\documentclass[10pt, journal, compsoc, twoside]{IEEEtran}
\IEEEoverridecommandlockouts
\usepackage{graphicx}
\usepackage{epstopdf}
\usepackage{cite}
\usepackage{amsmath,amssymb,amsfonts}
\usepackage{algorithmic}
\usepackage{graphicx}
\usepackage{textcomp}
\usepackage{xcolor}
\usepackage{enumitem,kantlipsum}
\usepackage{comment}
\usepackage{url}
\usepackage{amsthm}
\usepackage{balance}

\usepackage{caption}

\usepackage{subcaption}
\usepackage{multirow}
\usepackage{setspace}

\usepackage{soul}

\usepackage{mathtools}
\usepackage{booktabs}

\newtheorem{thm}{Theorem}
\newtheorem{lem}{Lemma}
\newtheorem{cor}{Corollary}
\theoremstyle{remark}
\newtheorem{rem}{Remark}
\theoremstyle{definition}

\newtheorem{asmp}{Assumption}

\newcommand{\xiaoxi}[1]{{\bf \color{red} #1}}
\newcommand{\weijie}[1]{{\bf \color{magenta} #1}}

\usepackage{fixltx2e}
\usepackage{accents}

\usepackage{mathtools}
\usepackage{bbm}
\usepackage{comment}
\usepackage{multirow}

\usepackage[linesnumbered,ruled,vlined]{algorithm2e}

\usepackage[tr]{optional}

\MakeRobust{\underaccent}
\def\BibTeX{{\rm B\kern-.05em{\sc i\kern-.025em b}\kern-.08em
		T\kern-.1667em\lower.7ex\hbox{E}\kern-.125emX}}

\begin{document}
	
% 	\title{Interplay of Batch Size and Local Update for Federated Learning in Resource Constrained Edge Computing Systems\\
% 		{\footnotesize \textsuperscript{*}Note: Sub-titles are not captured in Xplore and
% 			should not be used}
% 		\thanks{Identify applicable funding agency here. If none, delete this.}
% 	}
\title{DYNAMITE: Dynamic Interplay of Mini-Batch Size and Aggregation Frequency for Federated Learning with Static and Streaming Dataset}%\xiaoxi{we need to change the title}}

	\begin{comment}
	\author{\IEEEauthorblockN{1\textsuperscript{st} Given Name Surname}
		\IEEEauthorblockA{\textit{dept. name of organization (of Aff.)} \\
			\textit{name of organization (of Aff.)}\\
			City, Country \\
			email address or ORCID}
		\and
		\IEEEauthorblockN{2\textsuperscript{nd} Given Name Surname}
		\IEEEauthorblockA{\textit{dept. name of organization (of Aff.)} \\
			\textit{name of organization (of Aff.)}\\
			City, Country \\
			email address or ORCID}
		\and
		\IEEEauthorblockN{3\textsuperscript{rd} Given Name Surname}
		\IEEEauthorblockA{\textit{dept. name of organization (of Aff.)} \\
			\textit{name of organization (of Aff.)}\\
			City, Country \\
			email address or ORCID}
	}
	\end{comment}

\author{Weijie Liu,~\IEEEmembership{Student Member,~IEEE,}
        Xiaoxi Zhang,~\IEEEmembership{Member,~IEEE,}
        Jingpu Duan,~\IEEEmembership{Member,~IEEE,}
        Carlee Joe-Wong,~\IEEEmembership{Senior Member,~IEEE,}
        Zhi Zhou,~\IEEEmembership{Member,~IEEE,}
        and~Xu~Chen,~\IEEEmembership{Senior Member,~IEEE}% <-this % stops a space
        
\IEEEcompsocitemizethanks{
\IEEEcompsocthanksitem Weijie Liu, Xiaoxi Zhang, Zhi Zhou and Xu Chen are with the School of Computer Science and Engineering, Sun Yat-sen University, Guangzhou 510006,
China. (Corresponding author: Xiaoxi Zhang.) \protect\\
E-mail: liuwj55@mail2.sysu.edu.cn, \{zhangxx89, zhouzhi9, chenxu35\}
@mail.sysu.edu.cn.
\IEEEcompsocthanksitem Jingpu Duan is with the Institute of Future Networks, Southern University
of Science and Technology, Shenzhen 518055, China, and also with the
Department of Communications, Peng Cheng Laboratory, Shenzhen 518066,
China.\protect\\
E-mail: duanjp@sustech.edu.cn.
\IEEEcompsocthanksitem Carlee Joe-Wong is with Department of Electrical and
Computer Engineering, Carnegie Mellon University, CA 94035.\protect\\
E-mail: cjoewong@andrew.cmu.edu.
}% <-this % stops an unwanted space

% \thanks{Manuscript received 2023; revised 2023; accepted 2023; Date of publication 2023; date of current version 1 October 2023.\\
% This work extends our paper published in IEEE/ACM IWQoS 2023~\cite{weijie-iwqos23}, by additionally coping with dynamic data streams and limited edge storage, with a new algorithm DYNAMITE proposed and substantially more experiments conducted to verify our algorithm performance.\\
% Digital Object Identifier no. 10.1109/TMC.}
}

% \markboth{IEEE TRANSACTIONS ON MOBILE COMPUTING, VOL. XX, NO. XX, June 2023}%
% {Liu et al.: DYNAMITE: Dynamic Interplay of Mini-Batch Size and Aggregation Frequency for Federated Learning}

\IEEEtitleabstractindextext{
	\begin{abstract}
        Federated Learning (FL) is a distributed learning paradigm that can coordinate heterogeneous edge devices to perform model training without sharing private data. While prior works have focused on analyzing FL convergence with respect to hyperparameters like batch size and aggregation frequency, the joint effects of adjusting these parameters on model performance, training time, and resource consumption have been overlooked, especially when facing dynamic data streams and network characteristics. This paper introduces novel analytical models and optimization algorithms that leverage the interplay between batch size and aggregation frequency to navigate the trade-offs among convergence, cost, and completion time for dynamic FL training. We establish a new convergence bound for training error considering heterogeneous datasets across devices and derive closed-form solutions for co-optimized batch size and aggregation frequency that are consistent across all devices. Additionally, we design an efficient algorithm for assigning different batch configurations across devices, improving model accuracy and addressing the heterogeneity of both data and system characteristics. Further, we propose an adaptive control algorithm that dynamically estimates network states, efficiently samples appropriate data batches, and effectively adjusts batch sizes and aggregation frequency on the fly. Extensive experiments demonstrate the superiority of our offline optimal solutions and online adaptive algorithm.
	\end{abstract}

 	\begin{IEEEkeywords}
		Federated Learning, Edge Computing, Batch Size, Resource-Constrained
	\end{IEEEkeywords}
}
\maketitle
	
	\section{Introduction}
% More references are needed.
% Introduce FL and the edge environment, transition to system and statistical challenges. Mention the resource constraints
%Together with edge computing, Federated Learning (FL), as an emerging new learning paradigm, unleashes centralized computational resources for end devices. Servers can coordinate heterogeneous edge data and resources to deal with computation-intensive tasks, e.g., model training, while reduces transmission delay of massive data. 
%
\IEEEPARstart{F}{ederated} Learning (FL)~\cite{mcmahan2017communication,fedprox,li2019} has gained much attention as it enables distributed model training by multiple collaborative devices without exposing their raw data. In the meanwhile, with the increasing amount of data generated from different geographical locations and the proliferation of edge computing technologies~\cite{liu2021adaptive, ma2021adaptive}, deploying FL at edge devices has become a promising computation paradigm to facilitate data-driven applications (e.g., smart surveillance and personalized healthcare) while preserving data privacy. %\carlee{give example?}
Unlike traditional distributed machine learning (DML)~\cite{zhang2020machine,compress_gaia}, FL allows each training device (a.k.a. worker) to perform multiple local updates before uploading their model parameters to the central server in each aggregation round, and it does not require partitioning a central pool of data across distributed workers.

Despite its advantages, FL still faces two major challenges: 1) skewed distributions and unbalanced sizes of training data at different devices ({\em statistical challenge}), and 2) heterogeneous and limited edge resources ({\em system challenge}). The former is also referred to as non-independent-and-identical (non-i.i.d.) data, which has been analyzed for representative FL algorithms, especially FedAvg~\cite{li2019}. 
%However, they have not fully captured the non-i.i.d. properties in either their algorithm design or theoretical analyses. 
%However, existing FL algorithms fall short to account for the trade-offs between mini-batch sizes under these heterogeneity for practical edge FL.
Studies to address the system challenge have mainly focused on improving the learning efficiency by mitigating the impact of slow ``straggler'' devices on the wall-clock time of training and communication~\cite{ma2021adaptive,wang2019adaptive}.
%, with an important aspect being alleviating the problem of some devices being slow to complete their computations in an aggregation round, which can increase the wall-clock training time.  
%and navigate the trade-off among computation, communication and model accuracy in a more realistic environment, like resource constraints, non-i.i.d. data distribution, etc. 
%Besides, the computation and communication resources of edge systems are often limited. 
In addition, the cost due to either the energy consumed over a long training period \cite{mo2021energy,zeng2021energy} or operational charge paid to incentivize participating clients \cite{ruan2021wiopt,oort} can be prohibitive for FL at the edge~\cite{luo2021tackling}. 
%\carlee{Instead of ``bottleneck'' maybe just say they can be prohibitive/significant.} 
Thus, taking both time and cost into consideration when configuring training tasks on heterogeneous devices is of vital importance for FL algorithms.
To address these challenges simultaneously, we call for a full-fledged FL algorithm that can capture the three-way trade-off between convergence, training time, and cost expenditure. 
Recent works have analyzed the model convergence when varying different controls, e.g., balancing the number of local updates and aggregation rounds~\cite{wang2019adaptive}, or adjusting workers' mini-batch sizes under a time budget \cite{ma2021adaptive}, but these metrics are generally considered separately. 
%To address these challenges simultaneously, we call for a full-fledged FL algorithm that can capture the three-way trade-off between convergence, training time, and cost expenditure. 
In contrast, we propose to jointly optimize the aggregation frequency and mini-batch sizes, as they are the hyperparameters that determine the amount of data processed in each aggregation round and thus most affect these performance metrics.
%\carlee{maybe argue here that aggregation frequency and batch size are the hyperparameters that most affect runtime and cost (as well as convergence)--otherwise there may be questions about why these are what we optimize} \xiaoxi{Got it. Added the previous sentence.}

%Therefore, we propose to first quantify their impacts and then jointly optimize their interplay. 

\begin{figure}[t]
    \centering
    \includegraphics[width=8.5cm]{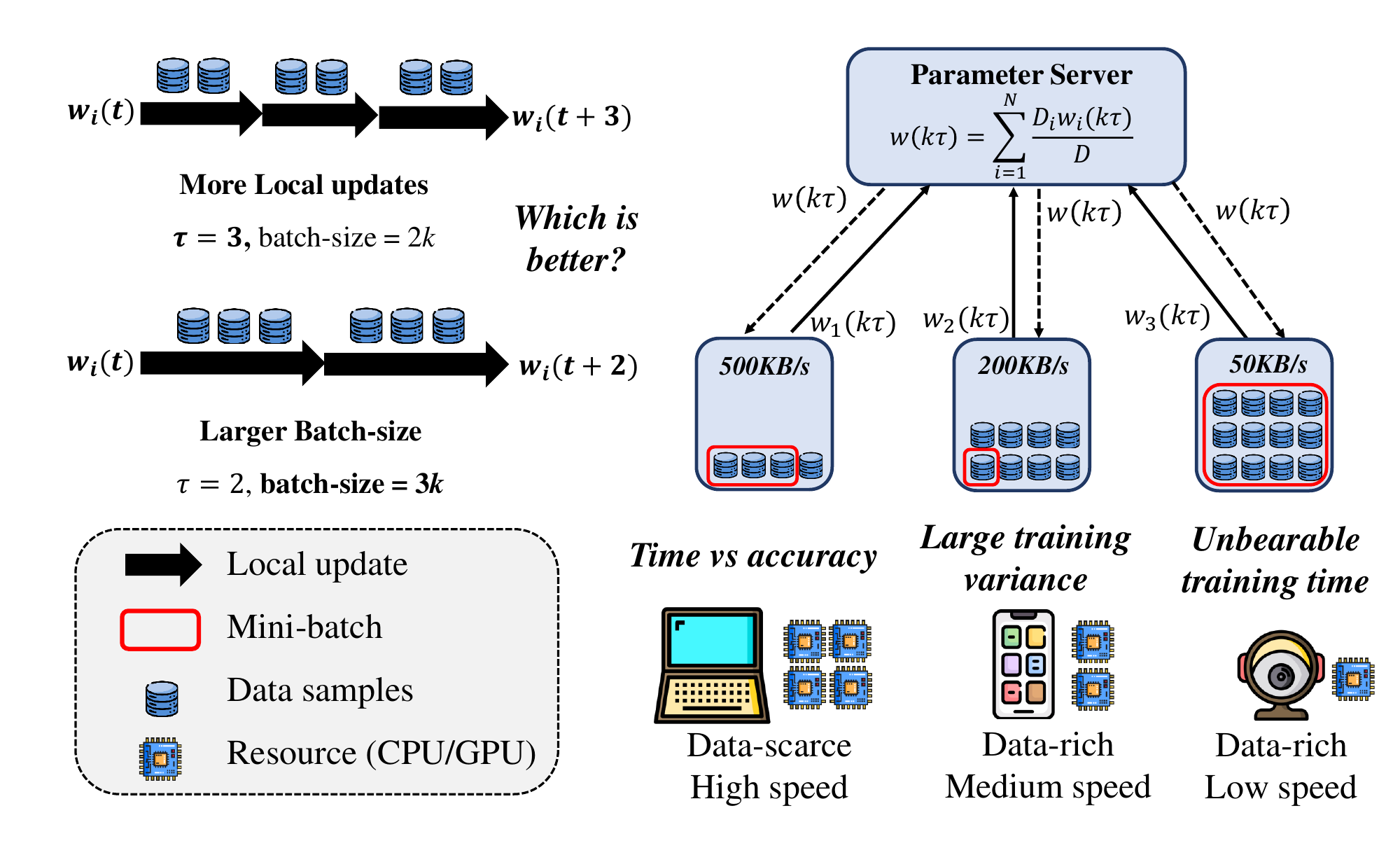}
    \caption{Left: Interplay of batch size and aggregation frequency; Right: Different batch sizes among clients with various computation and communication capabilities.}
    \label{fig:system-diagraph}
\end{figure}

%Bring up mini-batch size, number of local updates, and its potentially different contribution in striking a balance for the above mentioned three-way trade-off. Emphasize none of the literature well quantifies or addresses their interplay
% Besides, these prior theoretical works usually assume a {\em full-batch} training setting to achieve bounded convergence rates, but practical FL deployments generally adopt the mini-batch approach, motivating this work to study and bridge the gap of this inconsistency.  
%in their control algorithm and experimental validation, which brings huge inconsistency between the theoretical work and experiment. 
Further, we have the following intuitions about these performance metrics. % for synchronous FL.
%1) A larger mini-batch size and a larger number of local updates per round given the number of aggregation rounds both lead to more training samples and thus improve the local convergence of each client, but on the other hand increase the system overhead in different and perhaps complex forms. 2) Increasing the number of local updates may also result in a larger gap between the local and global models~\cite{li2019}, but this effect also depends on the batch size at each device. 
%with a larger mini-batch size and more local updates per aggregation round, the FL client can consume more training samples and obtain an improved local convergence. However, improved local convergence does not indicate better global performance. 
%For instance, increasing the number of local updates may result in a larger gap between the local and global models~\cite{li2019}, and this effect also depends on the batch size at each device. 
As illustrated in Figure \ref{fig:system-diagraph}-Left, increasing either the mini-batch size (interchangeably used with batch size in this paper) or the number of local updates can lead to more training samples processed and thus improve the local model accuracy. However, doing so can also increase the consumed cost and training time. 
%at the expense of a higher system overhead. %\carlee{It's not clear what ``local model convergence'' means.}
%However, improved local convergence does not indicate better global performance. 
Moreover, a larger number of local updates (lower aggregation frequency) may result in a larger gap between the local and global models~\cite{li2019}, hindering convergence, though this effect may depend on the batch size at each device. 
%Moreover, choosing these variables is even more complex if we consider time and cost as well. %\carlee{Maybe be more explicit and say that increasing either variable can increase per-iteration time and cost but might reduce the number of iterations needed for global model convergence. It would also be good to mention the effect on stragglers so that the transition to the next paragraph is smoother}
Therefore, we ask: {\em what is the best way to improve the FL model training when we can control both of these variables?} %, \weijie{as illustrated in Fig. \ref{fig:system-diagraph}}
% Prior theoretical works usually assume a {\em full-batch} training setting to achieve bounded convergence rates, but practical FL deployments generally adopt the mini-batch approach, motivating this work to bridge the gap of this inconsistency. 
To the best of our knowledge, this work proposes the first attempt to co-optimize mini-batch size and global aggregation frequency under dynamic edge networks, considering performance metrics of model accuracy, training time, and resource cost.

This work also reveals that strategically choosing {\em different mini-batch sizes among clients} is crucial to improve model accuracy, time, and cost expenditure. A motivating example might be performing an FL task for object detection on heterogeneous edge devices using their locally captured pictures. As illustrated in Figure \ref{fig:system-diagraph}-Right, the generally accepted ``no-straggler'' principle~\cite{ma2021adaptive}, which assigns the batch sizes of different FL devices for ensuring a uniform time per aggregation round \cite{ma2021adaptive,tyagi2020taming}, may not be optimal for this scenario.
%This principle always tends to assign more tasks, e.g. a bigger batch, to faster devices in order to alleviate the straggler effect . 
Specifically, the laptop with high training speed but relatively few data samples will have a large mini-batch while other data-rich devices such as the smartphone can only have a small mini-batch due to the relatively slow training speed. This could severely impede the convergence rate, as a small mini-batch size could introduce a high variance to the stochastic gradients (see Section \ref{sec:analysis}). 
On the other hand, if we neglect the clients' heterogeneous computing capacities %\carlee{hospitals aren't edge devices}
by simply setting a uniform batch size as what FL practitioners usually do~\cite{li2019,cohort_smith,luo2021tackling}, the straggler effects can be severe. Batch sizes, however, cannot help to limit battery usage and communication latency during model synchronization.
%Besides, these devices may have limited battery life and could incur high latency during communication for model synchronization. 
Therefore, jointly choosing the aggregation frequency in the meanwhile is also important for balancing the energy cost, training time, and model accuracy. %\carlee{``severe'' instead of unbearable? It might be bearable if the FL operator doesn't care much about training time.} . 
%Therefore, careful batch size assignment among the devices and jointly optimizing the aggregation frequency are crucial for navigating the trade-off between model accuracy and system overhead. 
To achieve this, we make the following \textbf{technical contributions}:
%To sum up, the main contributions in this paper are as follows.
%
\vspace{-0.5mm}
\begin{enumerate}[wide, labelwidth=!, labelindent=0pt]
\item \emph{New convergence bound with respect to batch size and global aggregation frequency (Section~\ref{sec:analysis}).}
We extend the FedAvg \cite{li2019} FL framework by allowing different clients to use different mini-batch sizes.  
%We analyze the interplay between mini-batch size and the number of local update steps, in terms of their combined effects on the model accuracy and resource consumption from a theoretical perspective. %
We capture FL clients' heterogeneity in the sizes and distributions of their datasets, based on which we then derive a novel convergence upper bound for the global model training, with respect to the aggregation frequency and batch sizes. Prior theoretical works usually assume a {\em full-batch} training setting to achieve bounded convergence rates, but practical FL deployments generally adopt the mini-batch approach. Our error bound can help bridge this inconsistency by quantifying the impacts of batch sizes considering the clients' heterogeneous data characteristics.

\item \emph{Novel closed-form results and algorithm design for co-optimizing the batch size and aggregation frequency (Section~\ref{sec:offline}).}
We propose an optimization model to capture the complex trade-offs among accuracy, completion (computation plus communication) time, and cost. 
%We first quantify the model accuracy w.r.t. mini-batch size and global aggregation frequency by using the above error bound, and 
Driven by our derived convergence bound, we provide closed-form solutions that co-optimize the batch size and aggregation frequency uniformly across clients. These results capture the interplay between these two control variables and can be easily adopted by FL developers.
We also propose an efficient algorithm to optimize {\em heterogeneous} batch sizes for different clients, which can further increase the model accuracy.%This also reveal that the ``no-straggler'' principle, which ensures a uniform time per aggregation round by assigning batch sizes of different clients \cite{ma2021adaptive,tyagi2020taming}, is not always effective. 

\item \emph{Online adaptive joint optimization algorithm (Sections~\ref{sec:online} and~\ref{sec:exp}).}
We design an adaptive control algorithm to dynamically choose the number of local updates and heterogeneous batch sizes among different clients, accommodating the online estimates of model convergence and system statistics across distributed training devices. Our algorithm can augment practical FL training strategies for both the cases of using static local datasets and dynamic data streams, with or without relying on limited data storage in a fluctuating edge network. Extensive experiments under different testbed settings demonstrate the superiority of our algorithms in terms of the accuracy, cost, and training time. 

\end{enumerate}

	\section{Related Work}
\label{sec:related}

% Xiaoxi: Due to space limit, this summary can be omitted.
%Our work is mainly related to prior works on convergence analysis for synchronous federated learning in resource-constrained edge computing systems, where theoretical results and control algorithm have been widely proposed and studied.

% Xiaoxi: This should be put into the intro
% Federated Learning was first proposed in \cite{mcmahan2017communication}, which can perform model training tasks by cooperating heterogeneous resources and data from geographically dispersed devices without transmitting private user data to a centralized data center. Unlike traditional distributed machine learning (DML), federated learning can update multiple local models and preserve the private data in every end device during the training process. Despite its advantages, federated learning still faces many challenges focused on two aspects: alleviating the straggler problem and navigating the trade-off among computation, communication and model accuracy in a more realistic environment, like resource constraints, non-i.i.d. data distribution, etc. 

{\bf Convergence analysis for FL} has been extensively studied in recent years. For instance, \cite{li2019} analyzes the convergence of the classic \emph{FedAvg} algorithm on non-i.i.d. data and establishes an $O(1/T)$ convergence bound for strongly convex and smooth problems. A refined FL framework \emph{FedProx}\cite{fedprox} has accounted for clients' different amounts of partial work, with provable convergence guarantees. Further, \cite{asyncFLopt} proves that the asynchronous \emph{FedAvg} has near-linear convergence to the global optimum for strongly convex optimization problems. A few other works propose FL algorithms and analysis for non-convex optimizations \cite{zhu2021delayed,reisizadeh2020fedpaq,fallah2020personalized}. %\cite{wang2019adaptive} obtained a novel convergence bound in FL with arbitrary local update steps between two global aggregations. 
These FL convergence analysis works mainly focus on the effect of the number of local updates or total number of iterations. % using either static or decay learning rate, mostly for convex objective functions.
%\xiaoxi{1. Introduce and cite 3-5 pioneer works, e.g., on the convergence, fedprox, etc for iid and non-iid data. These can be found in others' related works. 2. Emphasize that they analyzed the effect of local updates, total number of iterations, etc, for (mostly) convex objective functions, either using a static or decay learning rate. }

{\bf Improving the FL efficiency} has been studied in several directions, such as gradient compression \cite{compress_cmfl, compress_gaia, compress_qsgd, liu2020adaptivegradient} and hyperparameter selection \cite{ma2021adaptive, wang2019adaptive, adaptiveDRL}. %\carlee{maybe also mention client selection. Is this equivalent to a batch size or aggregation frequency of 0?} 
This work is orthogonal to the former (i.e., it can be combined with gradient compression), and falls in the latter regime, since we also aim to choose the best hyperparameters (i.e., batch-size and aggregation frequency). To optimize the learning speed, most studies choose hyperparameters to mitigate the effect of "straggler" devices, 
%mitigating straggler effect is one of the focal points in choosing hyperparameters. 
%in synchronous federated learning due to the computation and communication heterogeneity of the training devices. 
%In vanilla synchronous FL such as FedAvg, servers cannot execute the aggregation procedure until all the training devices upload their local updates, suffering from the straggler effect and low resource utilization.
 such as device sampling \cite{xia2020multi, ruan2021towards, luo2021tackling}, client selection \cite{oort,ruan2021wiopt}, and staleness control \cite{cipar2013solving,liu2021adaptive,ma2021fedsa}. 
 %Alternatively, a recent work \cite{ma2021adaptive} has presented an adaptive batch size selection algorithm to equalize the epoch time for each device.
 %Asynchronous \cite{liu2021adaptive} or semi-asynchronous FL \cite{ma2021fedsa} algorithm can also help by allowing stragglers to upload their local models asynchronously. 
%Partial update and device sampling strategies have been proposed to analyze the trade-off between the training devices and model accuracy.
%\xiaoxi{The following two sentences needs revision.}
%These methods may either significantly increase the training cost due to resource usage or sacrifice some training accuracy. 
Alternatively, recent works \cite{ma2021adaptive, park2022amble, shi2022talk} also consider optimizing batch sizes to improve FL efficiency by equalizing the epoch time for each device to mitigate the straggler effect. However, their works either lack theoretical analysis \cite{park2022amble} or neglect data heterogeneity and resource constraints across clients \cite{ma2021adaptive, shi2022talk}, which are important characteristics in edge systems. Several studies propose new strategies to handle streaming data for model training~\cite{jiang2022doll,zhou2020cefl,veloso2018self,defazio2014saga}. However, they do not focus on balancing the tradeoff among cost, accuracy, and training time.

{\bf Controlling FL under resource constraints} has risen as the main challenge for edge-enabled FL training.  
%Investigating the relationship among computation, communication and model accuracy in the edge computing environment instead of the data-center environment is another great challenge.
%
% Remove some sentences to shorten this section.
%
%Most of these prior results were proposed based on the theoretical analysis of different factors affecting the convergence rate and resource consumption. 
%Deploying federated learning at edge devices typically suffer from limited resources or high operational cost, which motivates 
An increasing number of studies have been proposed to improve FL accuracy under resource budgets, accounting for either completion time \cite{zhu2019broadband, wang2020optimizing, shi2020device} or operational cost \cite{mo2021energy, zeng2021energy}. Luo et al.~\cite{luo2021cost} propose a cost-effective FL design to choose the number of participants and local updates for total training cost minimization, respectively. Wang et al.~\cite{wang2019adaptive} derive a tractable convergence bound with an arbitrary number of local updates and design an algorithm for dynamically adjusting the aggregation frequency. Our work additionally analyzes the joint effect of mini-batch size on convergence, time, and cost metrics. A few recent works also consider choosing the mini-batch size. E.g, Ma et al. \cite{ma2021adaptive} propose a synchronous FL algorithm to adjust the batch size, and Liu et al. \cite{liu2020adaptivegradient} jointly optimize the batch size, gradient compression ratio, and spectrum allocation for wireless FL. Our work provides a new convergence bound with respect to {\em heterogeneous} clients' batch sizes and provides {\em both closed-form optimal solutions and online adaptive controls} for jointly selecting the aggregation frequency and batch sizes.

%a key factor in distributed gradient descent widely used in distributed machine learning, but always analyze the batch size in their algorithm or experimental validation. 

% Below is more suitable in intro, e.g., contributions. 
% Similar to the above research, our work focuses on the convergence rate and multiple resource consumption in non-i.i.d. synchronous FL, but in a more comprehensive way. We go beyond \cite{ma2021adaptive} and \cite{wang2019adaptive} by developing a convergence bound with both global aggregation frequency and arbitrary batch-size distribution scheme. In addition, we analyze the interplay between the these two factors and point out that the mini-batch size and local update steps could both explicitly affect the convergence rate in a statistical way in FL environment, then we propose a subtle, low-cost control algorithm to learn the optimal batch-size distribution scheme.

	\section{Preliminaries and Problem Formulation}
\label{sec:problem_formulation}
\subsection{Federated Learning}
We consider a parameter-server (PS) architecture, which consists of a set (defined as $\mathcal{N}$) of clients with $N = \left|\mathcal{N}\right|$ distributed edge devices (clients) and a centralized PS for global aggregation. Each device $i\in \mathcal{N}$ has a local data set $\mathcal{D}_i$ with $D_i$ data samples $\mathbf{x}_i=[\mathbf{x}_{i,1}, \mathbf{x}_{i,2}, ...,\mathbf{x}_{i,D_i}]$, and $\mathcal{D}_i$ is non-i.i.d. across $i$.
% Xiaoxi: Please verify the math definition in the following sentence.
We define the loss function for each sample $\mathbf{x}_{i,j}$ as $f(\mathbf{w},\mathbf{x}_{i,j})$ and the local loss function of device $i$ as:
\begin{equation}\label{eq:local_loss}
	F_i(\mathbf{w})=\frac{1}{D_i}\sum\limits_{j\in \mathcal{D}_i}f(\mathbf{w},\mathbf{x}_{i,j}).
\end{equation}

% Xiaoxi: please verify the revised paragraph. Besides, overlapping data samples seem fine if using this definition?
The ultimate goal is to train a shared (global) model $\mathbf{w}$ that minimizes the global loss function, defined as: 
% We denote $D_i$ and $D$ as the size of the local dataset $\mathcal{D}_i$ and the total dataset $\mathcal{D}=\sum_{i\in\mathcal{N}}D_i$ which includes all the data samples in every local dataset. We also assume that there is no overlap between two different local datasets and the global loss function can be defined as
%
\begin{equation}\label{eq:global_agg}
	F(\mathbf{w})=\sum\limits_{i\in\mathcal{N}}\frac{D_i}{D}F_i(\mathbf{w}),
\end{equation}
where $D$ is defined as $D=\sum_{i\in\mathcal{N}}D_i$.
%
% Given this global function $F(\mathbf{w})$, the FL problem is to find the optimal parameter $\mathbf{w}^*$, formally:
% %
% \begin{equation}\label{eq:fl_optimization}
% 	\mathbf{w}^* = \mathbf{argmin}_{\mathbf{w}}~F(\mathbf{w}).
% \end{equation}
%
% Xiaoxi: the following two paragraphs need to be shorten with important discussion possibly moved in the intro, as readers may skip this part if they already know FL well. 

As in the classic FedAvg\cite{mcmahan2017communication} framework, 
%based on the distributed gradient descent method also known as mini-batch SGD, which is one of the most common optimization algorithms in large-scale distributed machine learning tasks. 
clients divide their local data into mini-batches, perform multiple local updates, and upload their local models to the PS, which then broadcasts the updated global model to the clients by aggregating the local models. 
%Xiaoxi: merged below into the above paragraph to shorten the paper.
%The main difference between the traditional distributed machine learning algorithm and the FL algorithm is that every worker device in FL will perform $\tau$ rounds of local update steps ($ \tau>1$) before they upload their local model, while in DML, the server will perform global aggregation step after every local update step (i.e., $\tau=1$). %
%Recent works have investigated the effect of adaptive local update steps\cite{wang2019adaptive} and dynamic batch size \cite{ma2021adaptive} in federated learning individually. However, these p
Prior works either assume using the whole dataset for each round ({\em full-batch training})~\cite{wang2019adaptive,li2019} or simplify the effects of batch size on the convergence and training time in their analysis (e.g., \cite{ma2021adaptive}). 
%Therefore, they are unable to reveal the interplay between batch size and the number of local updates per round, which however is important to FL training on resource-constrained edge computing system as we will elaborate in Section \ref{sec:analysis}. 
Here we propose a more general FL setting by enabling customized batch sizes and the number of local updates. %\carlee{Maybe justify here why the number of local updates is the same for all users in this formulation} 
%All important notations in this paper are summarized in Table I.

\begin{table}[]
	\centering
	\caption{Main notations}
	\label{tab:my-table}
	\begin{tabular}{@{}ll@{}}
		\toprule
		$K$   & The number of communication rounds       \\ 
		$\tau$ & The number of local update steps     \\
		$T$   & The total number of iterations         \\
            $B_i$ & The maximum buffer size of device $i$     \\
            $\mathcal{B}_i$ & Buffer of device $i$     \\
		$s_i$  & Batch size of device $i$         \\
		$s_i^k$ & Batch size of device $i$ at round $k$ \\
		$\mathbf{s_k}$ & Batch size configuration at round $k$ \\
		$p_i$  & Computational capacity of device $i$ \\
		$t_{ci}$ & Computation time per update of device $i$      \\
		$t_{ui}$ & Communication time per round of device $i$    \\
		
		$\mathcal{D}_i$  & Local dataset of device $i$   \\
		$\mathcal{D}_i^k$  & Local data stream of device $i$ at round $k$    \\
		$\mathcal{D}$   & Entire dataset over all devices \\
		$\mathcal{D}_k$   & Entire data stream over all devices at round $k$\\
		$F(\mathbf{w})$& Global loss function \\
		$F_{i}(\mathbf{w})$& Local loss function  of device $i$    \\ 
		$F_{i,\mathcal{S}_i}(\mathbf{w})$& Local batch loss function of device $i$   \\ \bottomrule
	\end{tabular}
\end{table}

\subsection{Arbitrary batch size and aggregation frequency}

To capture different batch sizes across clients, we define the loss function $F_{i,\mathcal{S}_i}(\mathbf{w})$ under a mini-batch instead of the original local loss function $F_i(\mathbf{w})$ for each end device $i$:
\begin{equation}\label{eq:batch_loss}
	F_{i,\mathcal{S}_i}(\mathbf{w})=\frac{1}{s_i}\sum\limits_{j\in\mathcal{S}_i}f(\mathbf{w}, \mathbf{x}_{i,j}),
\end{equation}
where $\mathcal{S}_i$ denotes a mini-batch randomly selected from $\mathcal{D}_i$, and $s_i$ represents the size of $\mathcal{S}_i$. The full-batch training is a special case with $s_i=D_i$ and
%Here we note that the batch loss function generalizes the traditional loss function as 
$F_{i,\mathcal{S}_i}(\mathbf{w})=F_{i}(\mathbf{w})$. With a learning rate $\eta>0$, the local update rule is defined as:
\begin{equation}\label{eq:local_update}
	\mathbf{w}_i(t)=\mathbf{w}_i(t-1)-\eta g_i(\mathbf{w}_i(t-1)), t \neq k\tau
\end{equation}
where the batch gradient is $g_i(\mathbf{w}_i(t-1))\triangleq \nabla F_{i,\mathcal{S}_i}(\mathbf{w}_i(t-1))$. We consider a total of $K$ aggregation rounds (i.e., communication rounds) are performed in the FL training. The model update at each global aggregation step is:
 \begin{equation}\label{eq:aggreation}
 	\mathbf{w}(t)=\frac{\sum_{i=1}^{N}D_i\mathbf{w}_i(t)}{D},\, t=k\tau,
 \end{equation}
where $\tau$ is the number of local updates in each aggregation round, meaning that the PS only performs \eqref{eq:aggreation} and sends the global model $\mathbf{w}(t)$ to the clients at $t=k\tau,k=1,2,...,K$. 
%send this new global model back to all end devices for the following update round.
	
\subsection{Accuracy-time-and-cost joint optimization model}
\label{ssec:coopt-model}
% Xiaoxi: This subsection should be shorten and needs more references to justify the global cost constraint, explaining what kinds of cost we can capture and why they are important.  
%Our optimization problem mainly focused on two particular constraints, resource cost and completion time, which are the most critical factors in edge-enabled FL~\xiaoxi{cite references}.

%Compared to the traditional DML deployed in data centers, e
Compared to data centers, mobile edge devices usually have limited computing resources such as CPUs and GPUs. Their limited battery lives also restrict the energy available for FL. Moreover, edge devices in FL training often establish the connection with the PS through the Wide Area Network \cite{yuan2020hierarchical}%~\xiaoxi{cite papers}
, which could also incur high bandwidth costs in each communication round. 
%\carlee{This is not consistent with the hospital example in the introduction. Explicitly say that depending on whether we have a cross-silo or cross-device setting, we may have different costs and training times.} 
It is therefore necessary to consider both computation and communication costs. 

{\bf Limited budget for the total cost expenditure.} Formally, we suppose that $a$ units of computation cost are incurred (such as the cost of energy consumption) for processing a single sample, and $b$ units of bandwidth cost are consumed in each global aggregation step. %Then the consumed resource can be expressed as $K(a\tau s_{tot}+b)$ in total, where $K$ is the total communication rounds and 
Let $s_{tot}=\sum_{i\in\mathcal{N}}s_i$ represent the sum of batch sizes per iteration over all the devices. We consider that the total cost incurred by the entire training process cannot exceed a constant $R$, i.e., $K(a\tau s_{tot}+b)\le R$, which conforms to the definition of model training cost in \cite{schwartz2020green}. Here, $R$ can represent a cost budget of the energy consumption if the devices are owned by the FL owner, or the total rental fee of the edge devices if they are rented from another party. It can also be the budget of total monetary reward sent to participating clients~\cite{ruan2021wiopt}, e.g., for compensating clients' battery consumption and/or privacy losses~\cite{Li2017pricingdata}. %The parameters $a$ and $b$ represent the computation cost (per data sample) and the communication cost per aggregation round, respectively. 

{\bf Heterogeneous system capacities.} In practice, different edge devices can have heterogeneous computation and communication capacities, and %the straggler problem can be severe in synchronous FL, %Servers cannot perform the global aggregation step until all the selected clients have uploaded their local model, as 
the training time in each round is determined by the slowest device (straggler). Let $p_i$ denote the computation speed (number of samples processed per time) of device $i$. We then define $t_{ci}$ as the computation time of $i$ for a single local update and assume that it is proportional to the batch size, i.e., $t_{ci}= s_i/p_i$. Further, $t_{ui}$ is the communication time of each device $i$ incurred by synchronizing her local model with the PS. These definitions are consistent with practical system modelings  for FL training\cite{oort,ma2021adaptive}.
% \begin{equation}
% \label{eq:t_ci}
% 	t_{ci}=s_i/p_i, 
% \end{equation} 
% where $p_i$ is the computation speed (time per sample) of device $i$. %and the computation time is proportional to the batch size $s_i$, following the works~\xiaoxi{cite papers}. 
%Therefore the completion time constraint can be formulated as:
%
Suppose that the FL task owner has an expected completion time deadline $\theta$. We have the constraint on the completion time\footnote{To capture the randomness of computation/communication times of devices, \eqref{eq:time} can be re-written as the constraint on the expected completion time: $K\mathbb{E}[\max_i \left( \tau t_{ci}+t_{ui}\right)] \le \theta$; but for simplicity we consider that the variance of each device's runtime is small, compared with their differences among the devices, so that it suffices to consider posing the constraint on a deterministic form of the runtime, e.g., defining $t_{ci}$ and $t_{ui}$ as expected runtimes in the first place.}: %\carlee{This model is a bit unusual for one with stragglers, since the computation time is assumed to be deterministic. Maybe say that it can be written as $\mathbb{E}[\max_i time_i]$ but we assume deterministic times for simplicity, or because we assume inherently low compute capabilities at some devices dominate the effect of straggler noise.}
\begin{equation}\label{eq:time}
	\mathop{\mathbf{max}}\limits_{i\in\mathcal{N}}\;K(\tau t_{ci}+t_{ui})\leq\theta.
\end{equation}
Our goal is to find the optimal batch sizes $\mathbf{s}^*=[s_1,s_2,...,s_N]$ and the number of local update steps $\tau^*$ to minimize the gap between the expected global loss function $\mathbb{E}[F(\mathbf{w}(K\tau))]$
and the optimum $F^*$ after performing $K$ communication rounds, while satisfying the cost and completion time constraints. We define $[X]\triangleq\{1,\cdots, X\}$. Here we formulate the optimization problem as follows: 
\begin{align}
		\mathop{\textbf{Minimize}}_{\mathbf{s},\tau}\quad
		\mathbb{E}[F(\mathbf{w}(K\tau))]-F^*
		 \quad&\textbf{(Training error)}\label{eq:obj_origin}\\
		\textbf{S.t.}\quad
		\mathop{\mathbf{max}}\limits_{i\in\mathcal{N}}\;K(\tau t_{ci}+t_{ui})\leq\theta \quad&\textbf{(Completion time)}\label{eq:st_time}\\
		K(a\tau s_{tot}+b)\leq R \quad&\textbf{(Cost)}\label{eq:st_cost} \\
		s_i \in [D_i], \forall i , \tau \in [\tau_{max}] \quad&\textbf{(Feasibility)}\label{eq:st_fsb}
\end{align}
%where $R$ denotes the total resource budget.

To solve the above optimization problem, we need to first navigate the complex trade-offs among the expected error, completion time, and total cost incurred by the training process, via controlling our decision variables $\mathbf{s}$ (mini-batch size) and $\tau$ (the number of local updates).
%steps and global communication rounds, since it is obvious that all of these factors can directly increase the training time and system overhead but in different ways. Hence, we have to set these variables properly, especially in resource-constrained edge computing systems. More importantly, it is difficult for us to describe their combined effect and mutual interplay on the convergence rate in federated learning. 
We emphasize that, in addition to $\tau>1$  unlike centralized DML, edge FL faces heterogeneous distributions and sizes of local datasets ($D_i$), and thus may yield heterogeneous optimal mini-batch sizes $s_i$ across workers, which we shall show in Section \ref{sec:offline}. In contrast, the number of local updates $\tau$ needs to be uniform across clients, as unequal aggregation frequencies for different clients can cause objective inconsistency, i.e., the model converges to a mismatched objective function. While it is possible to address such inconsistencies during the aggregation process~\cite{wang2020inconsistency}, we do not consider such scenarios for the sake of simplicity. %\carlee{This can be corrected in the aggregation though. Maybe we can say that we do not consider such scenarios for simplicity.}
Our first challenge is then to simultaneously quantify the effects of the $\mathbf{s}$ and $\tau$ in the training error, formalized in our next section. %We analyze the convergence bound for arbitrary batch size and local update steps in Section IV, and solve the optimization problem in Section V based on this new bound. Further, we propose a low-cost control algorithm in Section VI to obtain a near-optimal batch size distribution and local update steps for online learning.
%\carlee{Explain why the aggregation frequency is assumed to be the same for each client, even if batch size is not.}

	\section{Training Error Bound Analysis}
\label{sec:analysis}
%Existing works do not provide training error analysis accounting for heterogeneous mini-batch sizes and the number of local updates simultaneously. 
In this section, we derive a new convergence bound to approximate \eqref{eq:obj_origin}, 
%inspired by \cite{wang2019adaptive,bottou} and additionally 
considering the effects of mini-batch sizes $s_i$ and the number of local update steps $\tau$. We first list our assumptions posed on the training model, which are generally adopted in pioneering FL works~\cite{kairouz2021advances,bottou}. We also evaluate the efficacy of our algorithm for training models that do not satisify these assumptions in Section \ref{sec:exp}.
%All the missing proofs can be found in \cite{TechReport}.

\begin{asmp}\emph{$\rho$-quadratic-continuous: For each client $i\in \mathcal{N}$ and some constant $\rho > 0$, the batch loss function $F_{i,\mathcal{S}_i}$ satisfies: }$\left \| F_{i,\mathcal{S}_i}(\mathbf{w}_1)-F_{i,\mathcal{S}_i}(\mathbf{w}_2)  \right \|\leq\rho\left \| \mathbf{w}_1-\mathbf{w}_2  \right \|_{2}^{2}\text{\emph{ for all }} \mathbf{w}_1, \mathbf{w}_2$.
\label{asmp1}
\end{asmp}
\begin{asmp}\emph{$\beta$-smooth: For each client $i\in \mathcal{N}$ and some constant $\beta > 0$, the batch loss function $F_{i,\mathcal{S}_i}$ satisfies: }$\left \| \nabla F_{i,\mathcal{S}_i}(\mathbf{w}_1)-\nabla F_{i,\mathcal{S}_i}(\mathbf{w}_2)  \right \| \leq \beta\left \| \mathbf{w}_1-\mathbf{w}_2  \right \| \text{\emph{ for all }} \mathbf{w}_1, \mathbf{w}_2$. \label{asmp2}
\end{asmp}
The local and global loss function satisfy the above assumptions straightforwardly due to the definition of $F_i(\cdot)$ and $F(\cdot)$.
\begin{asmp}\emph{Polyak-Łojasiewicz condition\cite{PL_condition}: There exists some constant $c$ that $ 0<c\le \beta ~\text{and}~ c \le 2\rho$, and for each device $i\in \mathcal{N}$, the global loss function $F(\mathbf{w})$ satisfies: }
\label{asmp3}
	$\left\|\nabla F(\mathbf{w})\right\|_2^2\geq2c(F(\mathbf{w})-F^*),~\forall \mathbf{w}$.
\end{asmp}
\begin{comment}
\begin{asmp}\emph{Strongly-convex: For each client $i\in \mathcal{N}$, $F_{i,\mathcal{S}_i}$ satisfies: }
\label{asmp3}
	$\exists 0<c\le \beta ~\text{and}~ c \le 2\rho, F_{i,\mathcal{S}_i}(\mathbf{w}_2)\geq F_{i,\mathcal{S}_i}(\mathbf{w}_1) + (\mathbf{w}_2-\mathbf{w}_1)^{T}\nabla F_{i,\mathcal{S}_i}(\mathbf{w}_1) + \frac{c}{2} \left \| \mathbf{w}_2-\mathbf{w}_1  \right \|^2  \text{\emph{ for all }} \mathbf{w}_1, \mathbf{w}_2$. 
\end{asmp}
\end{comment}
 % \carlee{Wouldn't satisfaction of these assumptions depend on the definition of the model? It might help to give examples of $F(w)$ and say we evaluate functions that do not satisfy these assumptions in our numerical evaluations. In particular the strong convexity does sort of contradict the quadratic continuity assumption, as strong convexity more or less imposes a quadratic lower-bound but quadratic continuity imposes a quadratic upper bound.}

\begin{asmp}\emph{First and Second Moment Limits: %For each client $i\in \mathcal{N}$, 
For some scalars $\mu_G\geq \mu>0$ and $M_i>0$, under any given model $\mathbf{w}$ and batch of data samples $\xi_t$ randomly selected from $\cup_i \mathcal{D}_i$ at step $t$, the global batch-gradient $g(\mathbf{w},\xi_t)$ and the variance of the gradient under any single data $\mathbf{x}_{i,j} \in \mathcal{D}_i$ of each client $i$, denoted as $\mathbb{V}_{\mathbf{x}_{i,j}}[\nabla f(\mathbf{w},\mathbf{x}_{i,j})]$, satisfy:} 
\label{asmp4}
	\begin{align*}
		\nabla F(\mathbf{w})^{T}\mathbb{E}_{\xi_t}[g(\mathbf{w},\xi_t)]\geq&\,\mu\left\|\nabla F(\mathbf{w})\right\|_2^2,\\
		\left\|\mathbb{E}_{\xi_t}[g(\mathbf{w},\xi_t)\right\|_{2}\leq&\,\mu_{G}\left\|\nabla F(\mathbf{w})\right\|_2,\\
		\mathbb{V}_{\mathbf{x}_{i,j}}[\nabla f(\mathbf{w},\mathbf{x}_{i,j})]\leq&\,M_i,~\forall i\in \mathcal{N}.
	\end{align*}
	%where $g_s(\mathbf{v}_{[k]}(t))$ represents a stochastic gradient computed with a single data sample.% and $\xi_t$ denotes a random seed to sample a set of data samples in mini-batch SGD.   
	%\xiaoxi{ Where's $\xi_t$ defined? Since $M_i$ is the maximum variance per sample, should we define $g_s$ first? }
\end{asmp}

\begin{asmp}{\emph{Bounded Gradient Divergence (non-i.i.d. degrees): Let $g(\mathbf{w})$ denote the global gradient under the dataset $\cup_i \mathcal{D}_i$. For some bounded scalar $\delta_{i}>0$, the local gradient of each client $i$ under her full dataset $\mathcal{D}_i$ satisfies:}}
\label{asmp5}
	\begin{equation*}
		\left \| g_{i}(\mathbf{w})-g(\mathbf{w})  \right \|\leq\delta_{i},~\forall \mathbf{w}, i.
	\end{equation*}
	%We also define $\delta=\sum_{i\in\mathcal{N}}\frac{D_i\delta_i}{D}$.
\end{asmp}

Based on the above assumptions, we show our first main result, an upper-bound of the training error with different batch sizes and uniform local update steps across devices, in the following theorem. 
\begin{thm}[Error bound with heterogeneous batch sizes $s_i$]\label{thm:bound-vanilla}
	Suppose that the loss functions satisfy Assumptions \ref{asmp1}--\ref{asmp5}. Assuming $F^*\geq0$, given a fixed learning rate $0\leq\eta\leq\frac{\mu}{\beta \mu_G^2}$ and the initial global parameter $\mathbf{w}(0)$, the expected error after $K$ aggregation rounds with $\tau$ local updates per round is:
	\begin{equation}\label{eq:bound_vanilla}
		\begin{split}
			\mathbb{E}[F(\mathbf{w}&(K\tau))]-F^*\leq q^{K\tau}\left[F(\mathbf{w}(0))-F^*\right]+\\
			&\frac{1-q^K}{1-q}\left(\frac{\beta\eta^2(1-q^{\tau})}{2D^2(1-q)}\sum\limits_{i\in \mathcal{N}}\frac{M_iD_i^2}{s_i}+\rho h(\tau)^2\right),
		\end{split}
	\end{equation}
where $q=1-\eta c\mu$, $h(\tau)=\frac{\delta}{\beta}\left((\eta \beta+1)^{\tau}-1\right)-\eta\delta \tau$, and $\delta=\sum_{i\in\mathcal{N}}\frac{D_i\delta_i}{D}$. Especially, when $\tau =1$, the above theorem is consistent with the DML convergence rate in prior works\cite{bottou}.
\end{thm}

We provide the full proof in Appendix \ref{sec:proof_vanilla}.

%We provide the basic idea of proving Theorem \ref{thm:bound-vanilla} below and defer the full proof to Appendix \ref{sec:proof_vanilla}.
\begin{comment}
\begin{proof}[Proof sketch]
 We first extend the convergence bound of DML \cite{li2019} with i.i.d. datasets and a uniform batch size to heterogeneous batch sizes in the non-i.i.d. scenario. In particular, we analyze the variance in model gradient computation among clients. Then, motivated by \cite{wang2019adaptive,ma2021adaptive}, we upper bound the gap of global loss between DML and FL, allowing clients to use different batch sizes. Formally, we capture the local bias resulted from local updates $\tau$, i.e., $F(\mathbf{w}(t))-F(\mathbf{v}_{[k]}(t))\leq \rho h(\tau)^2$, where $\mathbf{v}_{[k]}(t)$ is an auxiliary parameter vector that follows a centralized gradient descent. Finally, we combine these two bounds and apply them recursively over $K$ rounds to derive the bound w.r.t. both aggregation frequency and batch sizes. 
\end{proof}
\end{comment}

%\carlee{If si=Dis_i = D_i (i.e., full batch), are previously known bounds recoverd?} % not exactly the same since we combine the proof in both DML and FL.

Our bound (\ref{eq:bound_vanilla}) has a richer structure than those in \cite{wang2019adaptive,ma2021adaptive,liu2020adaptivegradient} to show the effects of $s_i$, $\tau$, and the data distributions. The first term is determined by the initial global loss, %qKτ[F(w(0))−F∗]q^{K\tau}\left[F(\mathbf{w}(0))-F^*\right], which  determined by the initial global parameter w(0)\mathbf{w}(0). This initial global loss 
which continuously decreases during the training process. The term associated with $M_i$ can be interpreted as the ``gradient variance loss'' resulting from the error of using a randomly selected batch to estimate the loss gradient under the entire local dataset. The last term $\rho h(\tau)^2$ can be regarded as the ``local bias'' which monotonically increases with $\tau$, since a larger $\tau$ means less frequent communications between the clients and server and thus a larger gap between the global and local models.

\section{An Offline Algorithm and Theories}
\label{sec:offline}
%In this section, we jointly optimize the mini-batch size $s$ and number of local updates per round $\tau$ in three cases. 
%While Theorem \ref{thm:bound-vanilla} provides a tractable way to optimize \eqref{eq:obj_origin}, to jointly optimize the batch size and number of local updates per round from the problem \eqref{eq:obj_origin}--\eqref{eq:st_fsb}, a more tractable way is to first assume that the parameters related to the model (in \eqref{eq:bound_vanilla}) and the edge system (in the optimization constraints) are known to us. Therefore, we consider the {\em offline} setting in this section, where all the above parameters are known constants. To 

In this section, we provide optimal solutions for co-optimized stationary batch sizes and the number of local updates in two cases. The total number of aggregation rounds $K$ is pre-determined in our problem~\cite{cohort_smith,wang2020inconsistency,ruan2021towards}. We assume they are all %\carlee{what does ``both'' refer to here?}
{\em offline} settings, where the parameters related to the model (in \eqref{eq:bound_vanilla}) and the system (in the optimization constraints) can be obtained, e.g., through pre-run tests~\cite{luo2021tackling, zhang2020machine}. We will design an adaptive control algorithm with parameters estimated online in Section \ref{sec:online}. 

%\carlee{``Given $K$'' is in each subsection heading and can probably be taken out of them and stated upfront instead}

\subsection{Case 1: Co-optimizing uniform $s$ and $\tau$}

We first consider the most common FL scenario in practice \cite{mcmahan2017communication ,fedprox} %~\xiaoxi{cite 2+ papers}
where every device has the same batch size $s$ and number of local updates per round $\tau$. %, under a given communication rounds $K$.  
%The optimization problem \eqref{eq:obj_origin}--\eqref{eq:st_fsb} can be simplified as follows: 
%
% \begin{align}\label{eq:opt_problem_offline}
% 		\mathop{\textbf{Minimize}}_{\mathbf{s},\tau}\quad
% 		 &\mathbb{E}[F(\mathbf{w}(K\tau))]-F^*
% 		\\
% 		\textbf{S.t.} \quad
% 		 &\mathop{\mathbf{max}}\limits_{i\in\mathcal{N}}\;K(\tau t_{ci}+t_{ui})\leq\theta,\, &t_{ci}=\frac{s_i}{p_i}\\
% 		&K(a\tau s_{tot}+b)\leq R \\
% 		&s \in [0,D_i], \tau \ge 0
% \end{align}
Based on our bound \eqref{eq:bound_vanilla}, we derive closed-form solutions of $s$ and $\tau$ in Theorem \ref{thm:opt_uniform}, by solving \eqref{eq:obj_origin}--\eqref{eq:st_fsb} with $s_i=s_{i'},~\forall i\neq i'$. 
\begin{thm}[Interplay of uniform $s$ and $\tau$]\label{thm:opt_uniform}
	Given the number of aggregation rounds $K$ and a feasible deadline ($\theta>Kt_{ui}$) and cost budget ($R>Kb$), the optimal uniform batch size $s^*$ and the number of local updates $\tau^*$ satisfy:
	\begin{align}
		&s^*(\tau) = \mathbf{min}\left\{\frac{R-Kb}{a\tau n},  \min\limits_{i\in\mathcal{N}}\left\{\frac{p_{i}(\theta-Kt_{ui})}{K\tau}\right\} \right\},\\
		&\tau_1=\lfloor\hat{\tau}\rfloor,\,\tau_2=\lceil\hat{\tau}\rceil,\,\frac{\partial f(\hat{\tau})}{\partial \tau} =0,\\
		&\tau^*=\mathop{\arg\min}\limits_{\tau\in\{\tau_1,\tau_2\}}f(\tau),\, s^*=\lfloor s^*(\tau^*)\rfloor,\label{eq:opt_tau}
	\end{align}
where $h(\tau)=\frac{\delta}{\beta}\left((\eta \beta+1)^{\tau}-1\right)-\eta\delta \tau$ and $f(\tau)=q^{K\tau}G(0)+\frac{1-q^K}{1-q}\left(\frac{\beta\eta^2(1-q^{\tau})}{2D^2(1-q)}\sum\limits_{i\in \mathcal{N}}\frac{M_iD_i^2}{s^*(\tau)}+\rho h(\tau)^2\right)$, $G(0) = \left[F(\mathbf{w}(0))-F^*\right]$ and $q = 1 -\eta c\mu$ as defined in Theorem~\ref{thm:bound-vanilla}.
\end{thm}
% 	\begin{equation*}
% 		s^*(\tau)= \mathbf{min}\left\{ \frac{R-Kb}{a\tau n}, \frac{p_{i}(\theta-Kt_{ui})}{K\tau} \right\},~\frac{\partial f(\tau^*)}{\partial \tau} =0,
% 	\end{equation*}

% 	\begin{equation*}
% 		\frac{\partial f(\tau^*)}{\partial \tau} =0
% 	\end{equation*}

%\begin{thm}[Uniform batch size, local update steps and communication rounds]
%	Suppose the total iteration $T$ is given, we can derive the optimal uniform batch size $s$, local update step $\tau$ and global aggregation round $K$ with time constraint $\theta$ and resource budget $R$.
%	\begin{equation*}
%		s^*(\tau)= \mathbf{min}\left\{ \frac{R-\frac{Tb}{\tau}}{aTN}, \min\limits_{i\in\mathcal{N}} p_i\left(\frac{\theta}{T}-\frac{t_{ui}}{\tau}\right)\right\}
%	\end{equation*}
%	
%	\begin{equation*}
%		\frac{\partial f(\tau^*)}{\partial \tau} =0, \quad K^*= \frac{T}{\tau^*}
%	\end{equation*}
%	\begin{equation*}
%		f(\tau) = q^{T}[F(\mathbf{w}(0))-F^*]+\frac{1-q^{T}}{1-q}\left(\frac{\beta\eta^2}{2D^2}\sum\limits_{i\in [N]}\frac{M_iD_i^2}{s^*(\tau)}+\rho h(\tau)\right)
%	\end{equation*}
%
%\end{thm}
We provide the full proof in Appendix \ref{sec:proof_uniform}.

%We provide the basic idea of proving Theorem \ref{thm:opt_uniform} below and defer the full proof to Appendix \ref{sec:proof_uniform}.
\begin{comment}
\begin{proof}[Proof Sketch]
Given that the objective function is monotonically decreasing with the client batch size, we can obtain the optimal uniform batch size $s^*(\tau)$ by finding its maximum value using the time and cost constraints. Then we can substitute $s^*(\tau)$ into the objective function, i.e., $f(\tau)$, which is a convex function when $\tau<2/log(1/q)$. Thus, we can solve the $\hat{\tau}$ to minimize the expected error bound by letting the derivative of $f(\tau)$ to be zero. However, $\hat{\tau}$ can be fractional, so we need to compare the values of $f(\lfloor\hat{\tau}\rfloor)$ and $f(\lceil\hat{\tau}\rceil)$ to find the optimal $\tau^*$. %Theorem \ref{thm:bs-gpu} can be proven by the same methods.
\end{proof}
\end{comment}
{\bf Our result quantitatively verifies an intuitive common practice that communicating with the PS every iteration ($\tau=1$) is the optimum, if the number of aggregation rounds $K$ and thus the total number of training iterations are sufficiently large (shown in Remark \ref{rmk:opt_tau_1}).}
\begin{rem}
\label{rmk:opt_tau_1}
	As the number of aggregation rounds $K$ increases, the optimal solution $\tau^*$, which is expressed in \eqref{eq:opt_tau}, will decrease to $1$, i.e., $\lim\limits_{K\to \infty}\tau^*=1$.
	%tend to decrease to even non-positive when $K\to \infty$. Thus, we have $\lim\limits_{K\to \infty}\tau^*=1$.
\end{rem}
\begin{proof}
When $K$ is small, the first term $q^{K\tau}G(0)$ dominates $f(\tau)$, and it monotonically decreases with $\tau$. As $K$ grows larger, the second term dominates $f(\tau)$ and monotonically increases with $\tau$. Thus the optimal local update steps $\tau^*$ decreases with the increase of $K$.
\end{proof}
%
%This is an quite reasonable and intuitive result which is also mentioned in Proposition 1 of \cite{wang2019adaptive}.
Remark \ref{rmk:opt_tau_1} can be intuitively explained as follows. 
When the number of communication round $K$ is small, e.g., due to the high communication cost or limited bandwidth, a bigger $\tau$ leads to a larger total number of model updates and thus higher accuracy while incurs limited communication cost. In contrast, if $K$ is sufficiently large, especially in the later stage of the training process, one should reduce $\tau$ and increase the batch size $s$, since a larger $\tau$ may increase the gap between the global and local models and thus incurs a larger final error. This implication is consistent with the intuition in \cite{wang2019adaptive}. %\carlee{This discussion is nice. We can also say here that the intuition of $s^\ast$ being as large as possible makes sense, to reduce the variance as permitted by the completion time/cost bounds.}

\subsection{Case 2: Co-optimizing $\tau$ and heterogeneous $s_i$}\label{sec:case3} %GPUs~\xiaoxi{why GPU}}
%\weijie{The title of this section and the beginning of this paragraph should be rephrased after the removal of the GPU case} In this case, we revisit our original optimization problem \eqref{eq:obj_origin}--\eqref{eq:st_fsb} without assuming training on powerful GPUs as Case 2. 
In this case, we generalize Case 1 by enabling different batch sizes assigned for different clients. Since edge devices can have different and potentially limited computation and communication capacities, %or even worse, multiple tasks compete for the same edge resources.
increasing the batch size at different clients lengthens the total computation time by different amounts. Following \cite{oort,ma2021adaptive}, the computation time of each step of local update can be modeled by $t_{ci} = s_i/p_i$.  
%This makes the co-optimization problem much more challenging than Case 2, and thus we fix $\tau$ in this case for better tractability. 
For the clarity of the following analysis, we first fix the value of $\tau$, then our optimization problem becomes:
%we cannot always expect that every edge participating device can always have sufficient and powerful GPU resources.
%
%Since the communication rounds $K$ and the local update steps $\tau$ are given, the above optimization problem in (19) is equivalent to the following convex problem.
%
% \textbf{Convex Problem}
\begin{align}
		\mathop{\textbf{Minimize}}_{\substack{\mathbf{s}=[s_1,s_2,...s_N]\\ \tau \in [\tau_{max}]}} \quad &\sum\limits_{i\in \mathcal{N}}\frac{M_iD_i^2}{s_i} \label{eq:obj_problem_case3}\\
		\mathbf{S.t.}\quad s_i &\leq p_i\left(\frac{\theta}{K\tau}-\frac{t_{ui}}{\tau}\right), s_i\in [D_i],~\forall i  \label{eq:st1_problem_case3}\\
		s_{tot}&=\sum\limits_{i\in \mathcal{N}}s_i \leq \left(R-Kb\right)/\left(a \tau\right)\label{eq:st2_problem_case3}
\end{align}
%Since \eqref{eq:obj_problem_case3}--\eqref{eq:st2_problem_case3} is a convex problem, we can solve it using traditional Lagrange multiplier method \cite{wiki:lagrange}. However, it can be quite complex and time-consuming. We instead 
Directly applying an integer programming optimizer such as Gurobi~\cite{gurobi} to solve \eqref{eq:obj_problem_case3}--\eqref{eq:st2_problem_case3} or using brute force algorithm may incur a high time complexity with at least $O(\kappa^{N}\tau_{max})$, where $\kappa = \frac{s_{tot}}{N}>>1$. Instead, we design a more efficient exact algorithm, as we state in the following theorem.
\begin{thm}\label{thm:opt_alg_1}
Given the number of aggregation rounds $K$ and the maximum number of local updates per round $\tau_{max}$, Algorithm \ref{alg:bs_dis} outputs the optimal batch sizes $\mathbf{s^*}$ and $\tau^*$ for FL training with at most $O(N^2 \tau_{max})$ time complexity. 
\end{thm}

Detailed proofs are all deferred to Appendix \ref{sec:proof_alg}.

{\bf Intuition of Algorithm \ref{alg:bs_dis}.} Since the objective function \eqref{eq:obj_problem_case3} decreases with $s_i$, the time constraint is transformed to \eqref{eq:st1_problem_case3}, which defines the largest batch size allowed for any device $i$ under deadline $\theta$, i.e., $s_{i}(\theta)=p_i\left(\frac{\theta}{K\tau}-\frac{t_{ui}}{\tau}\right)$. Similarly, the cost constraint \eqref{eq:st2_problem_case3} is equivalent to defining a total batch size under the cost budget $R$, i.e., $ \sum_{i\in\mathcal{N}} s_i =s_{tot}(R) =\frac{R-Kb}{a\tau}$. %To solve the optimal $\mathbf{s}$, we can start by 
If neglecting $s_{i}(\theta)$ firstly, 
%the optimal batch sizes $\mathbf{s^*}$ should satisfy $\frac{\sqrt{M_1}D_1}{s_1}=\frac{\sqrt{M_2}D_2}{s_2}=\cdots=\frac{\sqrt{M_N}D_N}{s_N}$, 
%Therefore, the optimal batch size $s_i^*$ of each device $i$ should be proportional to the constant $\sqrt{M_i}D_i$ related to its local dataset (Line 8), i.e., 
the Cauchy–Schwarz inequality yields: 
\begin{equation}
    \sum\limits_{i\in \mathcal{N}}\frac{M_iD_i^2}{s_i} \cdot \sum\limits_{i\in\mathcal{N}} s_i \ge \sum\limits_{i\in\mathcal{N}} \left(\sqrt{M_i}D_i\right)^2.
    \label{eq:MD}
\end{equation}
%Since the total batch size $\sum_{i\in\mathcal{N}} s_i = s_{tot}(R)$ is determined by the cost budget $R$, we 
Since $\sum_{i\in\mathcal{N}} \left(\sqrt{M_i}D_i\right)^2$ and $s_{tot}(R)$ are both constants, we can minimize the objective function $\sum_{i\in \mathcal{N}}\frac{M_iD_i^2}{s_i}$ when the equality holds with $\frac{\sqrt{M_1}D_1}{s_1}=\frac{\sqrt{M_2}D_2}{s_2}=\cdots=\frac{\sqrt{M_N}D_N}{s_N}$, i.e., $s_i\propto\sqrt{M_i}D_i$.
Then, considering $s_i(\theta)$, we need to reduce $s_i$ to $s_{i}(\theta)$ for {\em time-constrained devices} which have $s_i> s_{i}(\theta)$ (Lines \ref{line:s_violated}-\ref{line:reduce-s}). The $s_i$ of those devices will not be revised (Line \ref{line:remove}) since they reach the maximum allowed batch size. % have already had enough samples without violating the time constraint. 
In addition, we will re-assign (increase) the $s_i$ of other devices while keeping $s_i\le s_{i}(\theta)$ satisfied to make the best use of the extra data samples due to the reduced $s_i$ of those {\em time-constrained devices}, which is in fact a sub-problem of our original optimization problem. We can get the final solution by repeating the previous procedure recursively (Lines \ref{line:flag_0}-\ref{line:remove}), which can be proved optimal by using the Cauchy inequality again for $\sum_{i\in \mathcal{C}}s_i = s_{tot}(R)-\sum_{i\in \mathcal{N}\setminus \mathcal{C}} s_i(\theta)$, where $\mathcal{N}\setminus \mathcal{C}$ denotes the set of clients whose $s_i$ have been regulated to be equal to $s_i(\theta)$.
Since we always round down $s_i$ (line \ref{line_round}), we may still have some remaining resource budget. due to the round down operation in the previous steps.
We then increase the batch size of the device in the decreasing order of $\frac{M_iD_i^2}{s_i(s_i+1)}=\frac{M_iD_i^2}{s_i}-\frac{M_iD_i^2}{s_i+1}$ one at a time until the total batch size of all devices equals $s_{tot}(R)$ or $C=\emptyset$ (Lines \ref{line:re-assign}-\ref{line:re-assign_end}). Finally, we can find $\tau^*$ that yields the smallest error bound according to \eqref{eq:bound_vanilla}, by enumerating each feasible $\tau$ under which $s^*$ is optimized using the above method (Line \ref{line:search_tau}). 

\begin{algorithm}[t] \SetKwData{Left}{left}\SetKwData{This}{this}\SetKwData{Up}{up}
%\algsetup{linenosize=\tiny} \scriptsize
\SetKwFunction{Union}{Union}\SetKwFunction{FindCompress}{FindCompress} \SetKwInOut{Input}{Input}\SetKwInOut{Output}{Output}
	\Input{$\mathbf{G},M_i, D_i, K, \tau_{max}, a, b, R, \theta, p_i, t_{ui}, \forall i$ } 
	\Output{%Batch size distribution strategy 
	$\tau^*, \mathbf{s^*}=[s_1,s_2,...,s_N]$}
%	\BlankLine 
% \SetKwInOut{Initialize}{Initialize}	
% \Initialize{Set $C=\mathcal{N}$
% 	%as unassigned node,\; 
% 	$s_{tot}=\frac{R-Kb}{a\tau},  s_r=s_{tot}$}
	\ForEach{$\tau\in[1,\tau_{max}]$}{
	Set $C=\mathcal{N}$, $s_{tot}=\frac{R-Kb}{a\tau}, s_r=s_{tot}$\;
	\ForEach{node $i\in \mathcal{N}$}{$s_{i}(\theta)=\lfloor p_i\left(\frac{\theta}{K\tau}-\frac{t_{ui}}{\tau}\right)\rfloor$}
	%\emph{\,\tcc{The largest batch size for each $i$ under deadline $\theta$}}}
	\Repeat{$flag = 0$ or $C = \emptyset$}{ 
		$flag= 0$\label{line:flag_0}\;
		\ForEach {node $i\in C$}{\label{forins} 
			$s_i=\lfloor\frac{s_{r}\sqrt{M_i}D_i}{\sum_{i\in C}\sqrt{M_i}D_i}\rfloor$\label{line_round}\;
			
			\If{$s_i\geq s_{i}(\theta)$\label{line:s_violated}}{
				$s_i = s_{i}(\theta), s_r = s_{r}-s_i(\theta)$\label{line:reduce-s}\;
				Remove node $i$ from set $C$, $flag = 1$\label{line:remove}\; 
			}
		}
	}
	%\tcc{Assign remaining samples}
	\Repeat{$\sum_{i\in\mathcal{N}}s_i=s_{tot}$ or $C = \emptyset$\label{line:re-assign_end}}
	{Find $i^{'} = \text{argmax}_{i\in C}~ \frac{D_i^2}{s_i(s_i+1)}$, $s_{i^{'}}=s_{i^{'}}+1$ \label{line:re-assign} \; 
	%from the nodes $ i \in$ $C$ with the maximal $\frac{D_i^2}{s_i(s_i+1)}$
		\If{$s_{i^{'}}=s_{i^{'}}(\theta)$}{Remove node $i$ from set $C$\;}
	}

	}Find the optimum $(\tau^*,\mathbf{s}^*) = \text{argmin}_{(\tau,\mathbf{s})} \mathbf{G}$\label{line:search_tau}\; \tcc{Offline:$\mathbf{G}\triangleq$ \eqref{eq:bound_vanilla} Online:$\mathbf{G}\triangleq$ \eqref{eq:bound_marginal_approx}}
	
	\caption{An exact offline algorithm to Co-Optimize batch sizes and the number of local updates for FL training ({\bf CoOptFL})}
	\label{alg:bs_dis}
\end{algorithm}

%\begin{comment}
% In addition, for a special case where the computation time of each worker is a constant rather than $s_i/p_i$, e.g., training with powerful GPUs, Algorithm \ref{alg:bs_dis} can be simplified by using the following corollary to find the optimal $s_i^{\ast}$ and $\tau^{\ast}$.
In addition, we further consider a practical scenario where the FL training is performed with powerful GPUs, and the batch size should have little effect on the computation time  $t_{ci}$ \cite{cremanns2017deep}, which can be simplified to be a constant rather than $s_i/p_i$ in the time constraint \eqref{eq:st_time}. In this case, algorithm \ref{alg:bs_dis} can be simplified by using the following corollary to find the optimal $s_i^{\ast}$ and $\tau^{\ast}$. 

\begin{cor}[Optimal $\tau^* $ and $s_i^*$ with powerful GPUs]\label{cor:bs-gpu}
	Suppose that each client $i$ incurs a constant $t_{ci}$, e.g., running on powerful GPUs, the optimal $s_i^{\ast}$ and $\tau^{\ast}$ satisfy:  %(\xiaoxi{heterogeneous $t_{ci}$}?): 
	\begin{align}
		&s_i(\tau)=\left\lfloor\frac{s_{tot}(\tau)\sqrt{M_i}D_i}{\sum\limits_{i\in\mathcal{N}}\sqrt{M_i}D_i}\right\rfloor,\, s_{tot}(\tau)=\frac{R-Kb}{a\tau},\\
		&\tau_1=\lfloor\hat{\tau}\rfloor,\,\tau_2=\lceil\hat{\tau}\rceil,\,\frac{\partial f(\hat{\tau})}{\partial \tau} =0,\\ 
		&\tau^*=\min\left\{\mathop{\arg\min}\limits_{\tau\in\{\tau_1,\tau_2\}}f(\tau),\min\limits_{i\in\mathcal{N}}\left\{\frac{\theta-Kt_{ui}}{Kt_{ci}}\right\}\right\},\\
        &s_i^*=s_i(\tau^*),
% 		&s_i^* = \lfloor s_i(\tau^*)\rfloor
	\end{align}
where $h(\tau)=\frac{\delta}{\beta}\left((\eta \beta+1)^{\tau}-1\right)-\eta\delta \tau$, $f(\tau) = q^{K\tau}G(0)+\frac{1-q^K}{1-q}\left(\frac{\beta\eta^2(1-q^{\tau})}{2D^2(1-q)}\sum\limits_{i\in \mathcal{N}}\frac{M_iD_i^2}{s_i(\tau)}+\rho h(\tau)\right)$, $G(0) = \left[F(\mathbf{w}(0))-F^*\right]$ and $q = 1 -\eta c\mu$ as defined in Theorem~\ref{thm:bound-vanilla}.
\end{cor}
%\end{comment}

Detailed proof of Corollary \ref{cor:bs-gpu} is provided in Appendix \ref{sec:proof_uniform}.

{\bf Implication I.} If the time constraint is not the bottleneck (Line \ref{line_round} of Algorithm \ref{alg:bs_dis}), %or the training time is independent with $s_i$ (Corollary \ref{thm:bs-gpu}), 
$s_i^*$ is proportional to $\sqrt{M_i}D_i$. 
%This characterizes the heterogeneity of the sizes ($D_i$) and variances ($M_i$) of different clients' datasets. 
It is intuitive as devices with larger data sizes ($D_i$) have the potential to contribute more samples in each training iteration while more various data (with a larger $M_i$) needs a larger batch size to reduce the local variance of its computed gradients. %(details are proven \weijie{in appendix})% by Lemma 5 in \cite{TechReport}. 
{\bf This result reveals that either using full-batch ($s_i=D_i$) training \cite{li2019,ruan2021towards} or a uniform mini-batch size as FL practitioners usually adopt can be ineffective under non-i.i.d. clients' data.} %\xiaoxi{Moved the insight of the Theorem for GPU case here.}

{\bf Implication II.} 
%Compared to Case 2 and the existing works~\cite{ma2021adaptive,wang2019adaptive}, our Algorithm 1 is more comprehensive and realistic. Theorem \ref{thm:bs-gpu} captures the ``data heterogeneity'' of clients and sets the batch size for each device without considering its effect on computation time. We cannot do this in Case 3 since this could worsen the straggler problem and lead to low-utilization of the computation resources when considering limited resources. 
Other batch size assignment schemes (e.g.,\cite{ma2021adaptive}), on the other hand, focus on eliminating straggler effects brought by the system heterogeneity. They choose clients' batch sizes according to their computational capacity in order to minimize the average waiting time. {\bf However, this ``no-straggler'' strategy is sub-optimal when cost constraints are present, which are quite common in edge systems\cite{luo2021cost}. 
Using their strategy~\cite{ma2021adaptive}, devices with higher computation capacities but possibly a smaller $D_i\sqrt{M_i}$ of data always have bigger batch sizes, which could significantly undermine the model accuracy.} 
Our Algorithm 1 instead captures both the data heterogeneity $(D_i\sqrt{M_i})$ and system heterogeneity (mitigating the straggler effects), as well as navigating the trade-off between the completion time and resource consumption. %The algorithm can therefore simultaneously mitigate straggler effects and minimize the training error under a fixed cost budget.% and find the appropriate batch size distribution scheme within the time and resource constraints.

	\section{Online Adaptive Control Algorithm}
\label{sec:online}
%Xiaoxi: use `` '' rather than ' '

Section \ref{sec:offline} provides optimal solutions of batch sizes and the number of local updates, but it does not consider how to adapt them online with potentially unknown parameters, such as the computation speed $c_i$, communication time $t_{ui}$ (\emph{system dynamics}), and those associated with the model. Further, for the emerging applications of FL training under real-time data, e.g., video analytics~\cite{liu2020fedvision}, %\carlee{provide a citation of using FL for video here} 
we identify that limited on-device storage and online data streams (\emph{data dynamics}) need to be incorporated, especially for those performed on edge devices. For instance, the storage of smartphones can range from 128GB to 1TB, but the bit rates of data streams collected for running today's FL-supported video analytics tasks, can be as large as 3 gigabytes per minutes (1080p, 30fps, w/o compression). If the online training lasts longer, the edge storage may be used up\cite{gong2022ode} before the training ends. Moreover, it is not appropriate to use a static aggregation frequency and batch sizes solved from the offline optimization problem (see Section \ref{sec:offline}), given that the parameters related to the data and network dynamics are time-varying.

Therefore, in this section, we propose an adaptive algorithm to adjust $\mathbf{s}$ and $\tau$ at the beginning of each aggregation round, based on our online parameter estimation. It realizes a more practical edge FL training, supporting both conventional static heterogeneous local datasets and dynamic local data streams by considering fluctuating network characteristics (Section \ref{ssec:marginal_bound}). We also integrate a simple but efficient data sampling method to address the potential data insufficiency due to the limited storage of edge devices (Section \ref{ssec:par_est}). 
%will first introduce a ``marginal'' FL error bound different from the global error bound proposed in Theorem \ref{thm:bound-vanilla}. %and then apply it into our control algorithm. 
%We will then show the estimation of various parameters for this marginal bound, based on which we propose an online adaptive control algorithm to co-optimize the aggregation frequency and heterogeneous batch sizes.  

\subsection{FL with Streaming data and Limited Storage}
To adapt to the paradigm of federated learning on streaming data and limited on-device storage, we first slightly modify the definitions in traditional FL settings (see Section \ref{sec:problem_formulation}).

Similarly, we consider a parameter-server architecture, which consists of a set (defined as $\mathcal{N}$) of clients with $N$ distributed edge devices and a centralized PS for global aggregation. Each device $i\in \mathcal{N}$ has a local data stream $\mathcal{D}_{i}^{k}$ representing all $D_{i}^{k}$ data samples $\mathbf{x}_i=[\mathbf{x}_{i,1}, \mathbf{x}_{i,2}, ...,\mathbf{x}_{i,D_{i}^{k}}]$ that client $i$ received from round $1$ to round $k$, i.e. $\mathcal{D}_{i}^{1}\subseteq\mathcal{D}_{i}^{2}\subseteq...\subseteq\mathcal{D}_{i}^{k}$, and the local loss function of device $i$ and the global loss function at round $k$ can be defined as:
\begin{equation}\label{eq:local_loss_stream}
	F_{i}^{k}(\mathbf{w})=\frac{1}{D_{i}^{k}}\sum\limits_{j\in \mathcal{D}_{i}^{k}}f(\mathbf{w},\mathbf{x}_{i,j}).
\end{equation}

\begin{equation}\label{eq:global_agg_stream}
	F^{k}(\mathbf{w})=\sum\limits_{i\in\mathcal{N}}\frac{D_{i}^{k}}{D_{k}}F_{i}^{k}(\mathbf{w}),
\end{equation}
where $D_{k}$ is defined as $D_{k}=\sum_{i\in\mathcal{N}}D_{i}^{k}$. Besides, each client can select training samples from her local data stream $\mathcal{D}_{i}^{k}$ and store them into a buffer $\mathcal{B}_i$ with a limited size. %Thus, we have $\mathcal{B}_i\subseteq\mathcal{D}_{i}^{k}$ at every round $k$. 
Then we define the batch loss function $F_{i,\mathcal{S}_{i}^{k}}^{k}(\mathbf{w})$ under a mini-batch for each end device $i$:
\begin{equation}\label{eq:batch_loss_stream}
	F_{i,\mathcal{S}_{i}^{k}}^{k}(\mathbf{w})=\frac{1}{s_{i}^{k}}\sum\limits_{j\in\mathcal{S}_{i}^{k}}f(\mathbf{w}, \mathbf{x}_{i,j}).
\end{equation}

Unlike $\mathcal{S}_i$ in Eq.(\ref{eq:batch_loss}), $\mathcal{S}_{i}^{k}$ denotes a mini-batch randomly selected from the buffer $\mathcal{B}_i$ with dynamic and limited size of data samples of the client $i$ rather than its local data stream $\mathcal{D}_{i}^{k}$ due to the limited on-device storage; $B_i$ represents the maximum buffer size of $\mathcal{B}_i$ ($|\mathcal{B}_i|\leq B_i$) and $s_{i}^{k}$ denotes the size of the mini-batch $\mathcal{S}_{i}^{k}$. The buffer $\mathcal{B}_i$ of each client $i$ will be updated by sampling data samples from their local data stream in every communication round. With a learning rate $\eta>0$, we stick with the conventional rules of local update and global aggregation at round $k$, assuming $D_i^k$ is known to the corresponding client $i$:
\begin{equation}\label{eq:local_update_stream}
	\mathbf{w}_{i}(t)=\mathbf{w}_{i}(t-1)-\eta g_{i}^{k}(\mathbf{w}_{i}(t-1)),
\end{equation}
 \begin{equation}\label{eq:aggreation_stream}
 	\mathbf{w}(t)=\frac{\sum_{i=1}^{N}D_{i}^{k}\mathbf{w}_{i}(t)}{D_k},\, t=\sum_{k'=1}^k\tau_{k'}, \forall k = \{1,...,K\},
 \end{equation}
 %{\color{red}{It is unclear which value $\Tilde{K}$ should take, since (26-27) seem to be the update rules for any given $k$ rather than $\Tilde{K}$. So, should it be $t=\sum_{k'=1}^k\tau_{k'}, \forall k = \{1,...,K\}$}?}
where $g_{i}^{k}(\mathbf{w}_i(t-1))\triangleq \nabla F_{i,\mathcal{S}_{i}^{k}}^{k}(\mathbf{w}_i(t-1))$, and $\tau_k$ represents the number of local updates in communication round $k$.

The ultimate goal is to train a global model $\mathbf{w}$ that minimizes the global loss function $F^{K}(\mathbf{w})$ at the final round $K$ using data sampled from buffers $\mathcal{B}_i$ that have limited storage sizes and selectively store streaming data samples that incrementally arrive at the devices. %\carlee{does the test accuracy also change over time (i.e., is the optimal $F^\ast$ also changing)? It seems like it isn't from the below discussion.}

\subsection{Marginal Error bound and Problem Formulation}
\label{ssec:marginal_bound}
Revisiting our offline optimization problem \eqref{eq:obj_origin}--\eqref{eq:st_fsb}, the objective function derived in (\ref{eq:bound_vanilla}) with static parameters and decision variables %are all static and do not change during the training, which makes it 
is no longer suitable for our online setting.
%to guide the optimization problem in real environment. 
Therefore, we use a marginal upper bound, to quantify the gap between the optimum $F^*$ and the expected global loss %under the model 
that will be improved {\bf in aggregation round $k$}, formalized as $\mathbb{E}[F^{k}(\mathbf{w}^{(k)})] - F^*$. %\carlee{I think this needs more justification as to why we still evaluate convergence for the global model $F^\ast$. If the model is being used incrementally as it is being trained, then it would make more sense to evaluate it with respect to the current data distribution to date instead of the average over the entire datastream.} 
This performance metric is adopted to reflect the goal of making the best use of a limited buffer and adjusting our control variables $(\tau_k, \mathbf{s}_k)$ in order to be comparable with the optimal performance if having an unlimited size of buffer to store the entire updated dataset $\mathcal{D}_{i}^k, \forall i$. %~\xiaoxi{ $F(\mathbf{w}(K\tau))$ is simplified to be $F(\mathbf{w}^{(k)})$ here; $F(\mathbf{w}(T^{k-1}))$ is changed to $F(\mathbf{w}^{(k-1)})$, please verify}.
We derive the upper-bound of this gap in Lemma \ref{lem:marginal_bound}.

\begin{lem}[Marginal bound with heterogeneous batch size $s_i$ on streaming data]
\label{lem:marginal_bound}
	Suppose the loss function satisfies Assumptions \ref{asmp1}-\ref{asmp5} and $F^*\geq0$. %\carlee{are any of the parameters, e.g., for strong convexity, time-varying?} 
	For a fixed learning rate $0\leq\eta\leq\frac{\mu}{\beta\mu_{G}^2}$, %and given $G_{[k]}((k-1)$, \xiaoxi{$G_{[k]}((k-1)$ should be given, right?}
	the expected error of the empirical loss after $k$ global communication rounds with the number of updates $\tau_k$ and batch sizes $s_{i}^{k}$ for round $k$, defined as $\mathbb{E}[F^{k}(\mathbf{w}^{(k)})]-F^*$, is at most
	\begin{equation}\label{eq:bound_marginal}
		\begin{split}
			\mathbb{E}[F^{k}(\mathbf{w}^{(k)})]-F^*&\leq q^{\tau_k}\mathbb{E}[F^{k-1}(\mathbf{w}^{(k-1)})-F^*+\psi^{k}]\\
			&+\frac{\beta\eta^2(1-q^{\tau_k})}{2D_{k}^2(1-q)}\sum\limits_{i\in \mathcal{N}}\frac{M_{i}D_{i}^{k^2}}{s_{i}^{k}}+\rho h(\tau_k)^2,
		\end{split}
	\end{equation}
where $q=1-\eta c\mu$, $h(\tau_{k})=\frac{\delta}{\beta}\left((\eta \beta+1)^{\tau_k}-1\right)-\eta\delta \tau_k$, $F(\mathbf{w}^{(k-1)})\triangleq F(\mathbf{w}(\sum_{i=1}^{k-1}\tau_i))$, and $\psi^{k} = \mathbb{E}[F^{k}(\mathbf{w}^{(k-1)})-F^{k-1}(\mathbf{w}^{(k-1)})]$. 
% quantify the impact of the latest data samples in round $k$ on the global model, which can statistically describe the freshness and heterogeneity of these latest receiving data.  
\end{lem}

Detailed proof is deferred to Appendix \ref{sec:proof_vanilla}.

{\bf Intuition of Lemma \ref{lem:marginal_bound}.} Here, $\psi^k$ in \eqref{eq:bound_marginal} quantifies the impact of the latest data samples in round $k$ on the global model, which can statistically describe the freshness and heterogeneity of these latest receiving data. We point out that Lemma \ref{lem:marginal_bound} can easily adapt to static data set by setting $\psi^{k} = 0$ and remove all the superscripts $k$ and $k-1$ from $F(\cdot)$, $D$, and $s$. 
Compared to Theorem \ref{thm:bound-vanilla}, Lemma \ref{lem:marginal_bound} is defined for the setting, where the training data and network characteristics are time-varying. The lemma can well leverage the latest parameters collected from participated FL clients, and thus obtain better estimates of the unknown model and system parameters in each new aggregation round. 

Lemma \ref{lem:marginal_bound} provides the upper-bound of the error $\mathbb{E}[F^{k}(\mathbf{w}^{(k)})]-F^*$ incurred until the round $k$. Based on this, our optimization problem \eqref{eq:obj_origin}--\eqref{eq:st_fsb} can then be adapted to the following to solve for $\tau_k$ and $\mathbf{s}_k = [s_{1}^{k}, s_{2}^{k}, ...,s_{N}^{k}]$ used for each aggregation round $k\in [K]$.

\begin{align}
		\mathop{\textbf{Minimize}}_{\mathbf{s}_k,\tau_k}\quad
		&  \mathbb{E}[F^{k}(\mathbf{w}^{(k)})]-F^*
		\quad \text{(Approximated by \eqref{eq:bound_marginal})}\notag\\
		\textbf{S.t.} \
		& \mathop{\max}_{i\in\mathcal{N}}\;\sum_{k=1}^{K}(\tau_k s_{i}^{k}/p_i+t_{ui})\leq\theta,\, ~\forall i\label{eq:opt_problem_online}\\
		&\sum_{k=1}^{K}(a\tau_k \sum_{i\in \mathcal{N}}s_{i}^{k}+b)\leq R, s_{i}^{k}\le B_i,~\forall i\notag\\
            &\tau_k \in [\tau_{max}],~\forall k\notag
\end{align}
To solve \eqref{eq:opt_problem_online}, the remaining work is to estimate the unknown parameters $c_i, t_{ui}$, and those in \eqref{eq:bound_marginal} on the fly, as elaborated in Section \ref{ssec:par_est}.

\subsection{Online Parameter Estimation and Data Sampling}
\label{ssec:par_est}
 \begin{algorithm}[t] \SetKwData{Left}{left}\SetKwData{This}{this}\SetKwData{Up}{up} \SetKwFunction{Union}{Union}\SetKwFunction{FindCompress}{FindCompress} \SetKwInOut{Input}{Input}\SetKwInOut{Output}{Output}
	%\begin{small}
	\Input{$\theta, R, K, \tau_{max}, \eta$} 
	\Output{$\mathbf{w}(t)$}
\SetKwInOut{Initialize}{Initialize} \Initialize{$\theta_c\leftarrow\theta, R_c\leftarrow R, t \leftarrow0, k \leftarrow0$ \\ $\tau_1 \leftarrow1, \mathbf{w}(0), \mathbf{s_1}=[s_1^1,s_2^1,...,s_N^1]$}
    Receive $D_{i}^{0}, M_{i}$ from each device $ i\in \mathcal{N}$\;
    $D_0=\sum_{i=1}^{N}D_{i}^{0}$\;
	\Repeat{$k<K$ \textbf{or} $\theta_c<0$ \textbf{or} $R_c<0$ }{
		$k_0 \leftarrow k, k\leftarrow k+1$\;
		Send $\mathbf{w}(t),\tau_k,s_i^k,k$ to each node $i$\;
		$t_0 \leftarrow t$,\,%\emph{\quad //\,Save iteration index of last round.}\;
		%\tcc{Record the iter. index.}
		$t\leftarrow t+\tau_k$\;
		Receive $\mathbf{w}_i(t), p_i, D_{i}^{k}$ from each device $i\in \mathcal{N}$\;
            $D_k=\sum_{i=1}^{N}D_{i}^{k}$\;
		Execute global update according to (\ref{eq:aggreation_stream}) \;
		\If{$t_0>0$ \textbf{and} $k<K$} {
            \tcp{Parameters estimation}
			Receive from each device $i$: $\rho_i,\beta_i,c_i, M_{i}, F^{k_0}_{i,\mathcal{S}_i^{k_0}}(\mathbf{w}(t_0)), g_{i}^{k_0}(\mathbf{w}(t_0))$\label{line:server_collect}\;
			Calculate $g(\mathbf{w}(t_0))\leftarrow\frac{\sum_{i=1}^{N}D_{i}^{k_0}g_{i}^{k_0}(\mathbf{w}_i(t_0))}{D_{k_0}}$ 
			$\delta_{i}\leftarrow\left \| g_{i}^{k_0}(\mathbf{w}_i(t_0))-g(\mathbf{w}(t_0))  \right \|  $\label{line:PS_est_start} \;
			Estimate $\rho\leftarrow\frac{\sum_{i=1}^{N}D_{i}^{k_0}\rho_i}{D_{k_0}},\beta\leftarrow\frac{\sum_{i=1}^{N}D_{i}^{k_0}\beta_i}{D_{k_0}}$\label{line:server_para_start}\; 
			Estimate $c\leftarrow\frac{\sum_{i=1}^{N}D_{i}^{k_0}c_i}{D_{k_0}},\delta\leftarrow\frac{\sum_{i=1}^{N}D_{i}^{k_0}\delta_i}{D_{k_0}}$\label{line:server_para_end}\;
			Estimate remaining resources $\theta_c,R_c$ and communication time of each device $t_{ui}$\label{line:PS_est_end} \; 
			%\label{line:PS_est_end}\;\tcc{Linear search $\tau\in[1,\tau_m]$}
			Define function $\mathbf{G}$ to be \eqref{eq:bound_marginal_approx}\;
			$\tau_{k+1}, \mathbf{s}_{k+1}=$ {\bf CoOptFL}$(\mathbf{G}, M_{i}, D_{i}^{k}, K, \tau_{max}, a, b, R_c, \theta_c, p_i, t_{ui})$\label{line:CoOptFL}
		}
	}
	Send $STOP$ flag to all devices\;
	%\end{small}
	\caption{{\bf DYNAMITE} (Procedure at the PS)}
	\label{alg:online_server} 
\end{algorithm}

\DecMargin{1em}
\begin{algorithm}[t] \SetKwData{Left}{left}\SetKwData{This}{this}\SetKwData{Up}{up} \SetKwFunction{Union}{Union}\SetKwFunction{FindCompress}{FindCompress} \SetKwInOut{Input}{Input}\SetKwInOut{Output}{Output}
\SetKwInOut{Initialize}{Initialize}
\Initialize{$\mathcal{B}_i=\mathcal{D}_{i}^{0} ~(B_i \geq D_i^0), t \leftarrow 0$} 
    Estimate $M_{i}$ based on $\mathcal{D}_{i}^{0}$\;
	Send the stream size $D_{i}^{0}$ and $M_{i}$ to the server\;
	\Repeat{$STOP$ flag is received}{
		Receive $\mathbf{w}(t), \tau_k, s_{i}^{k}, k$ from the server\;
		$t_0 \leftarrow t$, $k_0 \leftarrow k-1$\;%\tcc{Record the iteration index}
% 		\emph{\quad //\,Save iteration index of last round.}\;
		\If{$t_0>0$}{
	%	\tcc{Estimate parameters}
		    $c_i\leftarrow\left\|\nabla F^{k_0}_{i,\mathcal{S}_i^{k_0}}(\mathbf{w}(t))\right\|^2/\,2F^{k_0}_{i,\mathcal{S}_i^{k_0}}(\mathbf{w}(t))$\label{line:client_est_start}\;
			$\rho_i\leftarrow \left \| F^{k_0}_{i,\mathcal{S}_i^{k_0}}(\mathbf{w}_i(t))-F^{k_0}_{i,\mathcal{S}_i^{k_0}}(\mathbf{w}(t))  \right \| /\left \| \mathbf{w}_i(t)-\mathbf{w}(t)  \right \|^2 $\;
			$\beta_i\leftarrow \left \| \nabla F^{k_0}_{i,\mathcal{S}_i^{k_0}}(\mathbf{w}_i(t))-\nabla F^{k_0}_{i,\mathcal{S}_i^{k_0}}(\mathbf{w}(t))  \right \| /\left \| \mathbf{w}_i(t)-\mathbf{w}(t)  \right \| $
		}
		$\mathbf{w}_i(t)\leftarrow\mathbf{w}(t)$\;
		$[\mathcal{B}_i, D_{i}^{k}] = \textbf{RS}(\mathcal{B}_{i}, B_i, D_{i}^{k_0})$\;
		\For{$r=1,2,...,\tau_k$}{
			$t\leftarrow t+1$\;
			Execute local update according to (\ref{eq:local_update_stream})\;
		}
		Record the average computation time $t_{ci}$ and estimate the computing capacity by $p_i = s_i/t_{ci}$\;
%		Estimate time and resource consumption $a_i, b_i, t_i, t_{u,i}, t_{b,i}$ for one local update at node $i$\;
		Send $\mathbf{w}_i(t), p_i, D_{i}^{k}$ to the parameter server\; 
		\If{$k_0>0$ \textbf{and} $F_{i,\mathcal{S}_i^{k}}^{k}(\mathbf{w}_i(t_0))-F^{k_0}_{i,\mathcal{S}_i^{k_0}}(\mathbf{w}(t_{0}))>\epsilon$}{
		Re-evaluate $M_{i}$ based on the current buffer $\mathcal{B}_{i}$\;
		}\label{line:M_recal}
		\If{$t_0>0$}{
			Send $\rho_i, \beta_i, c_i, M_{i}, F^{k_0}_{i,\mathcal{S}_i^{k_0}}(\mathbf{w}(t_0)),$ and $g^{k_0}_{i}(\mathbf{w}(t_0))$ to the PS \label{line:client_est_end}\;
		}
	}
	\caption{{\bf DYNAMITE} (Procedure at client $i$)}
	\label{alg:online_client}
\end{algorithm}
To simplify the problem \eqref{eq:opt_problem_online}, we first set $F^*=0$ %as a non-negative constant like 0 because 
as it is impossible to accurately evaluate it for model training. We then approximate the first term in (\ref{eq:bound_marginal}), i.e. $F^{k-1}(\mathbf{w}^{(k-1)})-F^*+\psi^{k} = F^{k}(\mathbf{w}^{(k-1)})=\frac{\sum_{i=1}^{N}D_i^kF_{i}^{k}(\mathbf{w}^{(k-1)})}{D_k}\approx\sum_{i=1}^{N}D_i^kF_{i,\mathcal{S}_i^{k-1}}^{k-1}(\mathbf{w}^{(k-1)})/D_k\triangleq \hat{F}^{k}(\mathbf{w}^{(k-1)})$ %\carlee{should the $\mathcal{S}_i$ have superscripts $k - 1$ here? Or are they cumulative over all prior rounds $t < k$?}
, by replacing the local loss $F_{i}^{k}(\cdot)$ with the batch loss $F_{i,\mathcal{S}_i^{k-1}}^{k-1}(\cdot)$, since it is impossible to calculate the local loss $F_{i}^{k}$ with $\mathcal{D}_{i}^{k}$ which have not been received at the end of the round $k-1$. Thus, we approximate $F_{i}^{k}(\cdot)$ by $F_i^{k-1}(\cdot)$, which uses $\mathcal{D}_i^{k-1}$ instead of $\mathcal{D}_i^{k}$, and we use $F_{i,\mathcal{S}_i^{k-1}}^{k-1}(\cdot)$ rather than $F_{i}^{k-1}(\cdot)$, since it can be quite time-consuming to calculate the exact value of $F_{i}^{k-1}(\cdot)$, %defined on the entire local dataset during the local update steps, 
especially when client $i$ has a large number of data samples. 
%\carlee{Does this mean you don't account for data distribution drift in the optimization? We could say that we make an assumption that the data distribution doesn't change much from round to round.} 
In this way, we use $\hat{F}^k(\mathbf{w}^{(k-1)})$ to capture the model drift due to the dynamic data, as it represents how good the old parameters $\mathbf{w}^{(k-1)}$ perform at the new data, which in turn reflects how well the model can generalize.

The estimation of $\rho, \beta, c$ and $\delta$ takes two steps. First, each client estimates these parameters $\rho_i, \beta_i, c_i$, and $g_i^{k}(\mathbf{w}(t))$ using the global model $\mathbf{w}(t)$ just received at the beginning of every round $k$ before synchronizing their local model $\mathbf{w}_i(t)$ with the global model. Consider that the network characteristics such as $t_{ui}$ and $t_{ci}$ are random variables. Since the training can take a large number of iterations, e.g., $10^5$, the estimates based on taking average of empirical measurements will be accurate, at least in probability converging to their true expectations, according to the law of large numbers. 
%\carlee{This is only true if you just care about the expected values of $t_{ui}$ and $t_{ci}$. That needs more justification for training time as training time involves the max over all clients.} 
One can also pick a good online estimation approach, such as OMD, FTRL, and bandits methods~\cite{oco}, which is not the focus of this work can thus omitted.
%\carlee{Estimating these parameters might be tricky in and of itself. Is there a requirement/assumption that the estimates be accurate? It may also be worth pointing out that the client doesn't give up much extra privacy by sending this information along with the gradient-based update.} 
Then the clients send these results back to the PS to calculate $\rho, \beta, c$ and $\delta$ as a weighted average of  $\rho_i, \beta_i, c_i$ and $\delta_i$ (see lines \ref{line:server_para_start}--\ref{line:server_para_end} in Algorithm \ref{alg:online_server}). Note that these parameter estimates do not expose extra information of clients' raw data beyond that exposed by sending the computed gradients. Finally our objective function (\ref{eq:bound_marginal}), %can be rewritten as $G(\tau, \mathbf{s})$ and 
can be approximated by the following error bound:
\begin{equation}\label{eq:bound_marginal_approx}
	\begin{aligned}
%		G(\tau, \mathbf{s})= 
%    q^{\tau_k}\frac{\sum_{i=1}^{N}D_iF_{i,\mathcal{S}_i}(\mathbf{w}^{(k-1)})}{D}
    q^{\tau_k}\hat{F}^{k}(\mathbf{w}^{(k-1)})
    + \frac{\beta\eta^2(1-q^{\tau_k})}{2D_{k}^2(1-q)}\sum\limits_{i\in \mathcal{N}}\frac{M_{i}D_{i}^{k^2}}{s_{i}^{k}}+\rho h(\tau_k)^2
	\end{aligned}
\end{equation}
where $q=1-\eta c\mu$, $h(\tau_{k})=\frac{\delta}{\beta}\left((\eta \beta+1)^{\tau_k}-1\right)-\eta\delta \tau_k$, $\hat{F}^{k}(\mathbf{w}^{(k-1)}) = \sum_{i=1}^{N}D_i^kF_{i,\mathcal{S}_i^{k-1}}^{k-1}(\mathbf{w}(\sum_{i=1}^{k-1}\tau_i))/D_k$.

For $\mu$ and $\mu_{G}$, when $g(\mathbf{w},\xi_t)$ is an unbiased estimate of $\nabla F(\mathbf{w})$, or $\nabla F_{i,\mathcal{S}_i^{k}}^k(\mathbf{w},\xi_t)$ is an unbiased estimate of $\nabla F_{i}^{k}(\mathbf{w})$, we have $\mu = \mu_{G} = 1$ , which are easily satisfied in the static data set case by randomly selecting each sample over the complete data set $D_{i}$ of each client $i$. However, this property can hardly hold in the streaming data case where each client $i$ can only select data samples from its limited buffer $\mathcal{B}_i$, after selecting data from their local data stream $\mathcal{D}_{i}^{k}$ and storing them at every FL round $k$. Regarding selecting data from $\mathcal{D}_{i}^k$ to $\mathcal{B}_i$, there are some straightforward data sampling methods, e.g., random sampling, which uniformly at random discards data stored in the buffer and replaces it with the latest-coming data, and FIFO sampling, which tends to preserve the data coming later while discarding those coming earlier. Such strategies will, however, inevitably lead to a biased global model. Hence, we adopt reservoir sampling \cite{vitter1985random} in our online algorithm to ensure that every data can have the same possibility of being stored in the buffer and thus selected into the batch during the whole training process. 

 \begin{algorithm}[h] \SetKwData{Left}{left}\SetKwData{This}{this}\SetKwData{Up}{up} \SetKwFunction{Union}{Union}\SetKwFunction{FindCompress}{FindCompress} \SetKwInOut{Input}{Input}\SetKwInOut{Output}{Output}
	%\begin{small}
	\Input{$\mathcal{B}_i, B_i, D_{i}^{k_0}$} 
	\Output{$\mathcal{B}_i, D_{i}^{k}$}
	\SetKwInOut{Initialize}{Initialize}
	\Initialize{$D_i^k = D_i^{k_0}$}
	\Repeat{the client performs local update steps.}{
	    	\For{every new data $\mathbf{x}$ received at round $k$ }{
                $D_{i}^{k} = D_{i}^{k} + 1$\;
                \eIf{$D_{i}^{k} < B_i$}{
                    Add $\mathbf{x}$ to the buffer $\mathcal{B}_i$\;
                }
                {Uniformly sample an integer $i$ in $[1,D_{i}^{k}]$\;
                    \eIf{$i \leq B_i$}{
                    Replace the $i_{th}$ data in buffer $\mathcal{B}_i$ with $\mathbf{x}$\;
                    }{Discard data $\mathbf{x}$\;
                    }
                }
		}
	}
	%\end{small}
	\caption{{\bf Reservoir Sampling (RS)}}
	\label{algo_reservoir} 
\end{algorithm}

We specifically note that these online parameter estimation based on the latest data stored in clients' buffers, especially for $\delta_i$ and $\delta$, can also help deal with the potentially changing gradient divergence or non-i.i.d. degree (Assumption \ref{asmp5}) across clients, showing the capability of our \textbf{DYNAMITE} of adapting with data distribution shift. %\carlee{Doesn't distribution shift aim to track an evolving data distribution? AdaCoOpt doesn't seem to do that; it appears to assume that the expected loss function is taken over the same data distributions.}

Many advanced data selection methods have also been proposed in prior works, such as loss-based sampling \cite{loshchilov2015online,shrivastava2016training}, importance-based sampling \cite{li2021sample}, gradient-norm sampling Mercury \cite{zeng2021mercury}, FedBalancer \cite{shin2022fedbalancer}, and the latest online streaming data selection method ODE\cite{gong2022ode}. However, these methods either need to evaluate all the data samples or incur a high time complexity during the selection process, which are not suitable for streaming data. %Besides, data selection is not the key contribution of this paper, and thus we do not take account of these complex sampling methods. Instead, 
Although data selection is not the focus of this work, our integrated sampling method (shown in Algorithm \ref{algo_reservoir}) is easy to implement and nicely preserves the good property (unbiased estimate of $\nabla F_{i}^{k}(\mathbf{w})$) that our online algorithm requires. We show the adaptiveness of our online control algorithm combined with different data sampling methods in experiments (Section \ref{sec:exp}).

\subsection{The workflow of our adaptive control algorithm}
    \label{ssec:AdaCoOpt}
	In this subsection, we present our Online Co-Optimization based FL algorithm, named {\bf DYNAMITE}, for the PS (Algorithm 2) and clients (Algorithm 3) to solve our refined batch size and aggregation frequency co-optimization problem shown in (\ref{eq:opt_problem_online}).%approximately within the resource and time budget by dynamically adjusting the local update steps and different batch size in edge nodes. 
	
	{\bf Algorithm \ref{alg:online_server}.} When the FL training starts, the PS initializes the remaining allowed completion time $\theta_c$ to be the deadline $\theta$, the remaining cost budget $R_c$ to be the total budget $R$, the current time $t$ to be zero, and the number of local updates per round $\tau$ to be one; the model weights are $\mathbf{w(0)}$ and batch sizes of all the clients ($\mathbf{s_1}$) are initialized to be the same. % (defined as $\mathbf{w}_0$ and $\mathbf{s_0}$).
	%and the size of every local dataset based on the system input and the message from each node. 
	In each aggregation round $k$, the server sends the global model $\mathbf{w}(t)$, number of local updates  $\tau_k$, batch size $s_{i}^{k}$, and current round index $k$ to the corresponding clients. Besides, it estimates the unknown parameters of the FL model (e.g., $\rho$, $\beta$, and $c$), network characteristics (e.g., $\theta_c, R_c$, and $t_{ui}$), and data distribution (non-i.i.d. degree $\delta$), shown in lines \ref{line:PS_est_start}--\ref{line:PS_est_end} of Algorithm \ref{alg:online_server}). After collecting related information from all the clients, the server updates $\tau$ and $\mathbf{s}$ using \textbf{CoOptFL} (line \ref{line:CoOptFL} in Algorithm \ref{alg:online_server}).   

	{\bf Algorithm \ref{alg:online_client}.} On the client side, each device $i$ first estimates $M_i$ through pre-run tests over its initial buffer $\mathcal{B}_i=\mathcal{D}_i^0$. Then it updates the buffer $\mathcal{B}_i$ by replacing the stored data with the new training data sampled from the local stream using reservoir sampling (Algorithm \ref{algo_reservoir}). Then, each client $i$ performs local updates and uploads the local model $\mathbf{w}_i(t)$ along with the estimated $c_i, \rho_i, \beta_i, M_{i},  F_{i,\mathcal{S}_i^{k_0}}^{k_0}(\mathbf{w}(t_0))$, and $g^{k_0}_{i}(\mathbf{w}(t_0))$, i.e.  $\nabla F_{i,\mathcal{S}_i^{k_0}}^{k_0}(\mathbf{w}(t_0))$, %after receiving the learning parameters and then upload the updated local model 
	to the PS (lines \ref{line:client_est_start}--\ref{line:client_est_end} of Algorithm \ref{alg:online_client}). We note that the client will re-evaluate the value of $M_i$ when the increase of the batch loss exceeds a threshold $\epsilon$ (Line \ref{line:M_recal} of Algorithm \ref{alg:online_client}), since it indicates that the local data distribution has changed significantly.  %~\xiaoxi{Complete this sentence, what parameters and which lines}.  
	%The server will then perform aggregation to update the global model and compute $\tau_{k+1}$ and $\mathbf{s}_{i}^{k+1}$ for the next round.
	%

 Finally, the server will perform aggregation step to update the global model and adopt our {\bf CoOptFL} (Algorithm \ref{alg:bs_dis}) with the estimated parameters to compute $\tau_{k+1}$ and $\mathbf{s}_{k+1}=[s_1^{k+1},s_2^{k+1},...,s_N^{k+1}]$ for all clients in the next round using the remaining budget $R_c$ and $\theta_c$ (line \ref{line:CoOptFL} in Algorithm \ref{alg:online_client}). %\xiaoxi{(replace R in \eqref{eq:opt_problem_online}?)}. 
	The key is to utilize the marginal error bound \eqref{eq:bound_marginal_approx} instead of the cumulative error bound \eqref{eq:bound_vanilla} when using our subroutine algorithm {\bf CoOptFL}. It finally outputs the optimal solution $(\tau^{*}$ and $\mathbf{s}^{*})$, %denoted by $(\tau^*,\mathbf{s}^*)$,
	which is the combination of ($\tau, \mathbf{s}$) that minimizes the value of (\ref{eq:bound_marginal_approx}) %(denoted by $G(\tau, \mathbf{s})$) 
	for the next aggregation round. Based on the adapted error bound shown in Lemma \ref{lem:marginal_bound} , this online adaptive control algorithm \textbf{DYNAMITE} (Algorithm \ref{alg:online_server} and \ref{alg:online_client}) also adapt to classic FL where clients have heterogeneous but static local data set $\mathcal{D}_i$ by removing the reservoir sampling process.% \weijie{We mention that the adaptive control of the batch size and local update step (Line \ref{line:CoOptFL} in Algorithm \ref{alg:online_client}) will perform in every $x$ communication rounds, where $x=1$ in our default algorithm but $x$ can be increased in practice to reduce excessive control overhead.}
	 
% 	 Motivated by \cite{wang2019adaptive}, we adopt a linear-search based method combined with our batch size optimization scheme (Algorithm \ref{alg:bs_dis}) round by round. %to find $\tau^*$ and $\mathbf{s}^*$ for next FL round effectively. 
% 	We first assume the value of $\tau=\hat{\tau}$ within a finite range $[1,\tau_m]$, where $\tau_m$ is a preset maximum value of $\tau$. For any feasible $\hat{\tau}_k\in[1,\tau_m]$, we can solve for the corresponding optimal $\hat{\mathbf{s}}_k$ using our Algorithm \ref{alg:bs_dis}, with $K$ and $R$ updated as the remaining communication rounds and cost budget. Then, the optimal pair $\left(\hat{\tau}_k, \hat{\mathbf{s}}_k\right)$ is the one that yields the minimum error bound \eqref{eq:bound_marginal_approx}.
	%Then we can find the optimal batch size distribution scheme $\hat{\mathbf{s}}$ corresponding to $\hat{\tau}$ with Algorithm \ref{alg:bs_dis}, using the remaining communication rounds and resource budgets. Thus, we can determine the optimal value of $\tau_{k}$ and $\mathbf{s}_k$ (denoted by $\tau^*$ and $\mathbf{s}^*$ by finding the best $\hat{\tau}$ and $\hat{\mathbf{s}}$ that minimizes $G(\tau, \mathbf{s})$ in (\ref{eq:bound_marginal_approx})

	\section{Experimental Validation}
\label{sec:exp}

In this section, we validate our theories and proposed algorithms in three parts: 1) Offline optimal local update step $\tau$ and uniform batch size $s$; 2) Optimal batch size assignment in CoOptFL (Algorithm \ref{alg:bs_dis}); 3) Online adaptive control algorithm DYNAMITE (Algorithms \ref{alg:online_server} and \ref{alg:online_client}) presented in Section~\ref{ssec:AdaCoOpt}. For the online adaptive control algorithm, we conduct experiments on both static datasets and dynamic data streams to demonstrate the superiority of our proposed algorithm \textbf{DYNAMITE}.
%In this section, we validate our theories and proposed algorithms in three parts: 1) Offline optimal local update step $\tau$ and uniform batch size $s$. 2) Offline optimal batch size assignment (Algorithm \ref{alg:bs_dis}); 3) Online adaptive control algorithm AdaCoOpt (Algorithms \ref{algo_disjdecomp} and \ref{alg:online_client}), presented and explained in Section~\ref{ssec:exp_results}. Before showing these, we first introduce our experiment setup in Section \ref{ssec:exp_setup}. %and then present our experimental results with explanations. 

\subsection{Experiment setup}
%Before showing the evaluation results, we first introduce our experiment setup.

\label{ssec:exp_setup}
\subsubsection{Testbed}
To simulate the system heterogeneity, we first conduct our experiments in a small-scale testbed with various types of edge devices, including 1 laptop PC (CPU: Intel i5-7300HQ 4-core @2.50GHz), 1 desktop PC (CPU: Intel i5-1135G7 8-core @2.40GHz), and 3 docker containers~\cite{merkel2014docker} launched from a workstation. We manually assign different numbers of CPU cores (3, 6, 12) to each container. The PS instance is deployed on the container with the most CPU cores, while the rest of the containers and devices are used as clients. 

To further evaluate our proposed algorithms \textbf{CoOptFL} and \textbf{DYNAMITE}, we conduct two larger scale experiments: 1) 100 clients simulated in a lab server cluster; and 2) a 20-client testbed deployed at 20 geo-distributed VM instances rented from Hetzner\cite{hetzner}, including six 1-vCPU instance (2GB RAM, 20GB storage), seven 2-vCPU instances (4GB RAM, 40GB storage), and seven 4-vCPU instances (8GB RAM, 80GB storage) for reflecting computational heterogeneity among clients. We deploy our PS on one of the 4-core instances. %\carlee{This still sounds rather small (maybe it's okay for cross-silo settings). Is it possible to simulate 100 clients by sampling their runtimes from the traces on these machines? What was the setup for measuring communication times?}
%To simulate the system heterogeneity, we conduct our experiments on various types of edge devices, including 1 laptop (CPU:Intel(R) i5-7300HQ 4-core @2.50GHz), 1 desktop PC (Intel(R) Core(TM) i5-1135G7 8-core @2.40GHz), and 3 docker containers on a workstation. We manually assign different numbers of CPU cores (3, 6, 12) to each docker and deploy the PS instance on the one with the most cpu cores while the remaining dockers and devices are specified as clients. In addition, we further conduct a larger scale experiment for the control algorithm, where 10 docker containers are launched as 10 clients, with the numbers and CPU configurations are (4 (1-core), 4 (2-core), 2 (4-core)). %~\xiaoxi{verify this}
%We implement all FL training models with Tensorflow. 

\begin{table*}[t]
\caption{Main parameter set-up. ``Time (s)'', ``time-c'', and ``cost-c'' represent time budget, time-constrained, and cost-constrained, respectively; Smooth-I (or Smooth-C) represents Smooth data arrival pattern in terms of the number of data samples per round and I.I.D. (or Continuous) configuration in terms of the arriving feature classes. Arrival rate: ``5k 100r'' denotes that 5000 samples will be received in every 100 rounds (Smooth) or in the 100th round (Burst).} %\carlee{what is ``r'' in the arrival rate rows?}}
\label{tab:param}
\begin{tabular}{|cc|ccccc|ccccc|}
\hline
\multicolumn{2}{|c|}{Dataset}                                                                          & \multicolumn{5}{c|}{MNIST/EMNIST}                                                                                                                                                                                                                                & \multicolumn{5}{c|}{CIFAR}                                                                                                                                                                                                                                        \\ \hline
\multicolumn{2}{|c|}{Configuration}                                                                    & \multicolumn{1}{c|}{Static} & \multicolumn{1}{c|}{Smooth-I}                                          & \multicolumn{1}{c|}{Smooth-C}                                          & \multicolumn{1}{c|}{Burst}                                              & Random & \multicolumn{1}{c|}{Static} & \multicolumn{1}{c|}{Smooth-I}                                          & \multicolumn{1}{c|}{Smooth-C}                                          & \multicolumn{1}{c|}{Burst}                                               & Random \\ \hline
\multicolumn{1}{|c|}{\multirow{2}{*}{\begin{tabular}[c]{@{}c@{}}Time\\ (s)\end{tabular}}}     & time-c & \multicolumn{1}{c|}{1k}     & \multicolumn{1}{c|}{500}                                               & \multicolumn{1}{c|}{800}                                               & \multicolumn{1}{c|}{500}                                                & 800    & \multicolumn{1}{c|}{1k}     & \multicolumn{1}{c|}{800}                                               & \multicolumn{1}{c|}{1k}                                                & \multicolumn{1}{c|}{500}                                                 & 1k     \\ \cline{2-12} 
\multicolumn{1}{|c|}{}                                                                        & cost-c & \multicolumn{1}{c|}{5k}     & \multicolumn{1}{c|}{5k}                                                & \multicolumn{1}{c|}{5k}                                                & \multicolumn{1}{c|}{5k}                                                 & 5k     & \multicolumn{1}{c|}{5k}     & \multicolumn{1}{c|}{5k}                                                & \multicolumn{1}{c|}{5k}                                                & \multicolumn{1}{c|}{5k}                                                  & 5k     \\ \hline
\multicolumn{1}{|c|}{\multirow{2}{*}{Cost}}                                                   & time-c & \multicolumn{1}{c|}{80k}    & \multicolumn{1}{c|}{80k}                                               & \multicolumn{1}{c|}{80k}                                               & \multicolumn{1}{c|}{80k}                                                & 80k    & \multicolumn{1}{c|}{80k}    & \multicolumn{1}{c|}{80k}                                               & \multicolumn{1}{c|}{80k}                                               & \multicolumn{1}{c|}{80k}                                                 & 80k    \\ \cline{2-12} 
\multicolumn{1}{|c|}{}                                                                        & cost-c & \multicolumn{1}{c|}{40k}    & \multicolumn{1}{c|}{32k}                                               & \multicolumn{1}{c|}{32k}                                               & \multicolumn{1}{c|}{32k}                                                & 32k    & \multicolumn{1}{c|}{40k}    & \multicolumn{1}{c|}{48k}                                               & \multicolumn{1}{c|}{48k}                                               & \multicolumn{1}{c|}{48k}                                                 & 48k    \\ \hline
\multicolumn{1}{|c|}{\multirow{2}{*}{\begin{tabular}[c]{@{}c@{}}Buffer\\ Size\end{tabular}}}  & time-c & \multicolumn{1}{c|}{-}      & \multicolumn{1}{c|}{10k}                                               & \multicolumn{1}{c|}{10k}                                               & \multicolumn{1}{c|}{25k}                                                & 10k    & \multicolumn{1}{c|}{-}      & \multicolumn{1}{c|}{10k}                                               & \multicolumn{1}{c|}{10k}                                               & \multicolumn{1}{c|}{25k}                                                 & 10k    \\ \cline{2-12} 
\multicolumn{1}{|c|}{}                                                                        & cost-c & \multicolumn{1}{c|}{-}      & \multicolumn{1}{c|}{10k}                                               & \multicolumn{1}{c|}{10k}                                               & \multicolumn{1}{c|}{25k}                                                & 10k    & \multicolumn{1}{c|}{-}      & \multicolumn{1}{c|}{10k}                                               & \multicolumn{1}{c|}{10k}                                               & \multicolumn{1}{c|}{25k}                                                 & 10k    \\ \hline
\multicolumn{1}{|c|}{\multirow{2}{*}{\begin{tabular}[c]{@{}c@{}}Step\\ size\end{tabular}}}    & time-c & \multicolumn{1}{c|}{5e-3}   & \multicolumn{1}{c|}{5e-3}                                              & \multicolumn{1}{c|}{5e-3}                                              & \multicolumn{1}{c|}{5e-3}                                               & 5e-3   & \multicolumn{1}{c|}{5e-3}   & \multicolumn{1}{c|}{5e-3}                                              & \multicolumn{1}{c|}{5e-3}                                              & \multicolumn{1}{c|}{5e-3}                                                & 5e-3   \\ \cline{2-12} 
\multicolumn{1}{|c|}{}                                                                        & cost-c & \multicolumn{1}{c|}{5e-3}   & \multicolumn{1}{c|}{5e-3}                                              & \multicolumn{1}{c|}{5e-3}                                              & \multicolumn{1}{c|}{5e-3}                                               & 5e-3   & \multicolumn{1}{c|}{5e-3}   & \multicolumn{1}{c|}{5e-3}                                              & \multicolumn{1}{c|}{5e-3}                                              & \multicolumn{1}{c|}{5e-3}                                                & 5e-3   \\ \hline
\multicolumn{1}{|c|}{\multirow{2}{*}{\begin{tabular}[c]{@{}c@{}}Arrival\\ rate\end{tabular}}} & time-c & \multicolumn{1}{c|}{-}      & \multicolumn{1}{c|}{\begin{tabular}[c]{@{}c@{}}5k\\ 100r\end{tabular}} & \multicolumn{1}{c|}{\begin{tabular}[c]{@{}c@{}}5k\\ 100r\end{tabular}} & \multicolumn{1}{c|}{\begin{tabular}[c]{@{}c@{}}50k\\ 500r\end{tabular}} & -      & \multicolumn{1}{c|}{-}      & \multicolumn{1}{c|}{\begin{tabular}[c]{@{}c@{}}3k\\ 100r\end{tabular}} & \multicolumn{1}{c|}{\begin{tabular}[c]{@{}c@{}}3k\\ 100r\end{tabular}} & \multicolumn{1}{c|}{\begin{tabular}[c]{@{}c@{}}25k\\ 1000r\end{tabular}} & -      \\ \cline{2-12} 
\multicolumn{1}{|c|}{}                                                                        & cost-c & \multicolumn{1}{c|}{-}      & \multicolumn{1}{c|}{\begin{tabular}[c]{@{}c@{}}5k\\ 100r\end{tabular}} & \multicolumn{1}{c|}{\begin{tabular}[c]{@{}c@{}}5k\\ 100r\end{tabular}} & \multicolumn{1}{c|}{\begin{tabular}[c]{@{}c@{}}50k\\ 500r\end{tabular}} & -      & \multicolumn{1}{c|}{-}      & \multicolumn{1}{c|}{\begin{tabular}[c]{@{}c@{}}3k\\ 100r\end{tabular}} & \multicolumn{1}{c|}{\begin{tabular}[c]{@{}c@{}}3k\\ 100r\end{tabular}} & \multicolumn{1}{c|}{\begin{tabular}[c]{@{}c@{}}25k\\ 1000r\end{tabular}} & -      \\ \hline
\end{tabular}
\end{table*}

\subsubsection{Models and datasets}
We implement all the FL training models with Tensorflow\cite{abadi2016tensorflow}. 
We use MNIST \cite{lecun1998gradient}, EMNIST\cite{cohen2017emnist} and CIFAR-10 datasets \cite{krizhevsky2009learning} to train a SVM model (loss function: $\frac{\lambda}{2}||\mathbf{w}||^2+\frac{1}{2}\max\{0;1-y_j\mathbf{w}^{\top}x_j\}^2, \lambda = 0.1$) and a 9-layer CNN model (two $5\times5\times32$ convolution layers, each followed by a $2\times2$ max pooling and a local response normalization layer, two fully connected layers ($z\times256$, $256\times10$, where z = 1568 for MNIST and EMNIST and z = 2048 for CIFAR-10) and a softmax output layer with 10 units). To evaluate our online algorithm \textbf{DYNAMITE}, we first adopt non-i.i.d. data distribution settings proposed in \cite{wang2019adaptive} to simulate data heterogeneity among clients for static dataset case. We further conduct extensive experiments on streaming data by transforming the static local dataset into dynamic data stream, where we constantly distribute data samples to FL clients from complete dataset with different data stream configurations and data arrival patterns (Section \ref{sec:stream_config}) during the whole FL training process.  To evaluate our (offline) algorithm {\bf CoOptFL}, we initialize its input parameters ($t_{ui}$, $M_i$, $\rho$, $c$, $\beta$, $\delta$, $p_i$, and $t_{ci}$) using the same estimation method as in our online algorithm {\bf DYNAMITE}. 
%We use MNIST and CIFAR10 datasets to train a convex SVM model and a non-convex CNN model respectively. To evaluate our algorithm {\bf CoOptFL}, we initialize its input parameters 
%$t_{ui}$, $M_i$, $\rho$, $c$, $\beta$, $\delta$,$p_i$, and $t_{ci}$ 
%by the estimates obtained from the same method in our online algorithm {\bf AdaCoOpt} but through pre-run tests rather than the online process. 

\subsubsection{Streaming Data configurations and arrival patterns}\label{sec:stream_config}
In this subsection, we introduce the design of our online streaming data configurations and data arrival patterns applied in our following experiments. We design two different data stream configurations (I.I.D. stream and Continuous stream) and three distinct data arrival pattern (Smooth, Burst, Random) to fully simulate the data dynamics in online FL training and demonstrate the adaptability of our online algorithm \textbf{DYNAMITE}.

{\em Two streaming data configurations (in terms of feature class).} \textbf{I.I.D. stream:} Clients will receive all classes of data samples in every interval (e.g. every 100 communication rounds; please see Table \ref{tab:param}) and the number of each class of the data is the same. \textbf{Continuous stream:} Every client receives the same single class of data samples during the training process at any given time, and the chosen class will gradually change over time, i.e., once the training data of the currently chosen class is processed, a new class will be uniformly at random chosen from the set of classes that are not chosen before for the training task. % will gradually change over time by sampling uniformly at random from the classes that clients haven't used for training. %\carlee{this still isn't clear to me. Does each client only receive data points from one class, and then this gradually changes to another class where the other class is chosen uniformly at random?}
%\carlee{how are they changed? Do you sample uniformly at random from classes this client hasn't seen before? How do you determine the number of classes sent to each client? What is a ``duplicated'' class?}. 
%Detailed setting of the interval and the number of arriving data samples are shown in Table \ref{tab:param} (Arrival rate).

{\em Three data arrival patterns (in terms of the number of data samples).} 
\textbf{Smooth arrival:} Clients will receive the same number of data samples at each regular interval. \textbf{Burst arrival:} Each client will receive a massive amount of data samples in a specific round %\xiaoxi{(explicitly show how short the period of time is, or explicitly say how much data is received in the xxxth round)} 
after the training starts, and few samples are received in other rounds. \textbf{Random-arrival:} The quantity and the arrival time of data samples of each client are uncertain; the arrival patterns are time-varying and heterogeneous across clients. 

Parameters used in this section are clarified in Section \ref{sec:exp_param} and shown in Table \ref{tab:param}. %\xiaoxi{But why are the numbers in the row of Time(s) all very large? 1k, 500, 800. Are they the numbers of data samples arrived per second? Or is Time(s) referred to as ``interval''? But why are they so large?}

%\carlee{does the test data distribution change over time? Or just the training data at each client?}

\begin{figure}
	\centering
	\begin{subfigure}[t]{0.48\linewidth}
		\includegraphics[width=\textwidth]{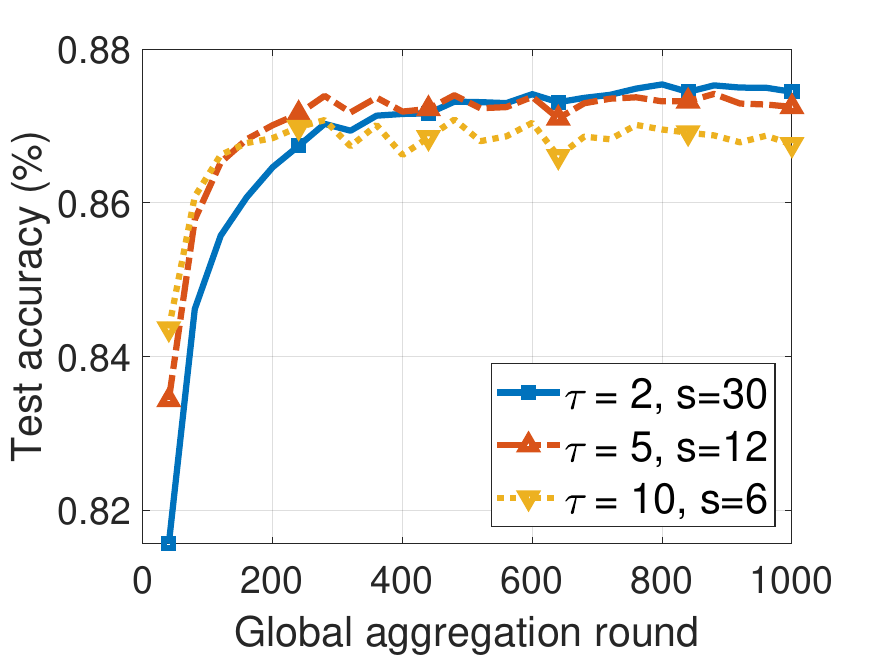}
		\caption{Test Accuracy}
		\label{fig:case1_acc}
	\end{subfigure}
	\begin{subfigure}[t]{0.48\linewidth}
		\includegraphics[width=\textwidth]{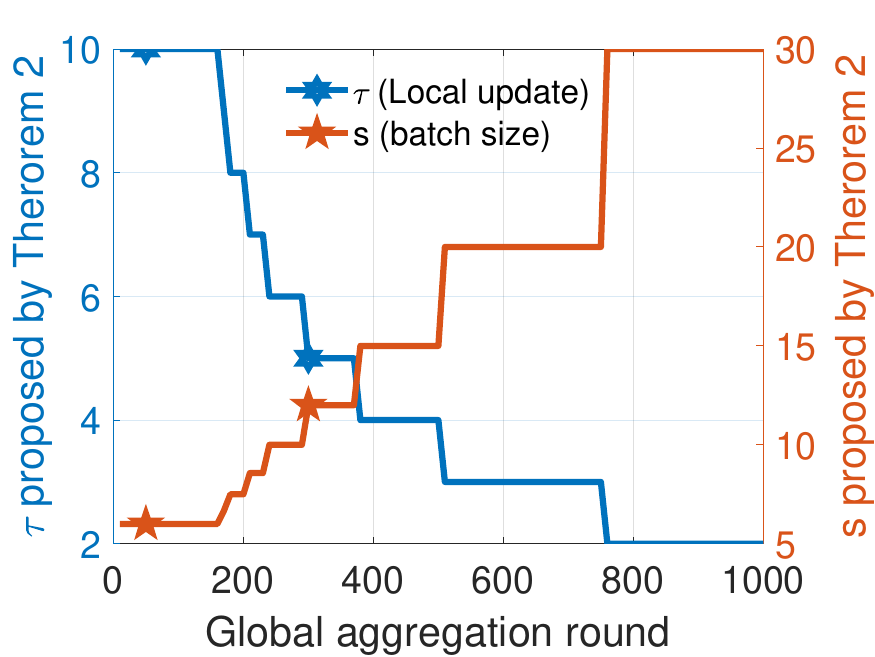}
		\caption{$\tau$, $s$ proposed in Theorem 2}
		\label{fig:case1_optau}
	\end{subfigure}
	\caption{Optimal $\tau$ and $s$ (squared-SVM, MNIST)}
	\label{fig:case1}
\end{figure}

\subsubsection{Baselines}
To demonstrate the effectiveness of our carefully chosen batch size configurations for different clients using {\bf CoOptFL}, we compare with {\bf Uniform}, a widely-adopted method with uniform batch size~\cite{wang2019adaptive} for all the clients, and with {\bf No-straggler}, a time-efficient batch size strategy proposed by \cite{ma2021adaptive}, and with \textbf{DBFL}, a dynamic batch size selection scheme proposed in \cite{shi2022talk}.

To evaluate the performance of our online algorithm {\bf DYNAMITE} under imperfect estimation of the model and system parameters, we compare with {\bf FedAvg}, which maintains $\tau$ and batch size unchanged, %and {remains unchanged through out the training, 
an adaptive aggregation control algorithm {\bf Dynamic-$\tau$} proposed by \cite{wang2019adaptive}, and the time-efficient {\bf No-straggler} algorithm in~\cite{ma2021adaptive}. 

To verify the impact of the data sampling methods in FL training with dynamic data stream, we also compare the performance of our adopted \textbf{Reservoir Sampling} (Algorithm \ref{algo_reservoir}) with other general sampling methods like \textbf{Random Sampling}, a straight-forward sampling strategy which uniformly at random discards a stored data sample and replace it with the latest one, and with \textbf{FIFO Sampling}, a classic data selection strategy to update clients' buffer simply following the ``First-in First-out'' principle.

\begin{figure}
	\centering
	\begin{subfigure}[t]{0.48\linewidth}
		\includegraphics[width=\textwidth]{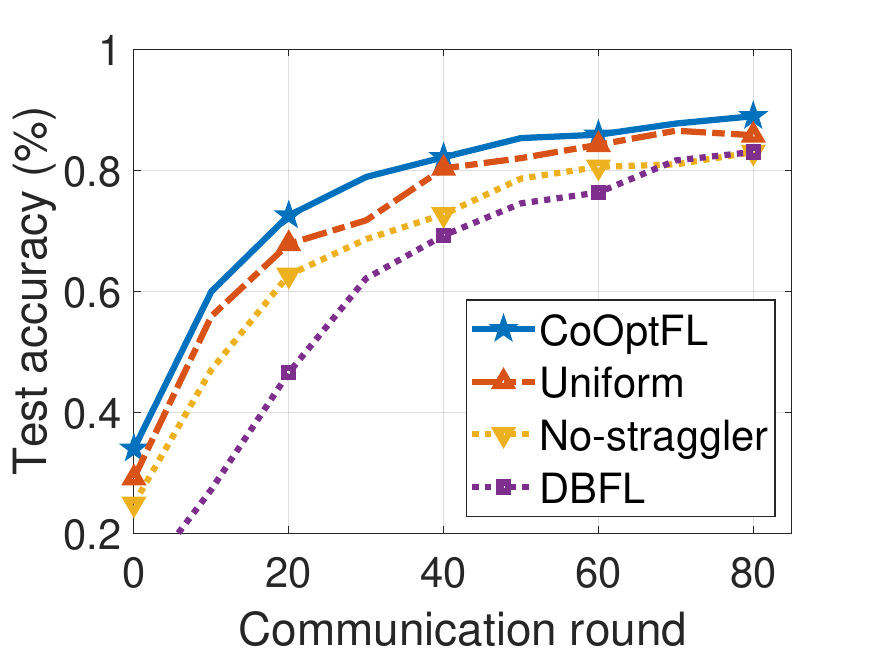}
		\caption{Accuracy (MNIST) 5-client}
	\end{subfigure}
	\begin{subfigure}[t]{0.48\linewidth}
		\includegraphics[width=\textwidth]{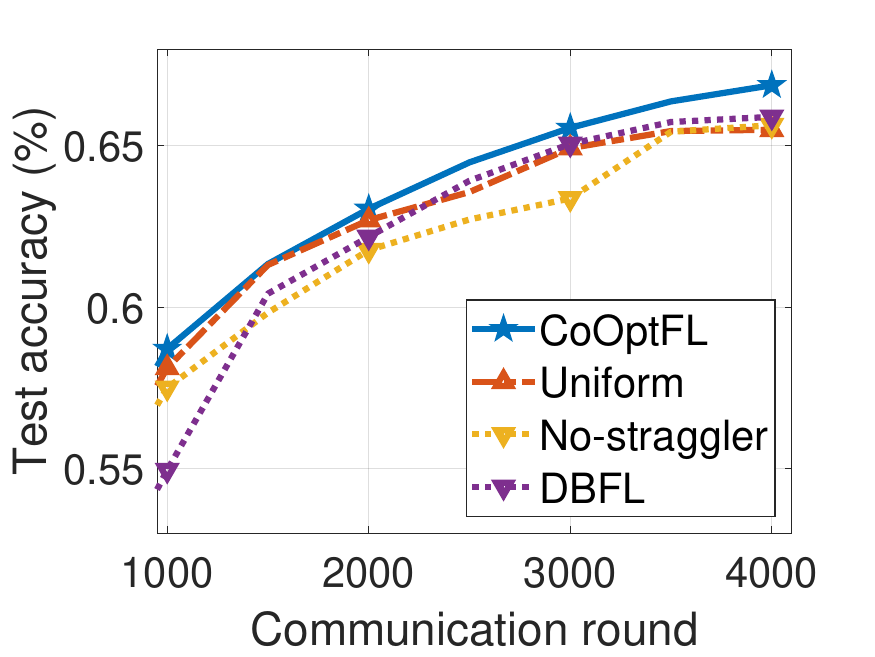}
		\caption{Accuracy (CIFAR) 20-client}
	\end{subfigure}
	\caption{Our batch size assignment in offline algorithm {\bf CoOptFL} achieves the highest accuracy for both datasets}
	\label{fig:bs-dis}
\end{figure}

\begin{figure*}
    \centering
	\begin{subfigure}[t]{0.24\linewidth}
		\includegraphics[width=\textwidth]{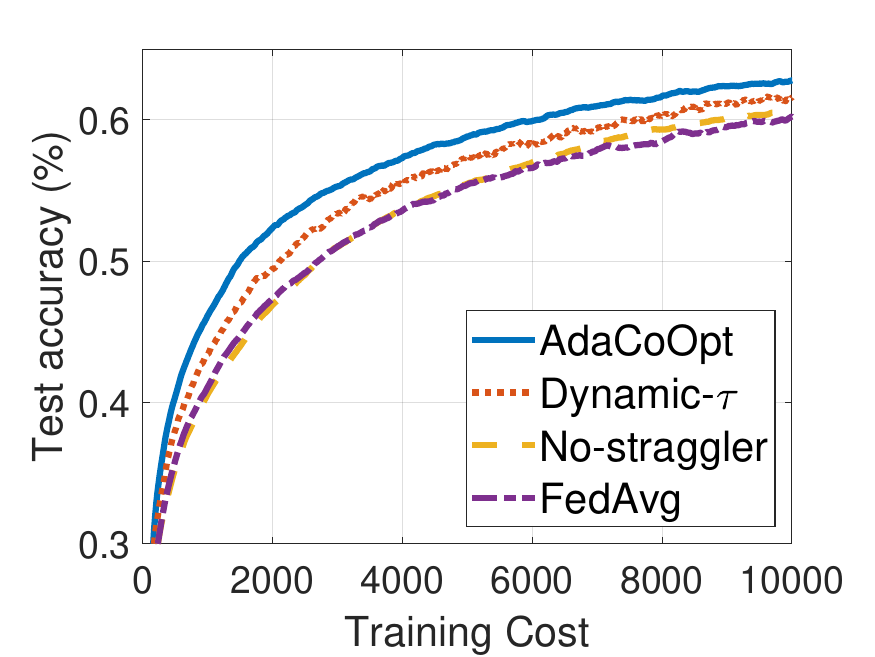}
		\caption{Cost-dominant (20-client)\\$R=10000, \theta=5000s$}
		\label{fig:online_resource_acc}
	\end{subfigure}
	\begin{subfigure}[t]{0.24\linewidth}
		\includegraphics[width=\textwidth]{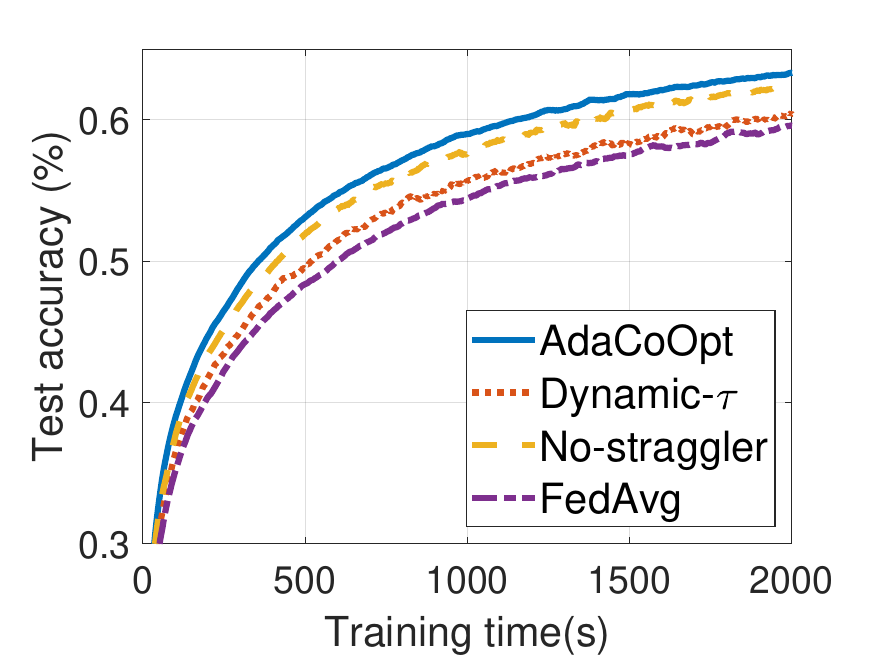}
		\caption{Time-dominant (20-client)\\$R=20000, \theta=2000s$}
	\end{subfigure}
	\begin{subfigure}[t]{0.24\linewidth}
		\includegraphics[width=\textwidth]{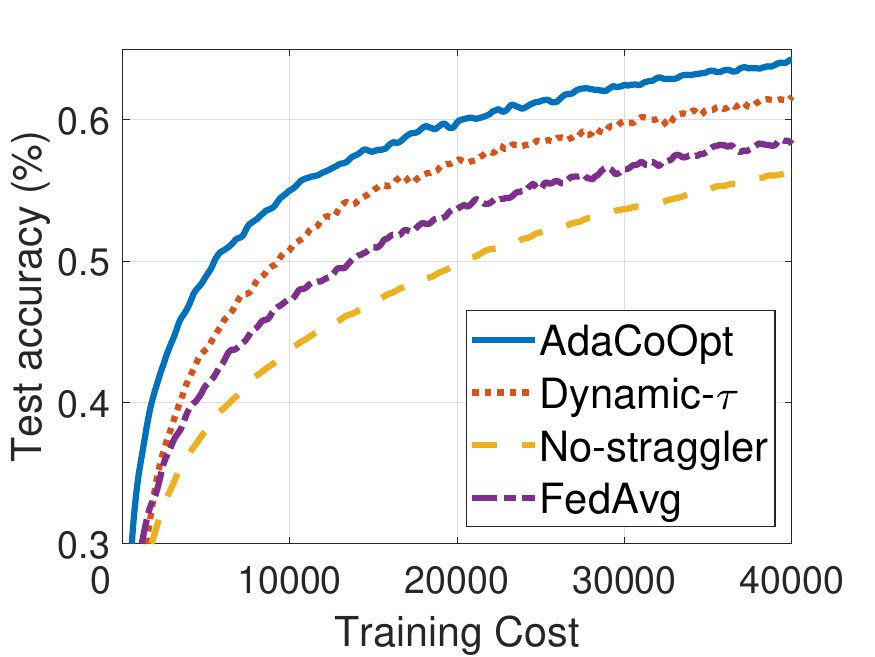}
		\caption{Cost-dominant (100-client)\\$R=40000, \theta=5000s$}
		\label{fig:online_similar_1}
	\end{subfigure}
	\begin{subfigure}[t]{0.24\linewidth}
		\includegraphics[width=\textwidth]{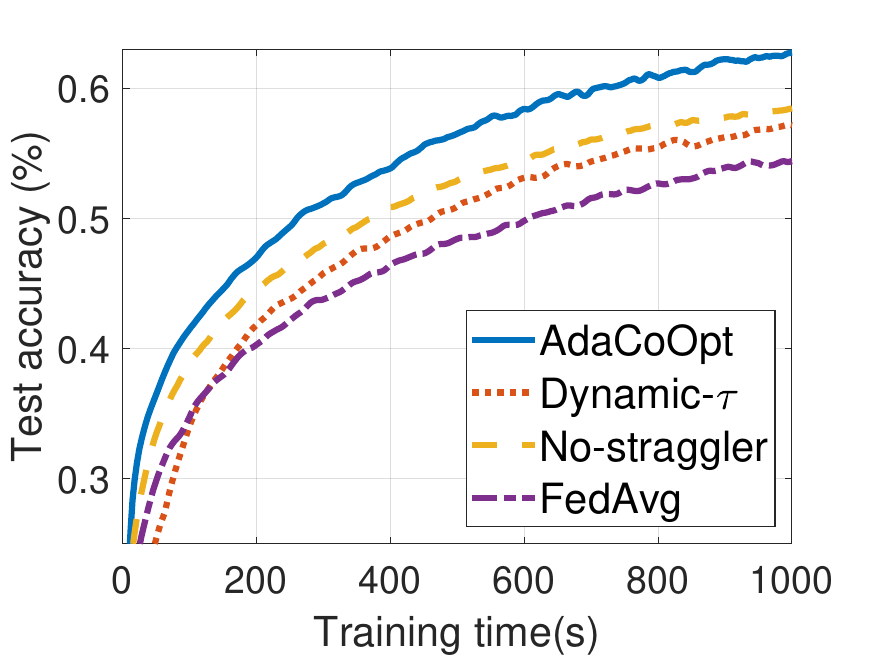}
		\caption{Time-dominant (100-client)\\$R=80000, \theta=1000s$}
	\end{subfigure}
	\caption{Our online algorithm {\bf DYNAMITE} achieves the highest accuracy under CIFAR-10 in both cost-sensitive and time-sensitive scenarios} 
	\label{fig:online-line}
\end{figure*}

\subsubsection{Parameters and run-time traces of the FL training.}
\label{sec:exp_param}
In our experiments, we initialize models with $\mathbf{w(0)}$ and set the default local update step $\tau = 2$, mini-batch size $s = 60$, and step size $\eta=0.005$ unless otherwise specified. %We also adopt the learning rate scaling rule applied in \cite{ma2021adaptive,balles2016coupling} to adjust the step-size among different clients according to their assigned batch sizes in the adaptive control algorithm. 
To evaluate the training cost and the completion time, we set the computation cost per sample $a=0.0005$ and the communication cost per round $b = |\mathcal{N}|/10$. On each of our testbeds, the clients are training the same FL model, but they have different resource configurations and run-times. In particular, the run-time logs of the 5-client and 20-client experiments are real; but in the 100-client simulation, we sample the run-time of each client from the run-time trace collected at the 20 VM instances located in different edge clusters to simulate the real-world communication and computation overhead. Other important parameters used in the experiments are presented in Table \ref{tab:param}. 

We clarify that Smooth-I and Smooth-C denote I.I.D stream and Continuous stream configurations under smooth arrival pattern, respectively. 
%``time-c'' and ``cost-c'' are the abbreviations of 'time-constrained' and 'cost-constrained', representing two different scenarios of our optimization problem. 
The arrival rate denotes the number of samples that clients received in every 100 round. %\xiaoxi{use training iteration or round or specify the length of the time interval, e.g., 1 second.} 
(Smooth arrival) or in a specific round (Burst arrival, EMNIST:Round 500, CIFAR: Round 1000).
%We use actual measured time to evaluate the computation and communication time in the test-bed experiment for cloud instances. Further, we will sample these actual measured time data for a more realistic time evaluation in the simulation experiments. 

\begin{figure}
    \centering
	\begin{subfigure}[t]{0.48\linewidth}
		\includegraphics[width=\textwidth]{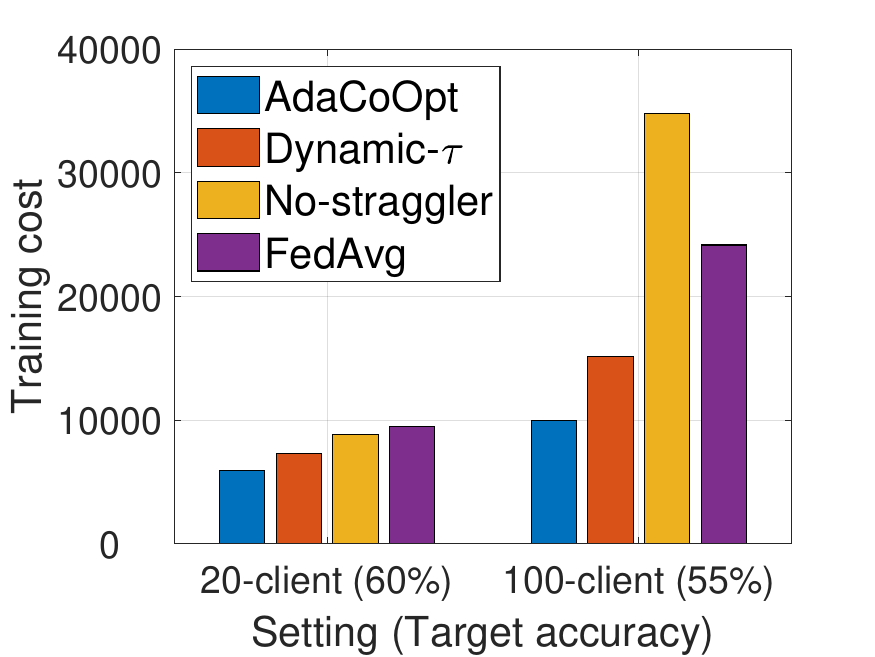}
		\caption{Cost-dominant scenario}
		\label{fig:online_similar_2}
	\end{subfigure}
	\begin{subfigure}[t]{0.48\linewidth}
		\includegraphics[width=\textwidth]{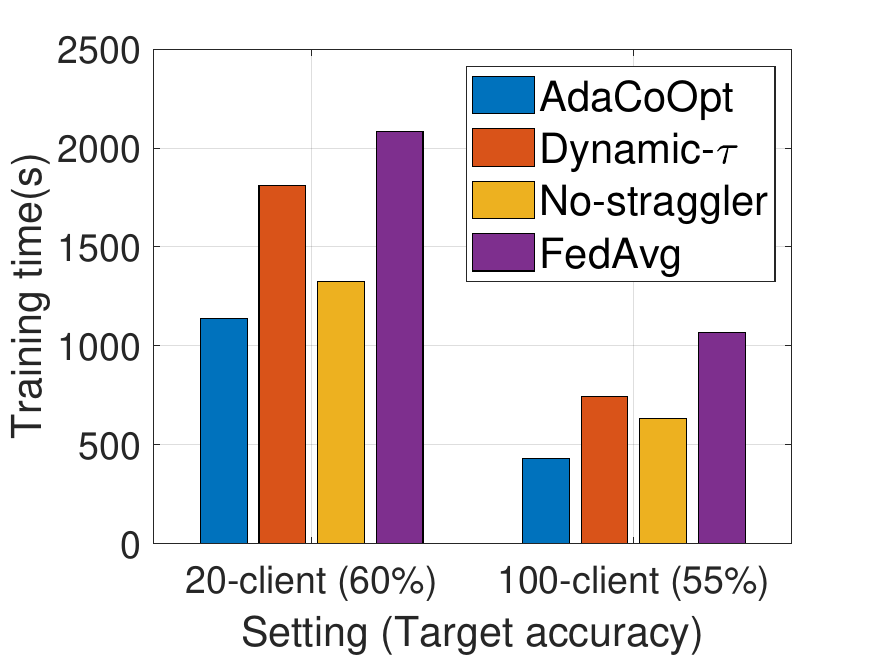}
		\caption{Time-dominant scenario}
		\label{vkt}
	\end{subfigure}
    \caption{Our online algorithm {\bf DYNAMITE} consumes minimal cost and time to achieve the target accuracy under CIFAR-10 in both cost-sensitive and time-sensitive scenarios} 
    \label{fig:online-bar}
\end{figure}

\subsection{Experimental results and interpretation}
\label{ssec:exp_results}

\subsubsection{\bf Optimal number of local updates per round $\tau$ and uniform batch size $s$}
We find the optimal combination $\tau^*$ and $s^*$ using our Theorem \ref{thm:opt_uniform} for a squared-SVM model training and testing under the MNIST dataset. We compare three different combinations: $\tau = 10, s=6$; $\tau = 5, s=12$; $\tau = 2, s=30$. 
 %where $(\tau = 5, s=12)$ is the optimum based on Theorem \ref{thm:opt_uniform} with parameters estimated through pre-run tests using the same estimation as our Algorithms \ref{algo_disjdecomp} and \ref{alg:online_client}. 
%
%
%~\xiaoxi{Use global aggregation round as xlabel}
Fig. \ref{fig:case1_acc} shows the optimal $\tau$ and $s$ combination varying $K$, e.g., $(\tau=10, s=6)$ for $K<50$, $(\tau=5, s=12)$ for $K\approx 300$, and $(\tau =2, s=30)$ for $K>800$ achieve the highest accuracy respectively. We also mark the optimal $\tau$ and batch size proposed in our Theorem \ref{thm:opt_uniform}  at the corresponding rounds in Fig. \ref{fig:case1_optau} for better visualization. Besides, Fig. \ref{fig:case1_optau} shows that the optimal $\tau$ decreases with the increase of $K$, supporting our theoretical result in Remark \ref{rmk:opt_tau_1}. %Moreover, Fig. \ref{fig:case1_optau} shows that $\tau$ proposed in our Theorem \ref{thm:opt_uniform} can well capture the varying optimal combination mentioned above.

\subsubsection{\bf Optimal heterogeneous batch sizes across clients}
%We use MNIST and CIFAR-10 datasets with $K=1000$ aggregation rounds and $\tau = 5$ local updates per round. 
We compare our offline algorithm {\bf CoOptFL} to {\bf No-straggler}\cite{ma2021adaptive}, which configures batch sizes $s_i$ according to clients' computing capacities ($s_i \propto p_i$) so as to eliminate the straggler effect across clients, {\bf Uniform} ($s=60$) and {\bf DBFL} (Initial incremental factor $= 1.1$) . We set the communication round $K=80$ and $K=3000$ for the MNIST and CIFAR-10 datasets, respectively. For fairness, we set the total batch size $s_{tot}=\sum_{i\in\mathcal{N}}{s_{i}}$ as a constant to ensure that clients will process the same amount of data samples $K\cdot s_{tot}$ in total while using different batch size assignment strategies.

Fig. \ref{fig:bs-dis} shows that {\bf CoOptFL} can converge faster and achieve better final testing accuracy compared to the baselines in both 5-client and 20-client settings. Note that {\bf No-straggler} always tends to assign bigger batch sizes to devices with higher computing capacities regardless of their non-i.i.d. data properties, which leads to a lower model accuracy and resource utilization than {\bf Uniform}, especially when devices with higher computing capacity have fewer and similar data samples (Theorem \ref{thm:opt_alg_1}). Similar results can be found in Fig.~\ref{fig:online_similar_1} and Fig.~\ref{fig:online_similar_2} in the following online experiments as well. The slower convergence speed of \textbf{DBFL} can be attributed to the smaller initial batch size configurations.  %\carlee{It would be interesting to show the batch sizes and how/whether they depend on $D_i$ and $M_i$} \xiaoxi{We find it hard to show it due to the space limit.}

\subsubsection{\bf Adaptive control for co-optimized aggregation frequency and heterogeneous batch sizes (static datasets)}
We first further compare our {\bf DYNAMITE} with three benchmarks for CIFAR-10 FL training using static dataset: the first two are vanilla FedAvg~\cite{li2019} and the time-efficient {\bf No-straggler} \cite{ma2021adaptive}; the third one is {\bf Dynamic-$\tau$} \cite{wang2019adaptive} which dynamically adjusts $\tau_k$ for each round $k$. %We use the same hyperparameters for baselines according to their original papers.% but uses a static batch size $s=50$. %~\xiaoxi{specify it}.
%We conduct this experiment on CIFAR10 dataset.
%(10000:10000:30000,ts=120)
%
\begin{comment}
\begin{figure}
    \label{fig:adaptive_alg_resource}
	\centering
	\begin{subfigure}[t]{0.48\linewidth}
		\includegraphics[width=\textwidth]{figure/online_resource_acc.png}
		\caption{Model accuracy}
		\label{fig:online_resource_acc}
	\end{subfigure}
	\begin{subfigure}[t]{0.48\linewidth}
		\includegraphics[width=\textwidth]{figure/online_resource_loss.png}
		\caption{Training loss}
		\label{fig:online_resource_loss}
	\end{subfigure}
	\caption{Adaptive algorithm (Resource Dominant)}	
\end{figure}
%
\begin{figure}
	\centering
	\begin{subfigure}[t]{0.45\linewidth}
		\includegraphics[width=\textwidth]{figure/online_time_acc.png}
		\caption{Accuracy}
	\end{subfigure}
	\begin{subfigure}[t]{0.45\linewidth}
		\includegraphics[width=\textwidth]{figure/online_time_loss.png}
		\caption{Training loss}
	\end{subfigure}
	\caption{Adaptive algorithm (Time Dominant)}	
	\label{fig:adaptive_alg_time}
\end{figure}
\end{comment}
%---------------------
%--------------------------
%
%
%We compared our adaptive control algorithm with the one in \cite{wang2019adaptive} and the static algorithm using both uniform and no-straggler batch size distribution. 

\begin{figure}[t]
	\centering
	\begin{subfigure}[t]{0.45\linewidth}
		\includegraphics[width=\textwidth]{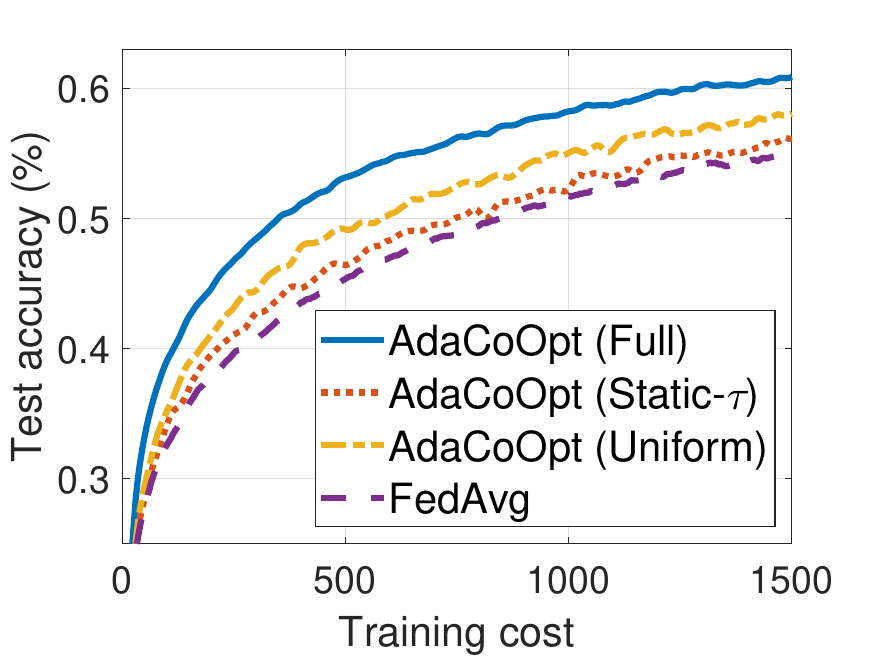}
		\caption{Cost-dominant scenario}
		\label{fig:ablation_cost}
	\end{subfigure}
	\begin{subfigure}[t]{0.45\linewidth}
		\includegraphics[width=\textwidth]{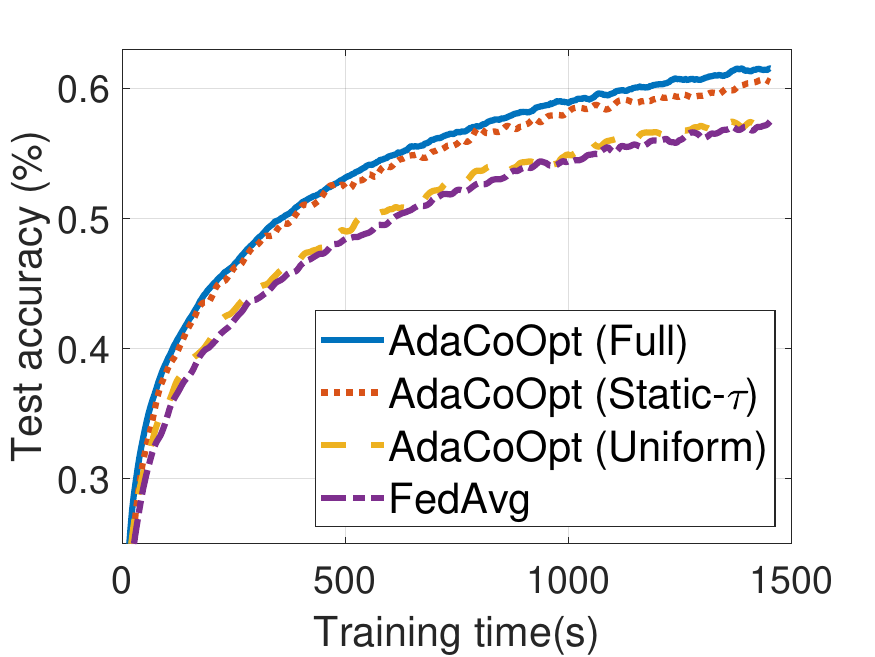}
		\caption{Time-dominant scenario}
		\label{fig:ablation_time}
	\end{subfigure}
	\caption{Ablation Experiment under CIFAR-10}
	\label{fig:ablation}
\end{figure}

\begin{figure*}[t]
    \centering
	\begin{subfigure}[t]{0.24\linewidth}
		\includegraphics[width=\textwidth]{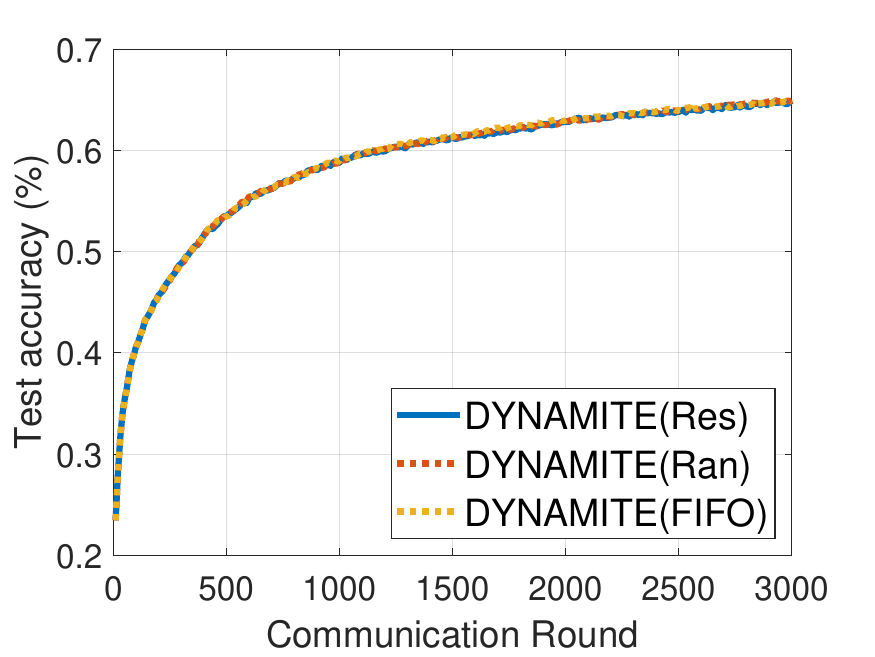}
		\caption{Smooth-IID}
		\label{fig:sampling-cifar-smoothi}
	\end{subfigure}
	\begin{subfigure}[t]{0.24\linewidth}
		\includegraphics[width=\textwidth]{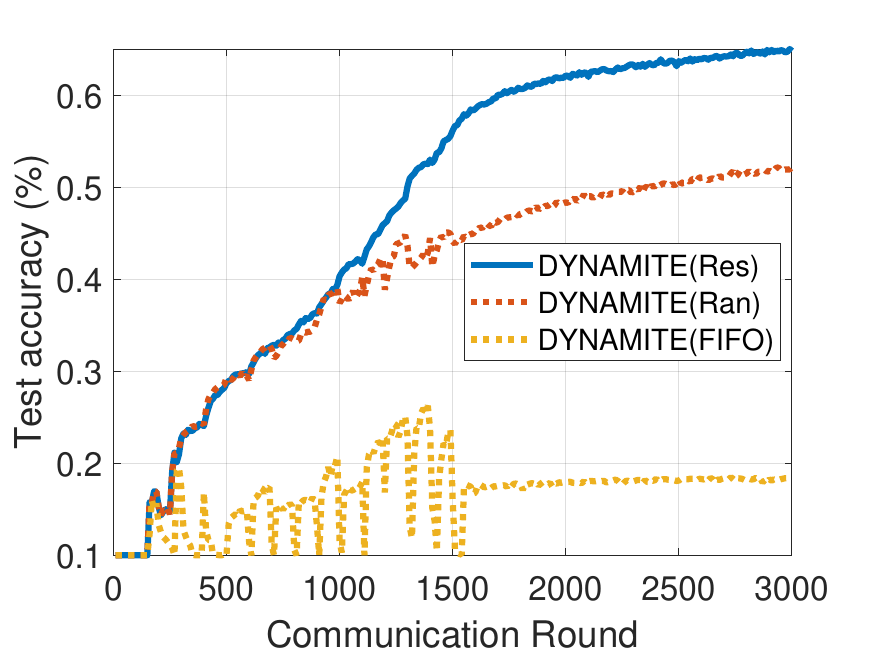}
		\caption{Smooth-Continuous}
		\label{fig:sampling-cifar-smoothc}
	\end{subfigure}
	\begin{subfigure}[t]{0.24\linewidth}
		\includegraphics[width=\textwidth]{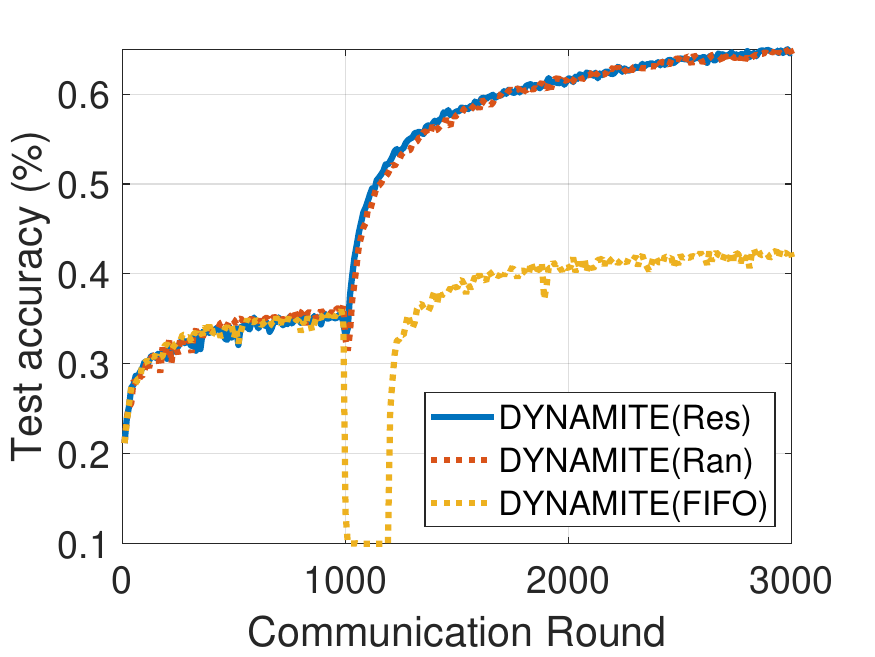}
		\caption{Burst-Continuous}
		\label{fig:sampling-cifar-burst}
	\end{subfigure}
	\begin{subfigure}[t]{0.24\linewidth}
		\includegraphics[width=\textwidth]{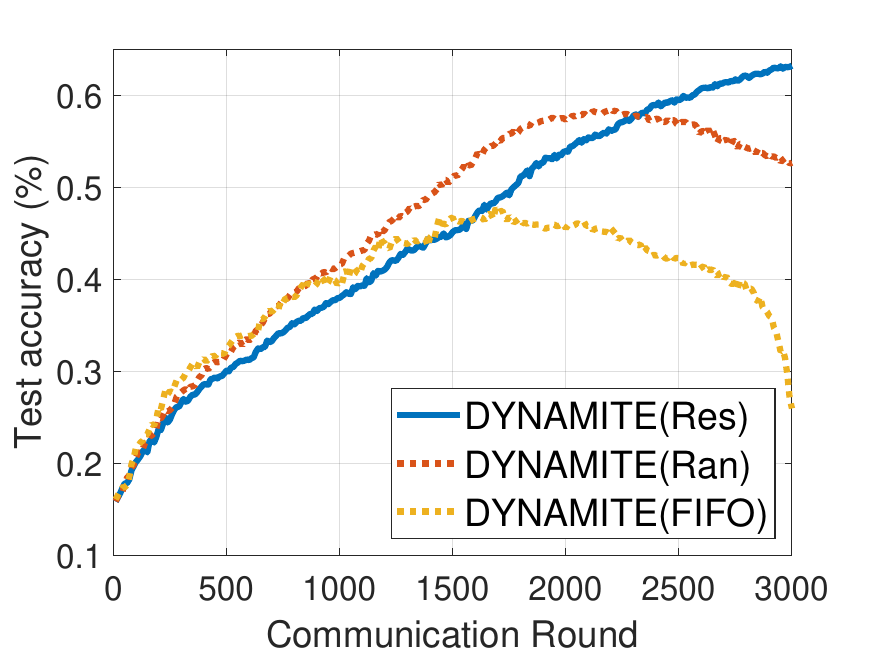}
		\caption{Random-Continuous}
		\label{fig:sampling-cifar-random}
	\end{subfigure}
	\caption{Sampling methods under CIFAR-10 (DYNAMITE)} 
	\label{fig:sampling-cifar}
\end{figure*}

\begin{figure*}[t]
    \centering
	\begin{subfigure}[t]{0.24\linewidth}
		\includegraphics[width=\textwidth]{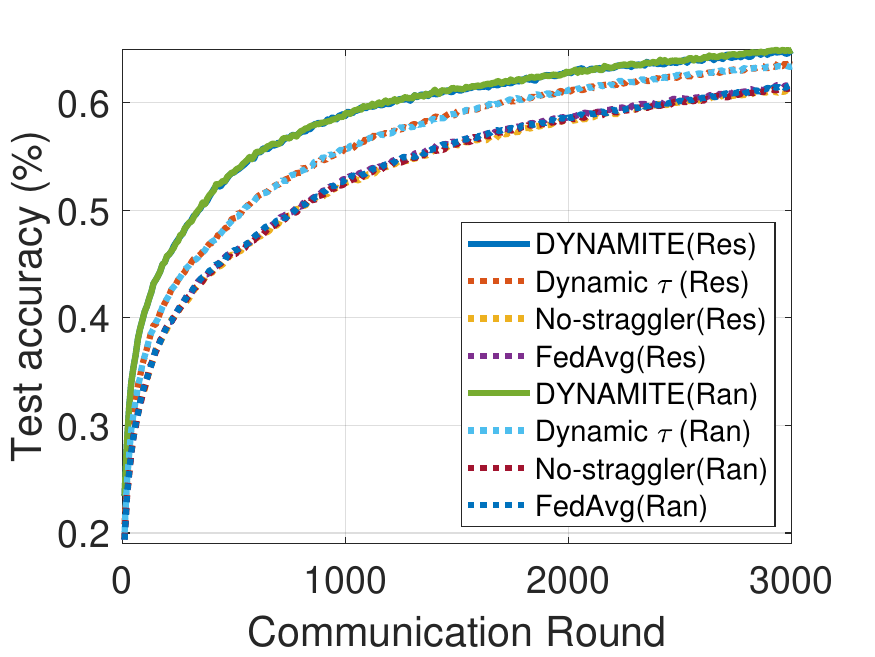}
		\caption{Smooth-IID}
		\label{fig:cifar-smoothi}
	\end{subfigure}
	\begin{subfigure}[t]{0.24\linewidth}
		\includegraphics[width=\textwidth]{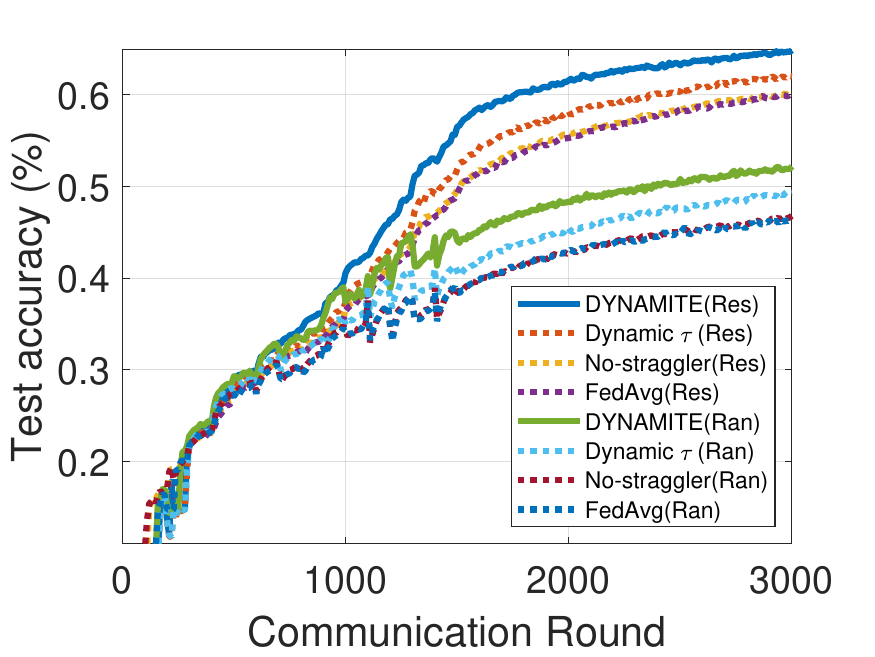}
		\caption{Smooth-Continuous}
		\label{fig:cifar-smoothc}
	\end{subfigure}
	\begin{subfigure}[t]{0.24\linewidth}
		\includegraphics[width=\textwidth]{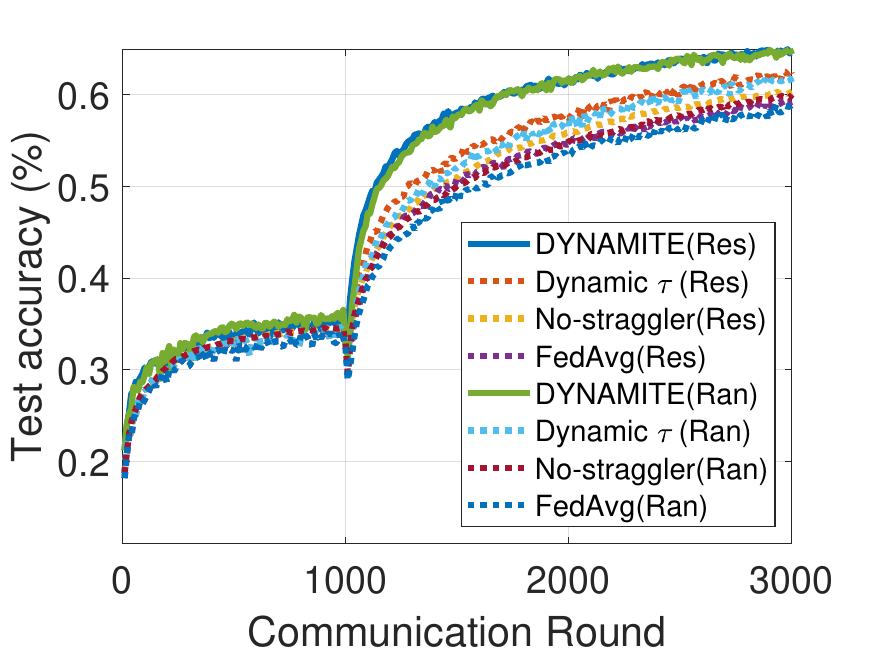}
		\caption{Burst-Continuous}
		\label{fig:cifar-burst}
	\end{subfigure}
	\begin{subfigure}[t]{0.24\linewidth}
		\includegraphics[width=\textwidth]{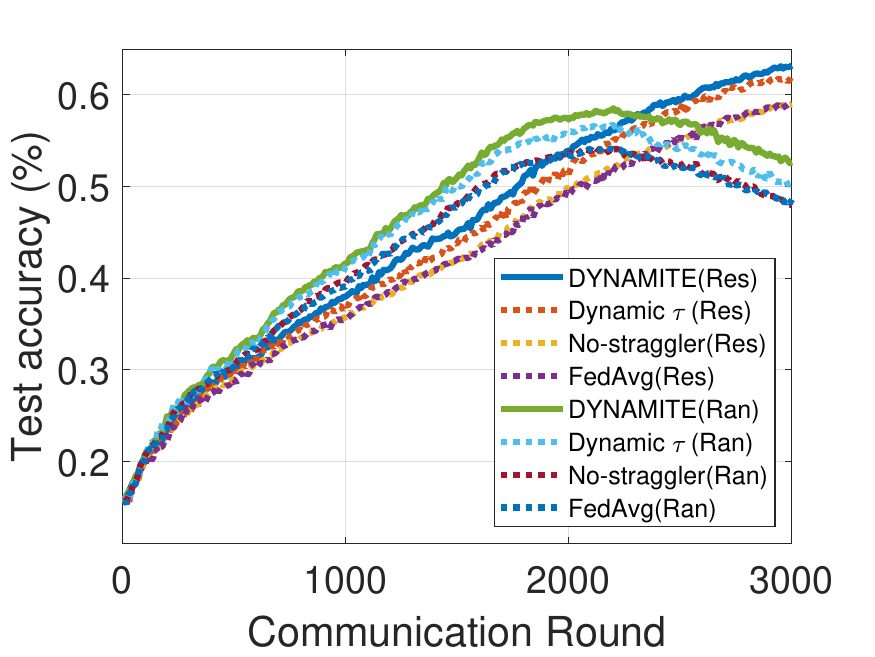}
		\caption{Random-Continuous}
		\label{fig:cifar-random}
	\end{subfigure}
	\caption{Sampling methods under CIFAR-10 (Extensive)} 
	\label{fig:stream-cifar}
\end{figure*}

\begin{figure}[t]
	\centering
		\begin{subfigure}[t]{0.45\linewidth}
		\includegraphics[width=\textwidth]{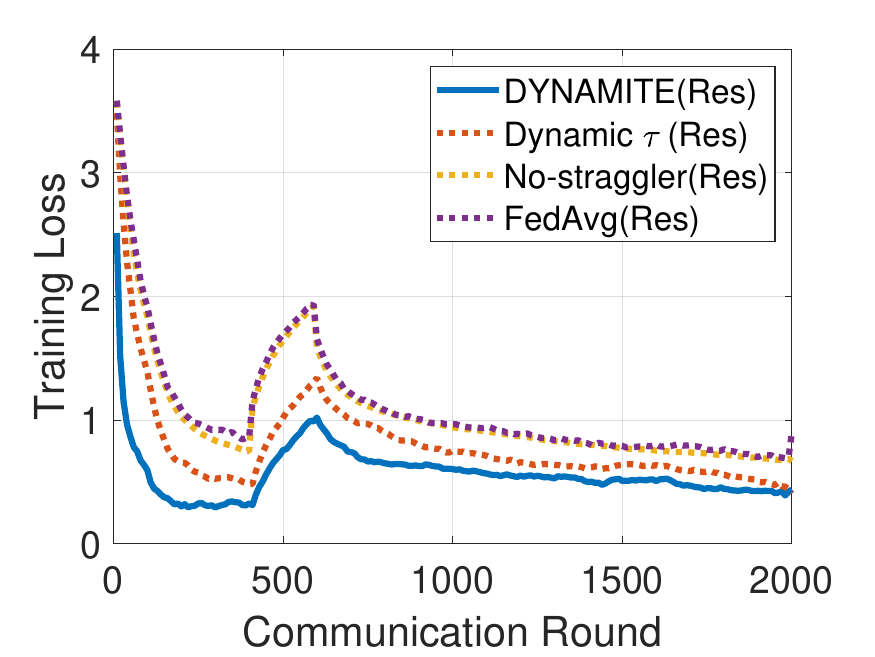}
		\caption{EMNIST}
	\end{subfigure}
	\begin{subfigure}[t]{0.45\linewidth}
		\includegraphics[width=\textwidth]{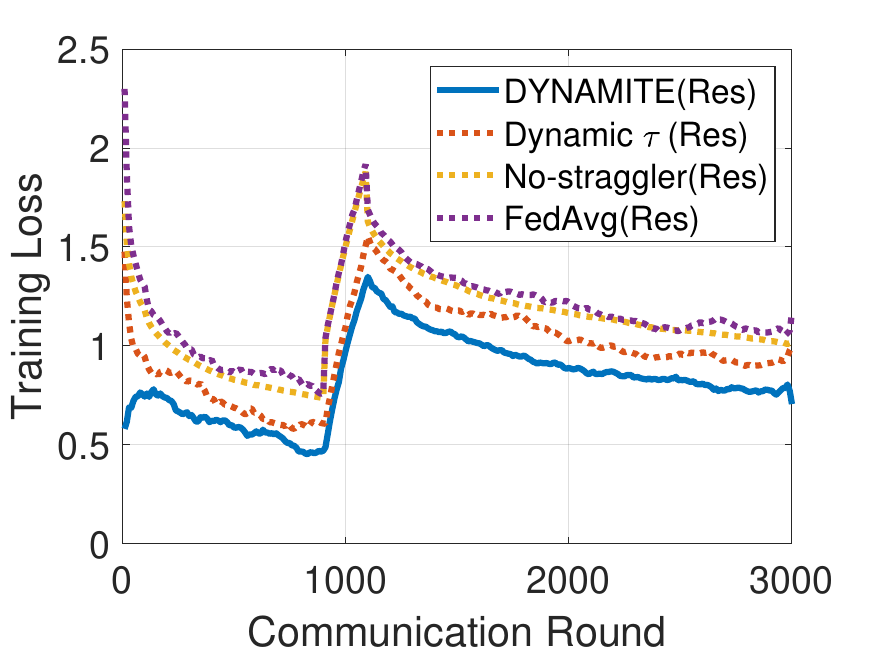}
		\caption{CIFAR-10}
	\end{subfigure}
	\caption{Training loss (Burst arrival)}
	\label{fig:train_loss_burst}
	
\end{figure}

We compare the strategies in two different scenarios of our optimization problem, where the cost constraint and time constraint dominates, respectively. We set different values of $R$ and $\theta$ to simulate these two different scenarios. Fig.\ref{fig:online-line} and Fig.\ref{fig:online-bar} together show that {\bf DYNAMITE} can outperform the baselines in both scenarios under different settings. For instance, in the cost-dominant scenario, {\bf DYNAMITE} can achieve a 2.7\%--7.9\% higher final test accuracy than {\bf FedAvg} and reduce the cost by 37.6\%--58\% when achieving the same accuracy. It achieves a 3.8\%--8.4\% higher final test accuracy and 45.4\%--59.6\% less completion time if achieving the same accuracy in the time-dominant scenario. These results indicate great adaptability of {\bf DYNAMITE}. %\weijie{Add comparison description between Dynamic tau and No-strag if needed}

Moreover, we conduct ablation experiments on both 20-client and 100-client settings to test the value of co-optimizing $\tau$ and $s_i$ of our {\bf DYNAMITE} in both cost-dominant and time-dominant scenarios. We compare our {\bf DYNAMITE} with {\bf DYNAMITE (Static-$\tau$)}, which only optimizes the batch sizes using a fixed aggregation frequency and {\bf DYNAMITE (Uniform)}, which uses a uniform batch size among clients, only adjusting the local update steps adaptively. Fig. \ref{fig:ablation_cost} shows that a timely adjusted global aggregation frequency ({\bf DYNAMITE (Uniform)}) can effectively reduce the training(communication) cost and thus is more critical in a cost-dominant training scenario. On the other hand, Fig. \ref{fig:ablation_time} shows that a careful batch size assignment ({\bf DYNAMITE (Static-$\tau$)}) can well capture the system and data heterogeneity
%and balance the trade-off between training time and model convergence 
so as to achieve a better model accuracy in a time-dominant scenario. These results also match our experiments in Fig. \ref{fig:online-line}, where {\bf Dynamic-$\tau$} performs better than {\bf No-straggler} in the cost-sensitive scenario, but worse than {\bf No-straggler} in the time-sensitive scenario.

\begin{table*}[]
\caption{Test accuracy in online FL training (Reservoir Sampling)}
\label{tab:stream_acc}
\begin{tabular}{cccccccccc}
\hline
\multicolumn{2}{c}{Dataset}                             & \multicolumn{4}{c}{EMNIST}                                        & \multicolumn{4}{c}{CIFAR}                                         \\ \hline
\multicolumn{2}{c}{Configuration}                  & Smooth-I     & Smooth-C       & Burst          & Random         & Smooth-I       & Smooth-C       & Burst          & Random         \\ \hline
\multirow{4}{*}{\begin{tabular}[c]{@{}c@{}}Cost\\ Constrained\end{tabular}} 
                                & FedAvg          & 0.785          & 0.761          & 0.774          & 0.764          & 0.610          & 0.599          & 0.592          & 0.59  \\
                                & No-straggler    & 0.785          & 0.763          & 0.780          & 0.768          & 0.618          & 0.601          & 0.603          & 0.593  \\
                                & Dynamic $\tau$  & 0.817          & 0.791          & 0.800          & 0.794          & 0.636          & 0.620          & 0.627          & 0.622  \\
                                & DYNAMITE        & \textbf{0.824} & \textbf{0.810} & \textbf{0.824} & \textbf{0.811} & \textbf{0.649} & \textbf{0.649} & \textbf{0.650} & \textbf{0.632} \\ \hline
\multirow{4}{*}{\begin{tabular}[c]{@{}c@{}}Time\\ Constrained\end{tabular}} 
                                & FedAvg          & 0.699          & 0.580          & 0.724          & 0.634          & 0.539          & 0.469          & 0.530          & 0.400          \\
                                & No-strag        & 0.776          & 0.662          & 0.750          & 0.701          & 0.608          & 0.534          & 0.598          & 0.456          \\
                                & Dynamic $\tau$  & 0.734          & 0.556          & 0.735          & 0.641          & 0.557          & 0.444          & 0.552          & 0.410          \\
                                & DYNAMITE        & \textbf{0.787} & \textbf{0.793} & \textbf{0.817} & \textbf{0.798} & \textbf{0.618} & \textbf{0.629} & \textbf{0.631} & \textbf{0.572} \\ \hline
\end{tabular}
\end{table*}

\begin{figure*}[t]
    \centering
	\begin{subfigure}[t]{0.24\linewidth}
		\includegraphics[width=\textwidth]{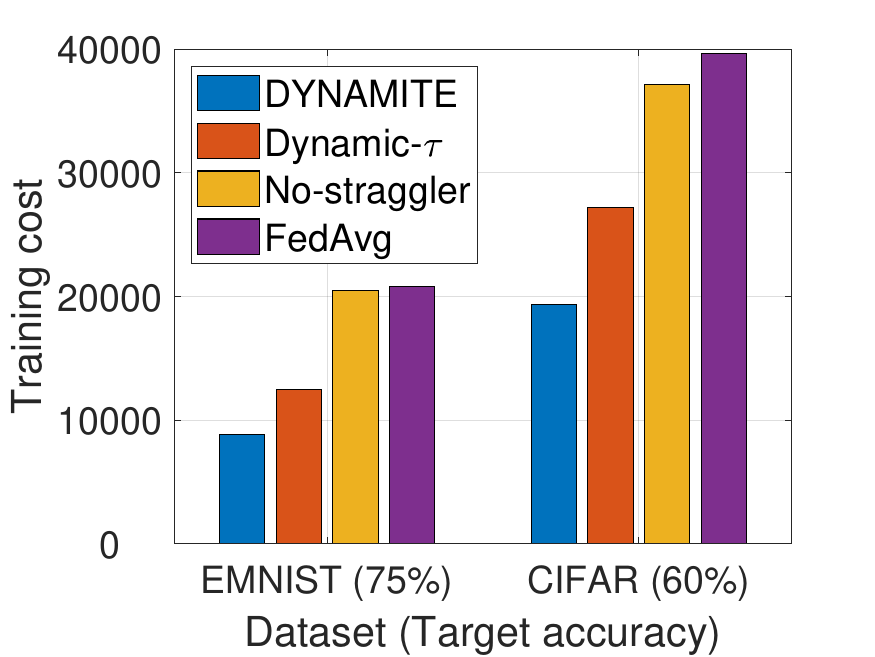}
		\caption{Smooth-IID}
		
	\end{subfigure}
	\begin{subfigure}[t]{0.24\linewidth}
		\includegraphics[width=\textwidth]{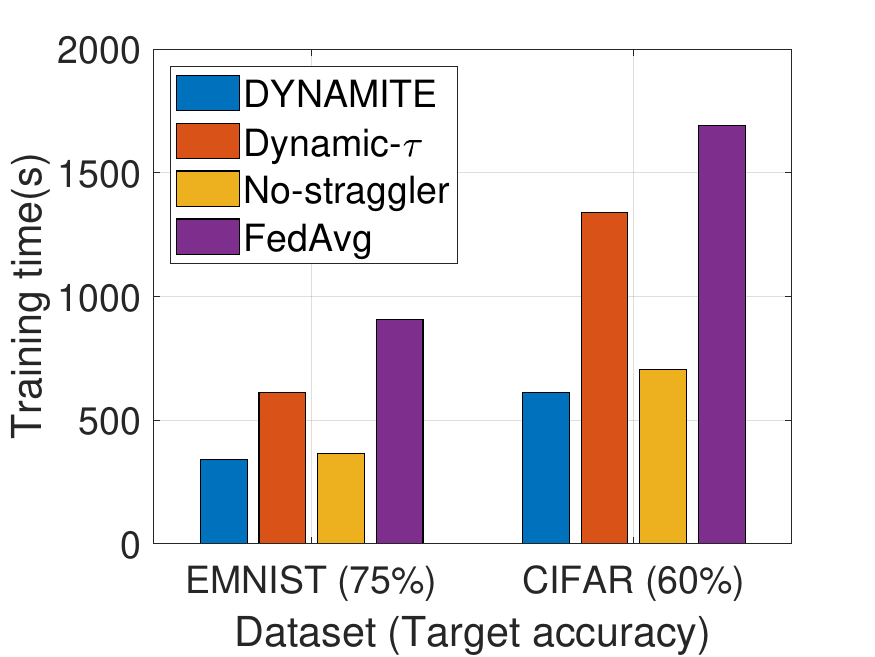}
		\caption{Smooth-IID}
		
	\end{subfigure}
	\begin{subfigure}[t]{0.24\linewidth}
		\includegraphics[width=\textwidth]{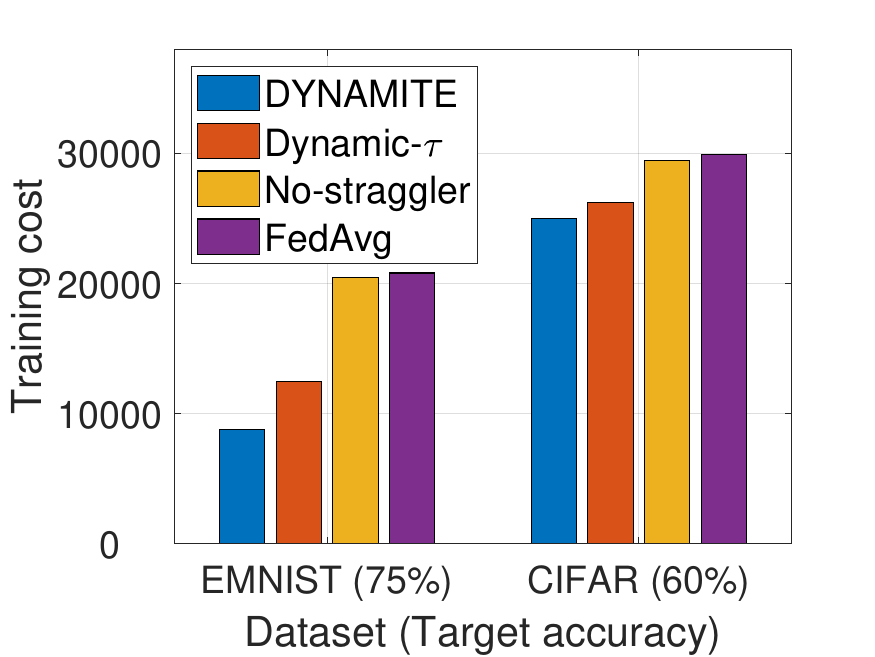}
		\caption{Smooth-Continuous}
		
	\end{subfigure}
	\begin{subfigure}[t]{0.24\linewidth}
		\includegraphics[width=\textwidth]{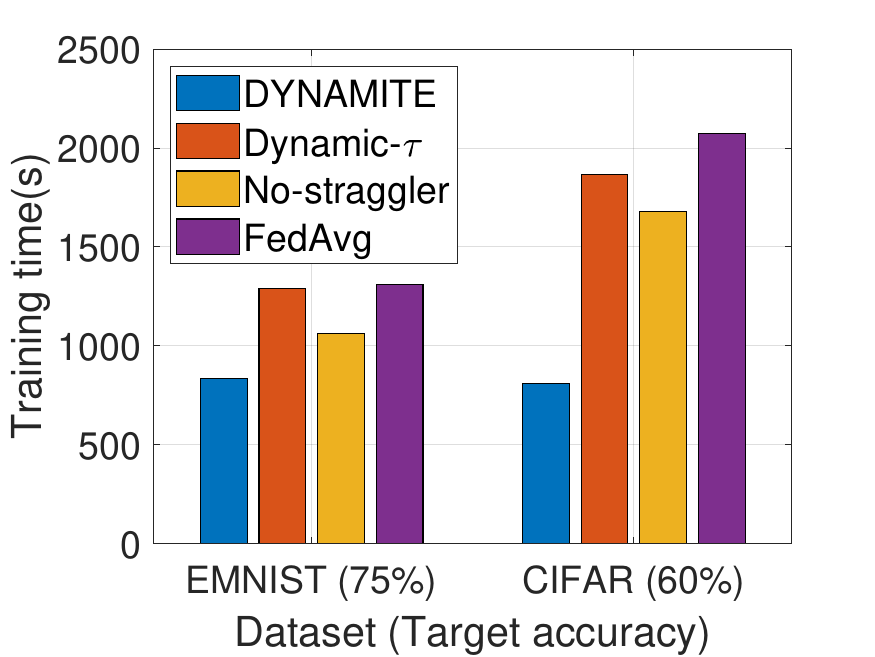}
		\caption{Smooth-Continuous}
		
	\end{subfigure}
	
	\begin{subfigure}[t]{0.24\linewidth}
		\includegraphics[width=\textwidth]{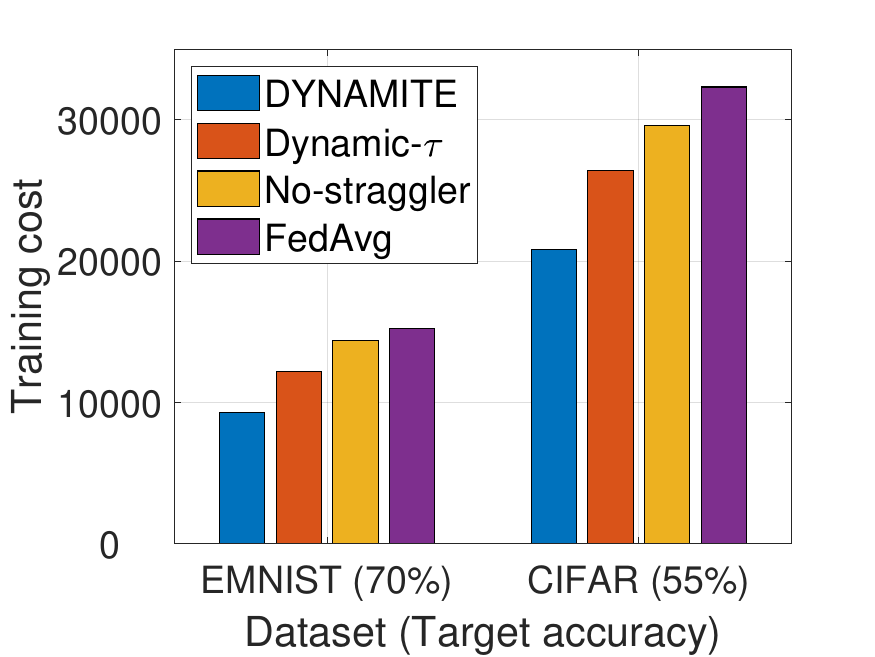}
		\caption{Burst-Continuous}
		
	\end{subfigure}
	\begin{subfigure}[t]{0.24\linewidth}
		\includegraphics[width=\textwidth]{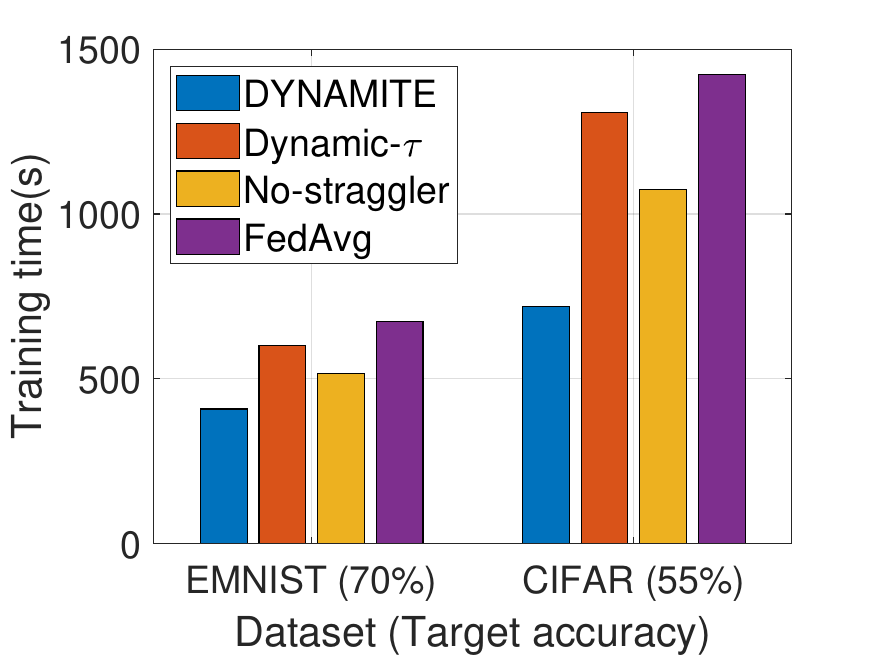}
		\caption{Burst-Continuous}
		
	\end{subfigure}
	\begin{subfigure}[t]{0.24\linewidth}
		\includegraphics[width=\textwidth]{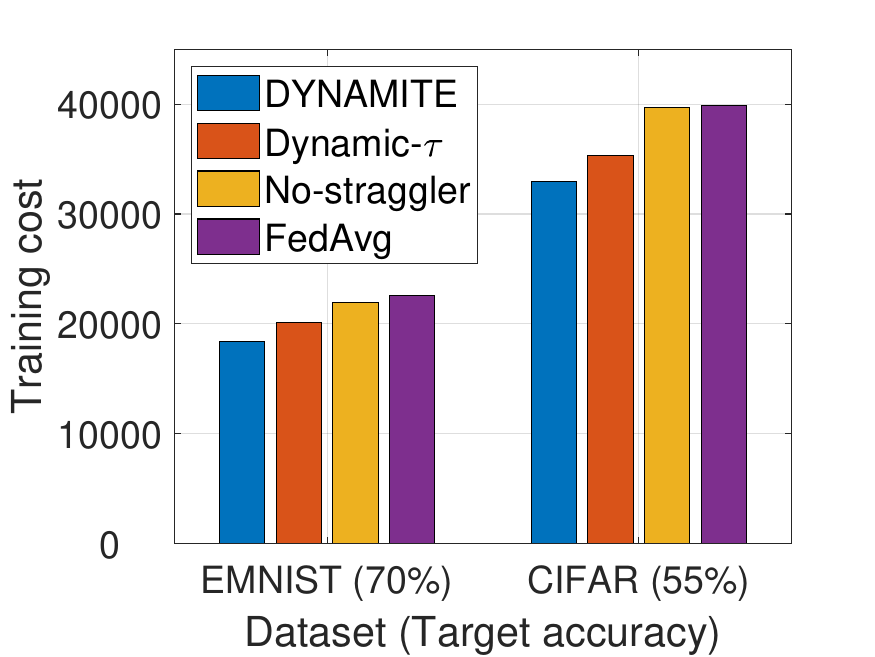}
		\caption{Random-Continuous}
		
	\end{subfigure}
	\begin{subfigure}[t]{0.24\linewidth}
		\includegraphics[width=\textwidth]{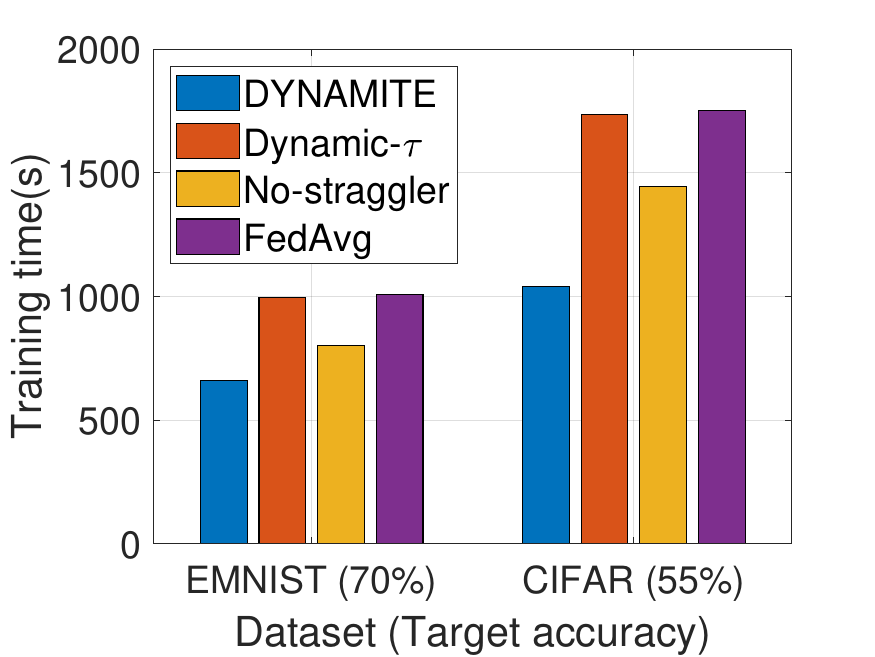}
		\caption{Random-Continuous}
		
	\end{subfigure}
	\caption{Training cost and time under EMNIST and CIFAR-10 (Reservoir Sampling)} 
	\label{fig:stream_bar}
\end{figure*}

\subsubsection{\bf Adaptive control for co-optimized aggregation frequency and heterogeneous batch sizes (streaming data)}
In this section, we further compare our \textbf{DYNAMITE} with three baselines using dynamic data streams with various data configurations presented in Section \ref{sec:stream_config}.    

{\bf Comparison of sampling methods.} Different from static datasets, data sampling strategies can be significant in FL training on dynamic data streams with limited on-device storage. Thus we first examine the impact of different data sampling strategies in different online settings. DYNAMITE(Res), DYNAMITE(Ran), and DYNAMITE(FIFO) are strategies of using DYNAMITE combined with sampling methods of Reservoir Sampling, Random Sampling, and FIFO, respectively. Fig.\ref{fig:sampling-cifar} shows that DYNAMITE(Res) which uses \textbf{Reservoir Sampling} has significant advantages over other sampling strategies, especially in the {\bf Continuous stream} settings, i.e., Smooth-Continuous, Burst-Continuous, Random-Continuous. It can be explained as follows. Both {\bf Random Sampling} and {\bf FIFO sampling} methods prefer to select data arrived later and discard data arrived earlier, which can easily lead to a biased model training and thus a biased global FL model. Fig. \ref{fig:sampling-cifar-smoothc} and Fig. \ref{fig:sampling-cifar-random} together show that biased sampling can not only lead to poor model performance but also catastrophic forgetting when training on a continuous data stream. On the other hand, if the data stream is i.i.d. (Fig \ref{fig:sampling-cifar-smoothi}), the performance difference among various sampling methods can be negligible, since the classes and the number of data samples that clients receive is always nearly the same, and thus how to select data is not of vital importance in this case. 

 We also notice that unlike other arrival patterns in continuous stream setting, DYNAMITE(Ran) and DYNAMITE(Res) can have similar performance in burst arrival setting (Fig. \ref{fig:sampling-cifar-burst}). We here present a reasonable interpretation to explain this result. First, {\bf Random Sampling} and {\bf Reservoir Sampling} are two similar methods in general. The major difference is that reservoir sampling guarantees that every data sample can have the same probability to be stored in the buffer by uniformly sampling both an up-coming and a previously stored data point. {\bf Random Sampling}, on the other hand, only discards data stored in the buffer uniformly at random, and it replaces a discarded data point with the latest-coming data. It then can inevitably result in biases, since the distribution of the selected data batch does not represent the full data stream. Consequently, model performance after every buffer update step is impeded, as presented in Fig. \ref{fig:sampling-cifar-smoothc} and \ref{fig:sampling-cifar-random}. However, clients receive data samples in a short period of time in {\bf Burst arrival} setting. So the clients have less frequent buffer updates compared to {\bf Smooth arrival} or {\bf Random arrival}, which eventually close the gap of these two similar sampling methods (Fig.~\ref{fig:sampling-cifar-burst}). 
 
 {\bf Full comparison of FL baselines and sampling methods.} In addition, we conduct extensive experiments on FL baselines (\textbf{Dynamic $\tau$, No-straggler and FedAvg}) combined with {\bf Reservoir Sampling} and {\bf Random Sampling} in Fig. \ref{fig:stream-cifar} under both EMNIST and CIFAR-10 datasets, revealing the importance and advantages of reservoir sampling.
%
%Therefore, we choose reservoir sampling method to further compare our \textbf{AdaCoOpt} with three benchmarks for EMNIST and CIFAR-10 FL training to demonstrate the effectiveness of our online control algorithm under streaming data scenario. 

{\bf Comparison in accuracy, cost, and run-time.} Moreover, we evaluate these control algorithms under different online stream settings presented in Section \ref{sec:stream_config}. Similar to the static dataset case, we compare these algorithms in two different scenarios, where the cost constraint and time constraint dominates respectively.  Table \ref{tab:stream_acc} and Figs. \ref{fig:stream-cifar}-\ref{fig:stream_bar} show that \textbf{DYNAMITE} can still outperform the baselines in both scenarios under different online data stream settings. \textbf{DYNAMITE} can achieve a 3.9\%--5.8\% higher final accuracy than FedAvg while reducing 16.7\%--51.2\% training cost in cost-dominant scenario, and a 7.9\%--21.3\% higher final accuracy with 39.4\%--63.8\% less completion time to achieve the same accuracy in time-dominant scenarios. Specifically, Fig. \ref{fig:train_loss_burst} shows that \textbf{DYNAMITE} can have a smoother training process and faster reboot in the burst scenario where a large number of training samples are fed into the clients suddenly.  Moreover, Fig. \ref{fig:stream-cifar} also reveals that our \textbf{DYNAMITE} still outperforms the baselines in different online data stream setting, either using {\bf Random Sampling} or {\bf Reservoir Sampling}, showing great adaptability of our \textbf{DYNAMITE}.

% In addition, for our larger scale experiments, our {\bf AdaCoOpt} can achieve a %1.2\%--4.6\xx\% higher test accuracy than {\bf No-straggler}, although incurring a moderately higher training time. It also reduces the training time by xx\%--xx\% compared to {\bf Dynamic $\tau$} and {\bf Uniform} while achieving up to 4.6\% higher test accuracy, verifying the scalability of our algorithm. 

%less cost than \textbf{Dynamic $\tau$} and spends 53\% less cost than the \textbf{Uniform} and \textbf{No-straggler}. 
%In the time-dominant scenario, our proposed algorithm spends 34\% less time than \textbf{Dynamic $\tau$} and \textbf{No-straggler}. Interestingly, when the time is the bottleneck, our algorithm also uses the strategy to ensure no straggler but has dynamic $\tau_k$ rather than static $\tau$ and thus outputs better combination of $\tau_k$ and $\mathbf{s}_k$ than {\bf No-straggler}. We also manage to fuse the \textbf{Dynamic $\tau$} algorithm in \cite{wang2019adaptive} and \textbf{\bf No-straggler} strategy to set a more competitive baseline in time-dominant scenario. Still, our proposed algorithm can spend 12\% less time than this competitive baseline.

	\section{Conclusion}
	This work proposes a novel framework to quantify and optimize the interplay of the number of local update steps and heterogeneous batch sizes across clients for federated learning performed at distributed edge devices. Technically, we derive a novel convergence bound with respect to those control variables, and analyze the performance metrics of cost and training time as well. 
	We then provide closed-form solutions for our joint optimization in special cases and propose an efficient exact algorithm for the general case. Our strategies consider both heterogeneous system characteristics and non-i.i.d. data, which can improve the common strategies that FL practitioners adopt. Moreover, we adapt our offline strategy to dynamically adjust the decisions on the fly, with superiority of several performances demonstrated in extensive experiments. 

	%\newpage
	
	%\balance
	%\opt{secon}{\input{short_appendix.tex}}
	%\opt{tr}{\input{9_full_appendix}}
	\bibliographystyle{IEEEtran}
	\bibliography{main.bib}

\begin{IEEEbiography}[{\includegraphics[width=1in,height=1.25in,clip,keepaspectratio]{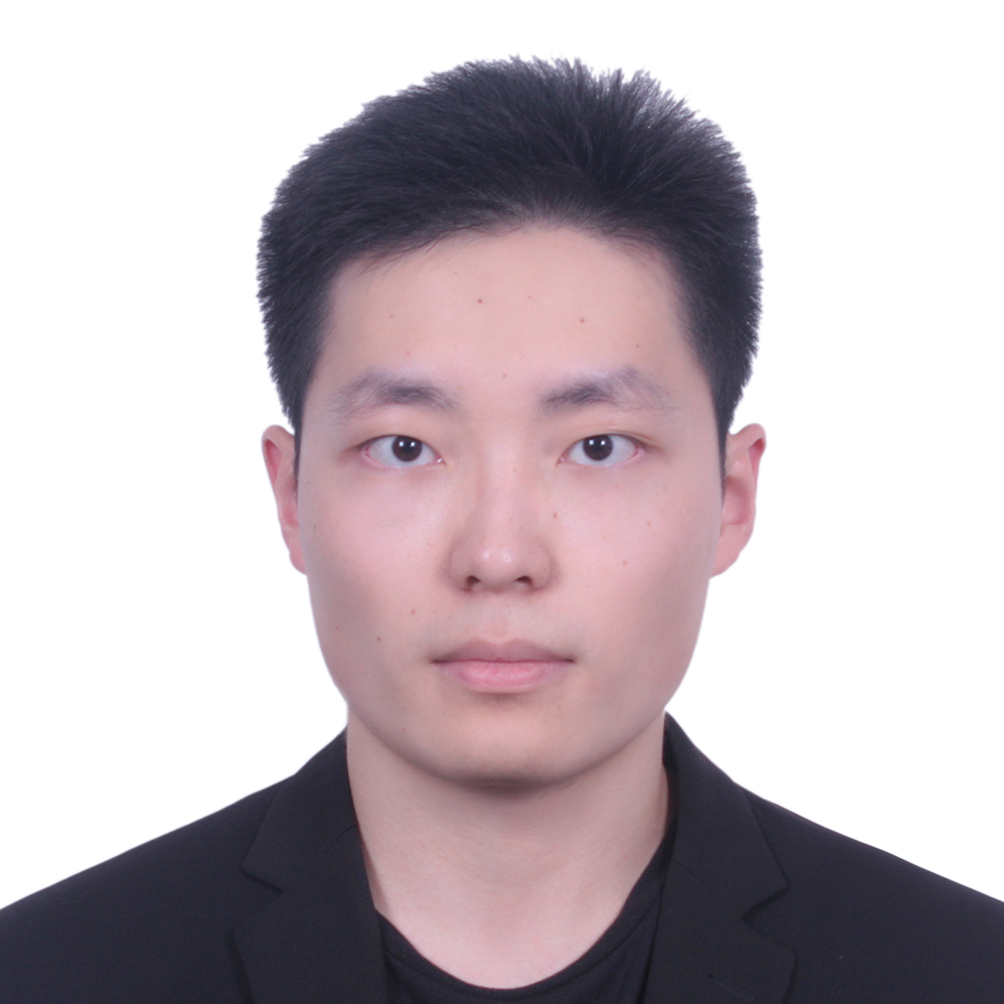}}]{Weijie Liu}
(Student Member, IEEE) received the B.E. degree in electronics and communication engineering from the Sun Yat-sen University in 2021. He is currently working toward the M.E. degree at the School of Computer Science and Engineering, Sun Yat-sen University, Guangzhou, China. His research interests include federated learning and edge computing.
\end{IEEEbiography}

\begin{IEEEbiography}[{\includegraphics[width=1in,height=1.25in,clip,keepaspectratio]{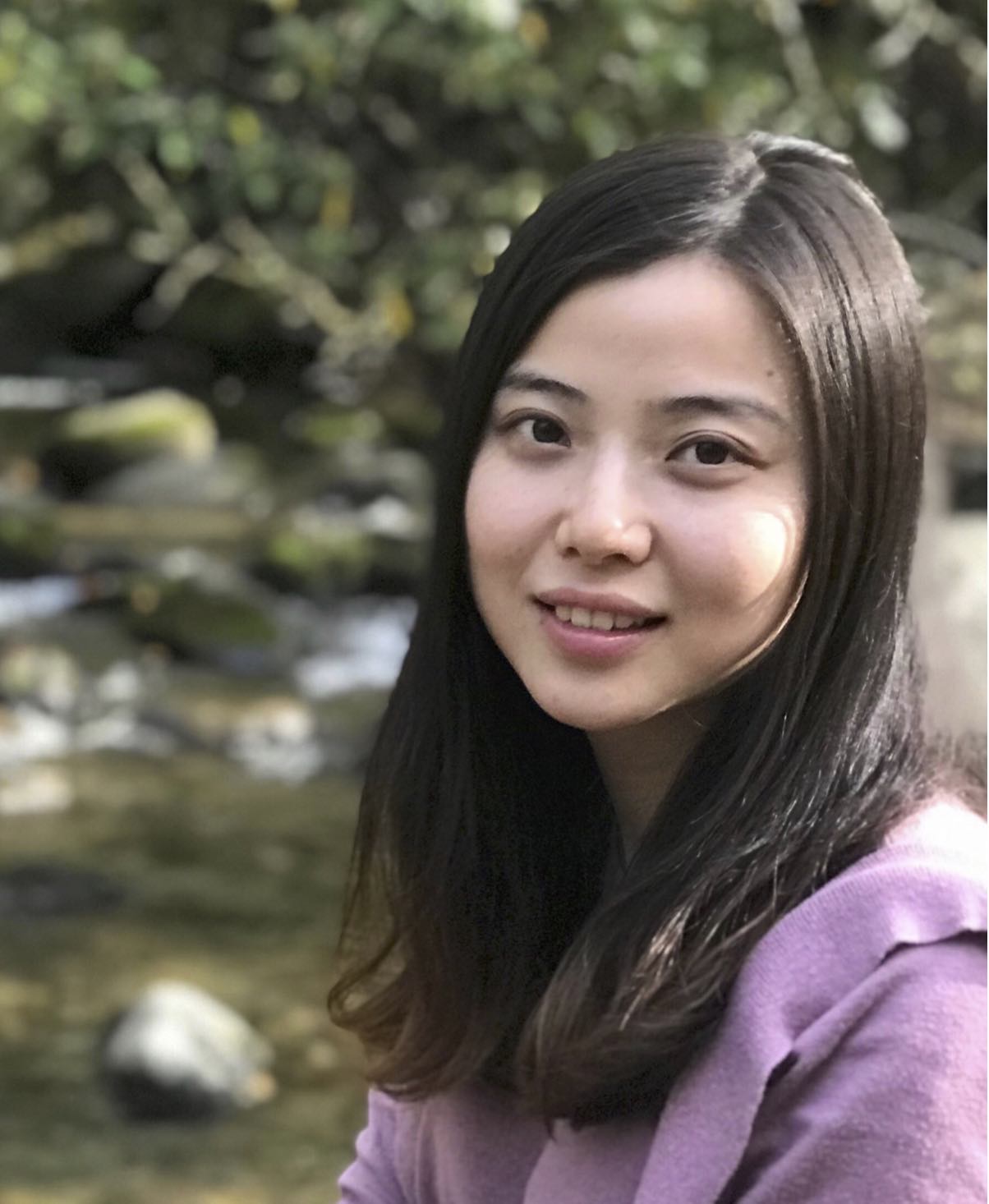}}]{Xiaoxi Zhang}
(Member, IEEE) received the B.E. degree in electronics and information engineering from the Huazhong University of Science and Technology in 2013 and the Ph.D. degree in computer science from The University of Hong Kong in 2017. She is currently an Associate Professor with the School of Computer Science and Engineering, Sun Yat-sen University. Before joining SYSU, she was
a Post-Doctoral Researcher with the Department of Electrical and Computer Engineering, Carnegie Mellon University. She is broadly interested in optimization and algorithm design for networked systems, including cloud and edge computing networks, NFV systems, and distributed machine learning systems.
\end{IEEEbiography}

\begin{IEEEbiography}[{\includegraphics[width=1in,height=1.25in,clip,keepaspectratio]{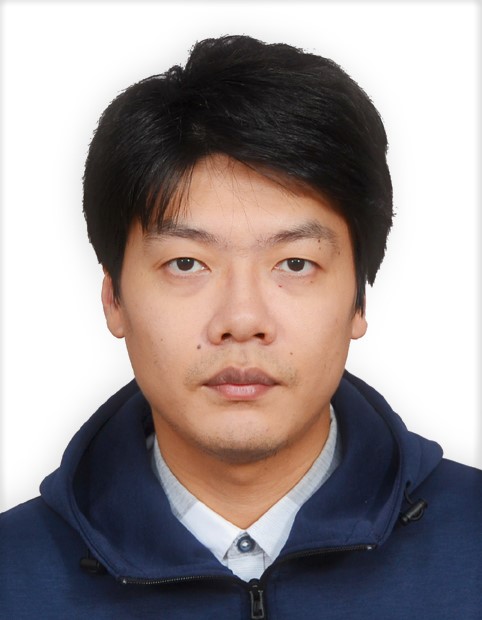}}]{Jingpu Duan}
(Member, IEEE) received the B.E. degree from the Huazhong University of Science and Technology, Wuhan, China, in 2013, and the
Ph.D. degree from the University of Hong Kong, Hong Kong, China, in 2018. He is currently a Research Assistant Professor with the Institute of Future Networks, Southern University of Science and Technology, Shenzhen, China. He also works with the Department of Communications, Pengcheng Laboratory, Shenzhen, China. His research interest includes designing and implementing high-performance networking systems.
\end{IEEEbiography}

\begin{IEEEbiography}[{\includegraphics[width=1in,height=1.25in,clip,keepaspectratio]{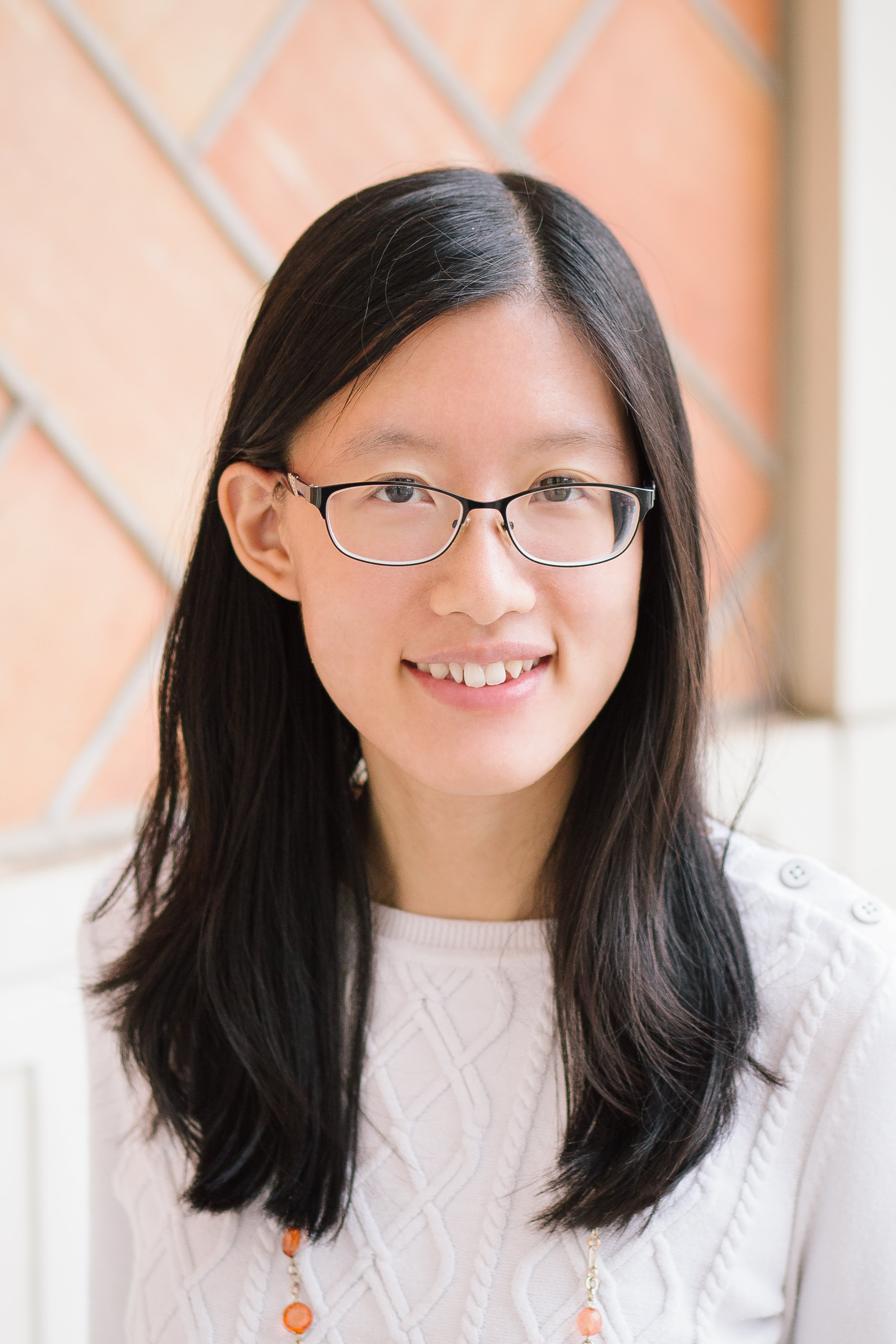}}]{Carlee Joe-Wong}
(Senior Member, IEEE) received the A.B. degree (magna cum laude) in mathematics, and the M.A. and Ph.D. degrees in applied and computational mathematics from Princeton University in 2011, 2013, and 2016, respectively. From 2013 to 2014, she was the Director of Advanced Research at DataMi, a startup she co-founded from her research on mobile data pricing. She is currently
the Robert E. Doherty Assistant Professor of electrical and computer engineering with Carnegie Mellon University. Her research interests lie in optimizing various types of networked systems, including applications of machine learning and pricing to cloud computing, mobile/wireless networks, and ridesharing networks. She received the NSF CAREER Award in 2018 and the ARO Young
Investigator Award in 2019.
\end{IEEEbiography}

\begin{IEEEbiography}[{\includegraphics[width=1in,height=1.25in,clip,keepaspectratio]{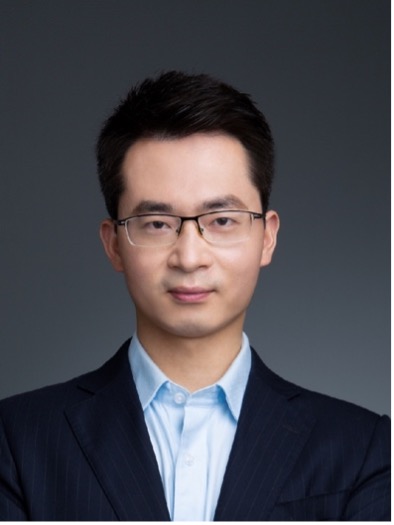}}]{Zhi Zhou}
(Member, IEEE) received the B.S., M.E., and Ph.D. degrees from the School of Computer Science and Technology, Huazhong University of Science and Technology (HUST), Wuhan, China, in 2012, 2014, and 2017, respectively. He is currently an associate professor with the School of Computer Science and Engineering, Sun Yat-sen University, Guangzhou, China. In 2016, he has been a visiting scholar with the University of Gottingen. His research interests include edge computing and cloud computing and distributed systems.
\end{IEEEbiography}

\begin{IEEEbiography}[{\includegraphics[width=1in,height=1.25in,clip,keepaspectratio]{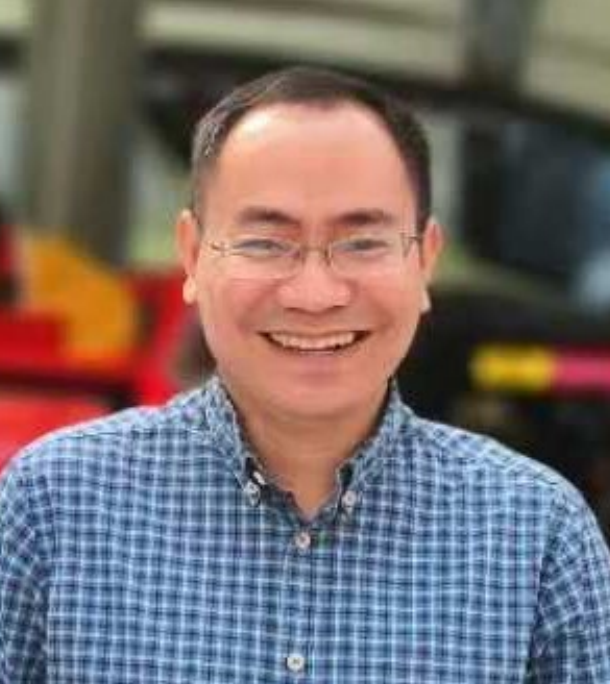}}]{Xu Chen}
(Senior Member, IEEE) received the Ph.D. degree in information engineering from the Chinese University of Hong Kong, Hong Kong, in 2012. He is currently a Full Professor with Sun Yat-sen University, Guangzhou, China, the Director of Institute of Advanced Networking and Computing Systems, and Vice Director of National and Local Joint Engineering Laboratory. From 2012 to 2014, he was a Postdoctoral Research Associate with Arizona State University, Tempe, AZ, USA, and from 2014 to 2016 a Humboldt Scholar Fellow with the Institute of Computer Science of the University of Goettingen, Germany. He was the recipient of the prestigious Humboldt Research Fellowship Awarded by the Alexander von Humboldt Foundation of Germany, 2014 Hong Kong Young Scientist Runner-up Award, 2017 IEEE Communication Society Asia-Pacific Outstanding Young Researcher Award, 2017 IEEE ComSoc Young Professional Best Paper Award, Honorable Mention Award of 2010 IEEE international conference on Intelligence and Security Informatics, Best Paper Runner-up Award of 2014 IEEE International Conference on Computer Communications, and Best Paper Award of 2017 IEEE International Conference on Communications. He is currently the Area Editor the IEEE OPEN JOURNAL OF THE Communications Society, he is an Associate Editor for the IEEE Transactions on Wireless Communications, IEEE Transactions on Vehicular Technology, IEEE Internet of Things Journal, and IEEE Journal on Selected Areas in Communications Series on Network Softwarization and Enablers.
\end{IEEEbiography}

\clearpage

\opt{tr}{\begin{appendices}
\allowdisplaybreaks[3]

\section{Proof of Theorem \ref{thm:bound-vanilla} and Lemma \ref{lem:marginal_bound}}
\label{sec:proof_vanilla}
\begin{lem}
	For any interval $[k]$, and t $\in [(k-1)\tau,k\tau)$,we have
	\begin{equation*}
		\| \mathbf{w}_i(t)-\mathbf{v}_{[k]}(t)\|\leq h_i(t-(k-1)\tau), 
	\end{equation*}
	where $h_{i}(x) = \frac{\delta_{i}}{\beta}\left((\eta \beta+1)^{x}-1\right)$.
\end{lem}
\noindent \emph{Proof.}
When $t=(k-1)\tau$, we have $\mathbf{w}_i(t)=\mathbf{v}_{[k]}(t)$ by the definition of $\mathbf{v}_{[k]}(t)$. Therefore we have $\| \mathbf{w}_i(t)-\mathbf{v}_{[k]}(t)\|= h_i(0)=0 $.
For the induction, we assume that
\begin{equation*}
	\| \mathbf{w}_i(t-1)-\mathbf{v}_{[k]}(t-1)\|\leq h_i(t-1-(k-1)\tau) 
\end{equation*}
We now show that $\| \mathbf{w}_i(t)-\mathbf{v}_{[k]}(t)\|\leq h_i(t-(k-1)\tau) $ holds for $t$. We have
\begin{align*}
	&\left\|\mathbf{w}_{i}(t)-\mathbf{v}_{[k]}(t)\right\|  \\
	=& \|\left(\mathbf{w}_{i}(t-1)-\eta g_{i}\left(\mathbf{w}_{i}(t-1)\right)\right) \\
	&\left.-\left(\mathbf{v}_{[k]}(t-1)-\eta g\left(\mathbf{v}_{[k]}(t-1)\right)\right) \| \right) \\
	=& \|\left(\mathbf{w}_{i}(t-1)-\mathbf{v}_{[k]}(t-1)\right)-\eta\left[g_{i}\left(\mathbf{w}_{i}(t-1)\right)\right.\\
	&\left.-g_{i}\left(\mathbf{v}_{[k]}(t-1)\right)+g_{i}\left(\mathbf{v}_{[k]}(t-1)\right)-g\left(\mathbf{v}_{[k]}(t-1)\right)\right] \| \\
	\leq &\left\|\mathbf{w}_{i}(t-1)-\mathbf{v}_{[k]}(t-1)\right\| \\
	&+\eta\left\|g_{i}\left(\mathbf{w}_{i}(t-1)\right)-g_{i}\left(\mathbf{v}_{[k]}(t-1)\right)\right\| \\
	&+\eta\left\|g_{i}\left(\mathbf{v}_{[k]}(t-1)\right)-g\left(\mathbf{v}_{[k]}(t-1)\right)\right\| \\
	\leq &(\eta \beta+1)\left\|\mathbf{w}_{i}(t-1)-\mathbf{v}_{[k]}(t-1)\right\|+\eta \delta_{i} \\
	\leq &(\eta \beta+1) g_{i}(t-1-(k-1) \tau)+\eta \delta_{i} \\
	=&(\eta \beta+1)\left(\frac{\delta_{i}}{\beta}\left((\eta \beta+1)^{t-1-(k-1) \tau}-1\right)\right)+\eta \delta_{i} \\
	=& \frac{\delta_{i}}{\beta}(\eta \beta+1)^{t-(k-1) \tau}-\frac{\delta_{i}}{\beta}(\eta \beta+1)+\eta \delta_{i} \\
	=& \frac{\delta_{i}}{\beta}(\eta \beta+1)^{t-(k-1) \tau}-\frac{\delta_{i}}{\beta} \\
	=& \frac{\delta_{i}}{\beta}\left((\eta \beta+1)^{t-(k-1) \tau}-1\right) \\
	=& h_{i}(t-(k-1) \tau).
\end{align*}
\begin{lem}
	For any interval $[k]$, and t $\in [(k-1)\tau,k\tau)$,we have
	\begin{equation*}
		\| \mathbf{w}(t)-\mathbf{v}_{[k]}(t)\|\leq h(t-(k-1)\tau), 
	\end{equation*}
	where $h(x) = \frac{\delta}{\beta}\left((\eta \beta+1)^{x}-1\right)-\eta\delta x$.
\end{lem}
\noindent\emph{Proof.}
\begin{align*}
	&\left\|\mathbf{w}(t)-\mathbf{v}_{[k]}(t)\right\|\\
	&=\| \mathbf{w}(t-1)-\eta \frac{\sum_{i} D_{i}g_{i}\left(\mathbf{w}_{i}(t-1)\right)}{D}-\mathbf{v}_{[k]}(t-1)\\
	&+\eta g\left(\mathbf{v}_{[k]}(t-1)\right) \|\\
	&=\| \mathbf{w}(t-1)-\mathbf{v}_{[k]}(t-1)\\
	&-\eta\left(\frac{\sum_{i} D_{i} g_{i}\left(\mathbf{w}_{i}(t-1)\right)}{D}-g\left(\mathbf{v}_{[k]}(t-1)\right)\right) \|\\
	&=\| \mathbf{w}(t-1)-\mathbf{v}_{[k]}(t-1)\\
	&-\eta\left(\frac{\sum_{i} D_{i}\left(g_{i}\left(\mathbf{w}_{i}(t-1)\right)-g_{i}\left(\mathbf{v}_{[k]}(t-1)\right)\right)}{D}\right) \|\\
	&\leq\left\|\mathbf{w}(t-1)-\mathbf{v}_{[k]}(t-1)\right\|\\
	&+\eta\left(\frac{\sum_{i} D_{i}\left\|g_{i}\left(\mathbf{w}_{i}(t-1)\right)-g_{i}\left(\mathbf{v}_{[k]}(t-1)\right)\right\|}{D}\right)\\
	&\leq\left\|\mathbf{w}(t-1)-\mathbf{v}_{[k]}(t-1)\right\|\\
	&+\eta \beta\left(\frac{\sum_{i} D_{i}\left\|\mathbf{w}_{i}(t-1)-\mathbf{v}_{[k]}(t-1)\right\|}{D}\right)\\
	&\leq\left\|\mathbf{w}(t-1)-\mathbf{v}_{[k]}(t-1)\right\|\\
	&+\eta \beta \frac{{\sum}_{i} D_{i} h_{i}(t-1-(k-1) \tau)}{D}\\
	&=\left\|\mathbf{w}(t-1)-\mathbf{v}_{[k]}(t-1)\right\|\\
	&+\eta \beta\left(\frac{\sum_{i} D_{i} \frac{\delta_{i}}{\beta}\left((\eta \beta+1)^{t-1-(k-1)\tau}-1\right)}{D}\right)\\
	&=\left\|\mathbf{w}(t-1)-\mathbf{v}_{[k]}(t-1)\right\|\\
	&+\eta\left(\frac{\sum_{i} D_{i} \delta_{i}}{D}\right)\left((\eta \beta+1)^{t-1-(k-1)\tau}-1\right)\\
	&=\left\|\mathbf{w}(t-1)-\mathbf{v}_{[k]}(t-1)\right\|+\eta \delta\left((\eta \beta+1)^{t-1-(k-1) \tau}-1\right).
\end{align*}
Equivalently,
\begin{align*}
	&\left\|\mathbf{w}(t)-\mathbf{v}_{[k]}(t)\right\|-\left\|\mathbf{w}(t-1)-\mathbf{v}_{[k]}(t-1)\right\|\\
	&\leq\eta \delta\left((\eta \beta+1)^{t-1-(k-1)\tau}-1\right)
\end{align*}
Summing up the above equation over different values of t, we have
\begin{align*}
	&\left\|\mathbf{w}(t)-\mathbf{v}_{[k]}(t)\right\| \\
	&=\sum_{y=(k-1) \tau+1}^{t}\left\|\mathbf{w}(y)-\mathbf{v}_{[k]}(y)\right\|-\left\|\mathbf{w}(y-1)-\mathbf{v}_{[k]}(y-1)\right\| \\
	&\leq \eta \delta \sum_{y=(k-1) \tau+1}^{t}\left((\eta \beta+1)^{y-1-(k-1) \tau}-1\right) \\
	&=\eta \delta \sum_{z=1}^{t-(k-1) \tau}\left((\eta \beta+1)^{z-1}-1\right) \\
	&=\eta \delta \sum_{z=1}^{t-(k-1) \tau}(\eta \beta+1)^{z-1}-\eta \delta(t-(k-1) \tau) \\
	&=\eta \delta \frac{\left(1-(\eta \beta+1)^{t-(k-1) \tau}\right)}{-\eta \beta}-\eta \delta(t-(k-1) \tau) \\
	&=\eta \delta \frac{(\eta \beta+1)^{t-(k-1) \tau}-1}{\eta \beta}-\eta \delta(t-(k-1) \tau) \\
	&=\frac{\delta}{\beta}\left((\eta \beta+1)^{t-(k-1) \tau}-1\right)-\eta \delta(t-(k-1) \tau) \\
	&=h(t-(k-1) \tau).
\end{align*}
Using the $\rho$-quadratic-Lipschitz property of $F_i(x)$, we have
\begin{equation*}
	F(\mathbf{w}(t))-F(\mathbf{v}_{[k]}(t))\leq\rho h(t-(k-1) \tau)^{2}.
\end{equation*}

\begin{lem}
	\label{lem:var_s}
	\begin{equation*}
		\mathbb{E}\left[\left\| g(\mathbf{v}_{[k]}(t),\xi_t) \right\|_2^2\right]\leq\frac{1}{D^2}\sum\limits_{i\in [N]}\frac{M_iD_i^2}{s_i}+\mu_G^2\left\|\nabla F_{i}(\mathbf{w})\right\|^2.
	\end{equation*}
\end{lem}
\noindent\emph{Proof.}
\begin{equation*}
	\because g_i(\mathbf{v}_{[k]}(t),\xi_t)=\sum\limits_{\mathbf{x}_{i,j}\in \mathcal{S}_i}\frac{\nabla f(\mathbf{w},\mathbf{x}_{i,j})}{s_i}
\end{equation*}
\begin{equation*}
	\therefore\mathbb{V}\left[g_i(\mathbf{v}_{[k]}(t),\xi_t)\right]\leq \frac{M_i}{s_i}
\end{equation*}
\begin{align*}
	\because g(\mathbf{v}_{[k]}(t),\xi_t)=&\sum\limits_{i=1}^{N}\frac{D_ig_i(\mathbf{v}_{[k]}(t),\xi_t)}{D} \\
	\therefore\mathbb{V}\left[g(\mathbf{v}_{[k]}(t),\xi_t)\right]\leq&\sum\limits_{i\in [N]}\frac{D_i^2M_i}{D^2s_i}
\end{align*}
%\xiaoxi{Changed $c_i$ to $D_i/D$ since $c$ is the strongly convexity parameter.}
Given the above, we have:
\begin{align*}
	&\mathbb{E}\left[\left\| g(\mathbf{v}_{[k]}(t),\xi_t) \right\|_2^2\right]\\
	=&\left\|\mathbb{E}_{\xi_t}[g(\mathbf{v}_{[k]}(t),\xi_t)\right\|_{2}^{2}+\mathbb{V}\left[g(\mathbf{v}_{[k]}(t),\xi_t)\right]\\
	\leq&\;\mu_G^2\left\|\nabla F_{i}(\mathbf{w})\right\|^2+\frac{1}{D^2}\sum\limits_{i\in [N]}\frac{M_iD_i^2}{s_i}.
\end{align*}

\begin{lem}
	For any interval $[k]$, and t $\in [(k-1)\tau,k\tau)$, when $\eta\leq\frac{\mu}{\beta M_G}$, $M_G=\mu_{G}^2$, we have
	\begin{align*}
		&\mathbb{E}[F(\mathbf{v}_{[k]}(t+1))]-F^*\\
		\leq &(1-\eta c\mu)\left(\mathbb{E}[F(\mathbf{v}_{[k]}(t))]-F^*\right)
		+\frac{\beta\eta^2 }{2D^2}\sum\limits_{i\in [N]}\frac{M_iD_i^2}{s_i}.
	\end{align*}
\end{lem}
\noindent\emph{Proof.}
\begin{align*}
	&F(\mathbf{v}_{[k]}(t+1))-F(\mathbf{v}_{[k]}(t))\\
	\leq&\nabla F(\mathbf{v}_{[k]}(t))^{T}(\mathbf{v}_{[k]}(t+1)-\mathbf{v}_{[k]}(t))\\
	+&\frac{\beta}{2}\left\|\mathbf{v}_{[k]}(t+1)-\mathbf{v}_{[k]}(t)\right\|^{2}\\
	=&-\eta\nabla F(\mathbf{v}_{[k]}(t))^{T}g(\mathbf{v}_{[k]}(t),\xi_t)+\frac{\beta\eta^2}{2}\left\| g(\mathbf{v}_{[k]}(t),\xi_t) \right\|^{2}.
\end{align*}
Taking the expectation on the $t_{th}$ sample process $\xi_t$ and using the PŁ-condition: $2c(F(\mathbf{w})-F^*)\leq\left\|\nabla F(\mathbf{w})\right\|_2^2$.
\begin{align*}
	&\mathbb{E}_{\xi_t}[F(\mathbf{v}_{[k]}(t+1))]-F(\mathbf{v}_{[k]}(t)\\
	\leq&-\eta\nabla F(\mathbf{v}_{[k]}(t))^{T}\mathbb{E}_{\xi_t}[g(\mathbf{v}_{[k]}(t),\xi_t)]\\
	&+\frac{\beta\eta^2}{2}\mathbb{E}_{\xi_t}[\left\| g(\mathbf{v}_{[k]}(t),\xi_t) \right\|^{2}]\\
	\leq&-\eta(\mu-\frac{\beta\eta M_G}{2})\left\| F(\mathbf{v}_{[k]}(t)) \right\|^{2}+\frac{\beta\eta^2}{2D^2}\sum\limits_{i\in [N]}\frac{M_iD_i^2}{s_i}\\
	\leq&-\eta c\mu(F(\mathbf{v}_{[k]}(t))-F^*)+\frac{\beta\eta^2 }{2D^2}\sum\limits_{i\in [N]}\frac{M_iD_i^2}{s_i}.
\end{align*}
Subtract $F^*$ from both sides, then taking total expectation.
\begin{align*}
	\mathbb{E}[F(\mathbf{v}_{[k]}(t+1))]-F^*&\leq (1-\eta c\mu)\left(\mathbb{E}[F(\mathbf{v}_{[k]}(t))]-F^*\right)\\
	&+\frac{\beta\eta^2 }{2D^2}\sum\limits_{i\in [N]}\frac{M_iD_i^2}{s_i}.
\end{align*}
% \textbf{Now we are ready to prove our error bound presented in Theorem 1.}
% \noindent\emph{Proof of Theorem 1}
Let $G_{[k]}(t)=\mathbb{E}[F(\mathbf{v}_{[k]}(t))]-F^*$, $q=1-\eta c\mu$, we have 
\begin{equation*}
	G_{[k]}(t+1)\leq qG_{[k]}(t)+\frac{\beta\eta^2}{2D^2}\sum\limits_{i\in [N]}\frac{M_iD_i^2}{s_i}.
\end{equation*}
Apply the above inequality recursively over $\tau$ local updates.
\begin{equation*}
	G_{[k]}(k\tau)\leq q^{\tau}G_{[k]}((k-1)\tau)+\frac{1-q^{\tau}}{1-q}\cdot\frac{\beta\eta^2 }{2D^2}\sum\limits_{i\in [N]}\frac{M_iD_i^2}{s_i}.
\end{equation*}
By the definition of $G_{[k]}(k\tau)$ and $F(\mathbf{v}_{[k]}(t+1))$, we have
\begin{align*}
	&G_{[k+1]}(k\tau)-G_{[k]}(k\tau)=F(\mathbf{v}_{[k+1]}(k\tau))-F(\mathbf{v}_{[k]}(k\tau))\\
	=&F(\mathbf{w}(k\tau))-F(\mathbf{v}_{[k]}(k\tau))\\
	\leq& \rho h(\tau)^{2}.
\end{align*}		

Combining the equations above, we can have

\begin{align}
	&G_{[k+1]}(k\tau)\leq G_{[k]}(k\tau)+\rho h(\tau)^{2}\\
	\leq&q^{\tau}G_{[k]}((k-1)\tau)+\frac{1-q^{\tau}}{1-q}\cdot\frac{\beta\eta^2 }{2D^2}\sum\limits_{i\in [N]}\frac{M_iD_i^2}{s_i}+\rho h(\tau)^{2}.
	\label{eq:marginal_t}
\end{align}	

Apply the above inequality recursively over all $K$ communication rounds.
\begin{align*}
	&\mathbb{E}[F(\mathbf{w}(K\tau))]-F^*\leq q^{K\tau}\left(F(\mathbf{w}(0)-F^*\right)\\
	+&\frac{1-q^{K\tau}}{1-q^{\tau}}\left(\frac{1-q^{\tau}}{1-q}\cdot\frac{\beta\eta^2}{2D^2}\sum\limits_{i\in [N]}\frac{M_iD_i^2}{s_i}+\rho h(\tau)^{2}\right).
\end{align*}
Due to the following inequality:
\begin{equation*}
	\frac{1-q^{K\tau}}{1-q^{\tau}}\leq\frac{1-q^{K}}{1-q}, 
\end{equation*}
Theorem \ref{thm:bound-vanilla} is as follows: 
\begin{align*}
	&\mathbb{E}[F(\mathbf{w}(K\tau))]-F^*\leq q^{K\tau}\left(F(\mathbf{w}(0)-F^*\right)\\
	+&\frac{1-q^{K}}{1-q}\left(\frac{1-q^{\tau}}{1-q}\cdot\frac{\beta\eta^2}{2D^2}\sum\limits_{i\in [N]}\frac{M_iD_i^2}{s_i}+\rho h(\tau)^{2}\right),
\end{align*}
where we have:
\begin{equation*}
	h(\tau) \triangleq \frac{\delta}{\beta}\left((\eta \beta+1)^{\tau}-1\right)-\eta\delta \tau.
\end{equation*}

The above inequality (\ref{eq:marginal_t}) can derive a marginal error bound which adapts to dynamic local update steps $\tau_k$, mini-batch size $s_i^k$ and streaming data set $\mathcal{D}_i^{k}$ in every FL round $k$ by rewriting (\ref{eq:marginal_t}) for a single FL round $k$ as follows:

\begin{align*}
	\mathbb{E}[F^k(\mathbf{w}^{(k)})-F^*]&\leq q^{\tau_k}\left(\mathbb{E}[F^k(\mathbf{w}^{(k-1)})-F^*]\right)
	\\&+\frac{1-q^{\tau_k}}{1-q}\cdot\frac{\beta\eta^2}{2D_{k}^2}\sum\limits_{i\in [N]}\frac{M_i^{k}D_i^{k^2}}{s_i^k}+\rho h(\tau_k)^{2},
\end{align*}
where $\mathbf{w}^{(k)}\triangleq \mathbf{w}(\sum_{i=1}^{k}\tau_i)$. Then we define $\psi^{k} = \mathbb{E}[F^{k}(\mathbf{w}^{(k-1)})-F^{k-1}(\mathbf{w}^{(k-1)})]$ to quantify and describe the freshness and heterogeneity of the latest receiving data in round $k$ and the Lemma \ref{lem:marginal_bound} is as follows:

\begin{equation}
		\begin{split}
			\mathbb{E}[F^{k}(\mathbf{w}^{(k)})]-F^*&\leq q^{\tau_k}[\mathbb{E}[F^{k-1}(\mathbf{w}^{(k-1)})]-F^*+\psi^{k}]\\
			&+\frac{\beta\eta^2(1-q^{\tau_k})}{2D_{k}^2(1-q)}\sum\limits_{i\in \mathcal{N}}\frac{M_{i}D_{i}^{k^2}}{s_{i}^{k}}+\rho h(\tau_k)^2.
		\end{split}
	\end{equation}

\section{Proof of Theorem \ref{thm:opt_uniform} and Corollary \ref{cor:bs-gpu}}
\label{sec:proof_uniform}

\begin{figure*}[t]
  \centering
%---------------------------------------
% Reservoir
  \begin{subfigure}{0.32\textwidth}
    \centering
    \includegraphics[width=\linewidth]{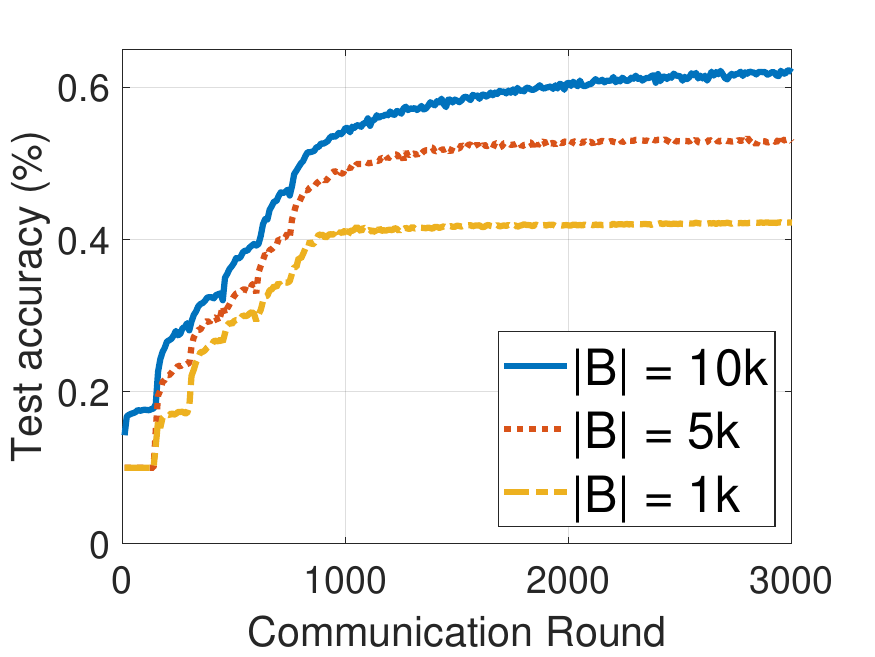}
    \caption{Smooth arrival - Reservoir}
    
  \end{subfigure}
  \hfill
  \begin{subfigure}{0.32\textwidth}
    \centering
    \includegraphics[width=\linewidth]{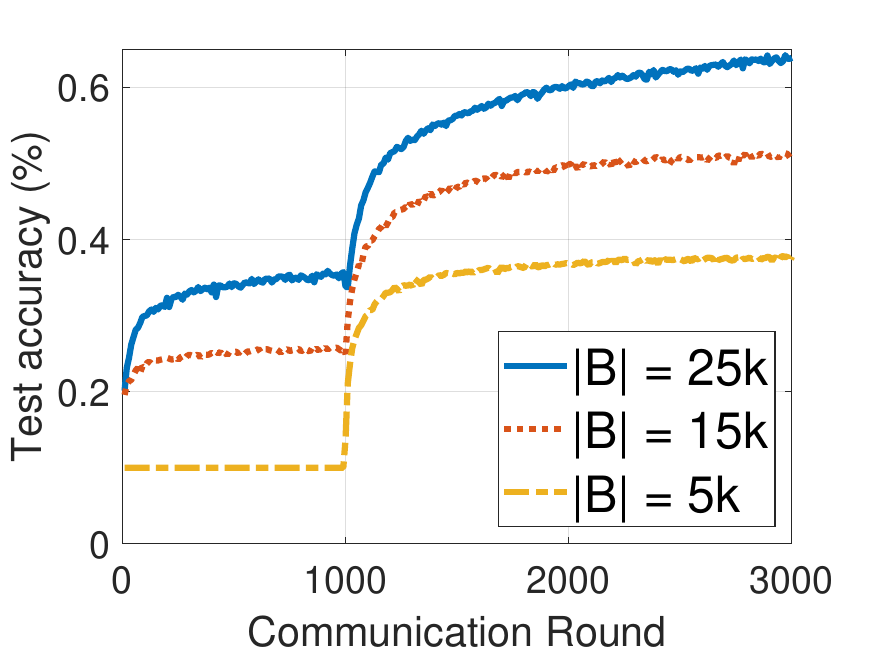}
    \caption{Burst arrival - Reservoir}
   
  \end{subfigure}
  \hfill
  \begin{subfigure}{0.32\textwidth}
    \centering
    \includegraphics[width=\linewidth]{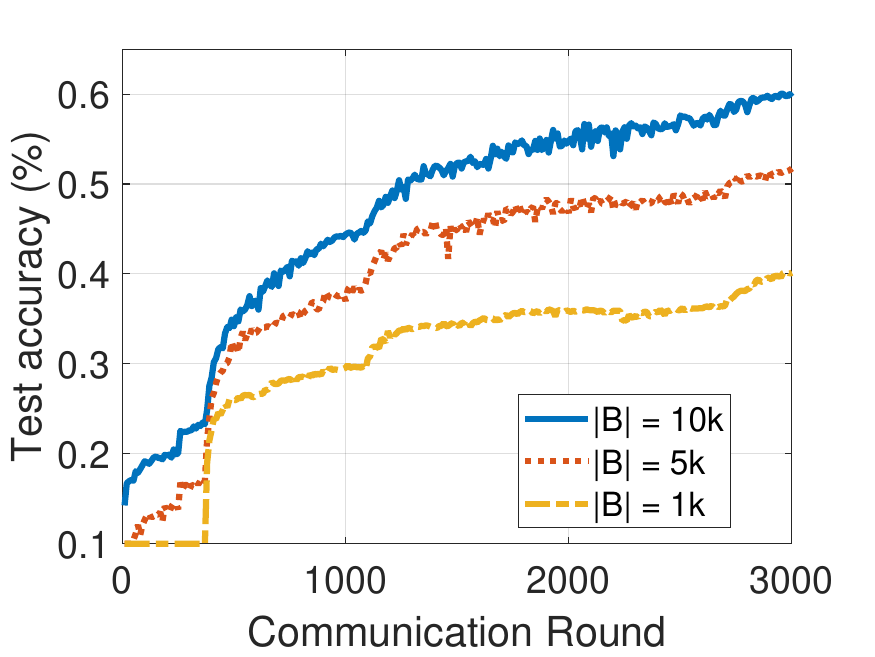}
    \caption{Random arrival - Reservoir}
    
  \end{subfigure}
%------------------------------------------------
% Random Sampling
  \begin{subfigure}{0.32\textwidth}
    \centering
    \includegraphics[width=\linewidth]{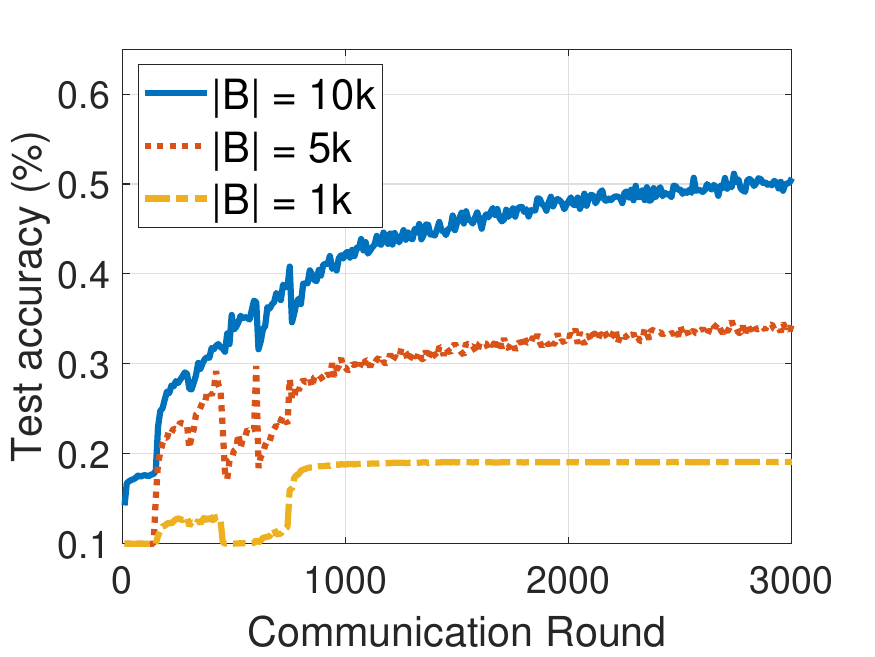}
    \caption{Smooth arrival - Random}
    
  \end{subfigure}
  \hfill
  \begin{subfigure}{0.32\textwidth}
    \centering
    \includegraphics[width=\linewidth]{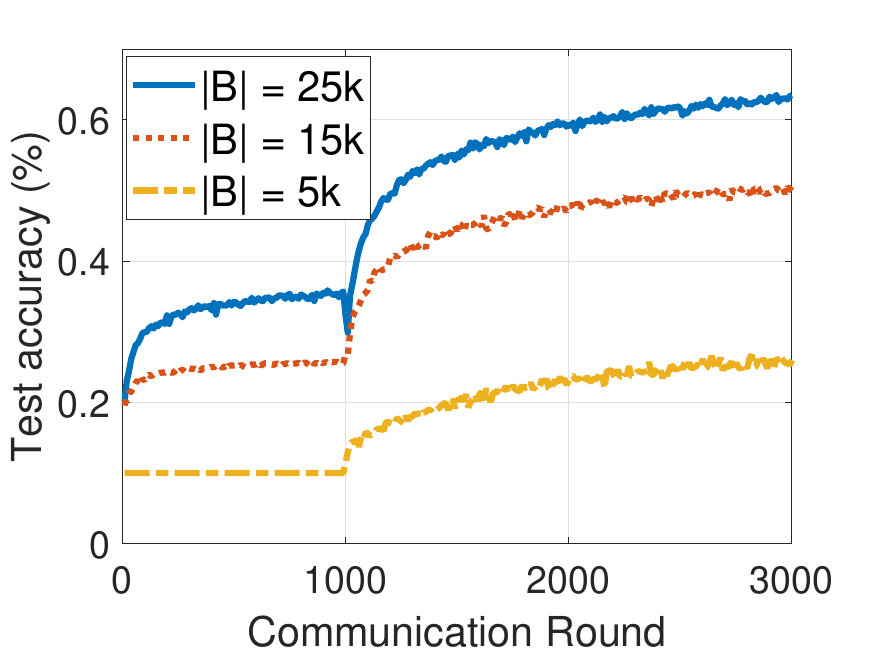}
    \caption{Burst arrival - Random}
    
  \end{subfigure}
  \hfill
  \begin{subfigure}{0.32\textwidth}
    \centering
    \includegraphics[width=\linewidth]{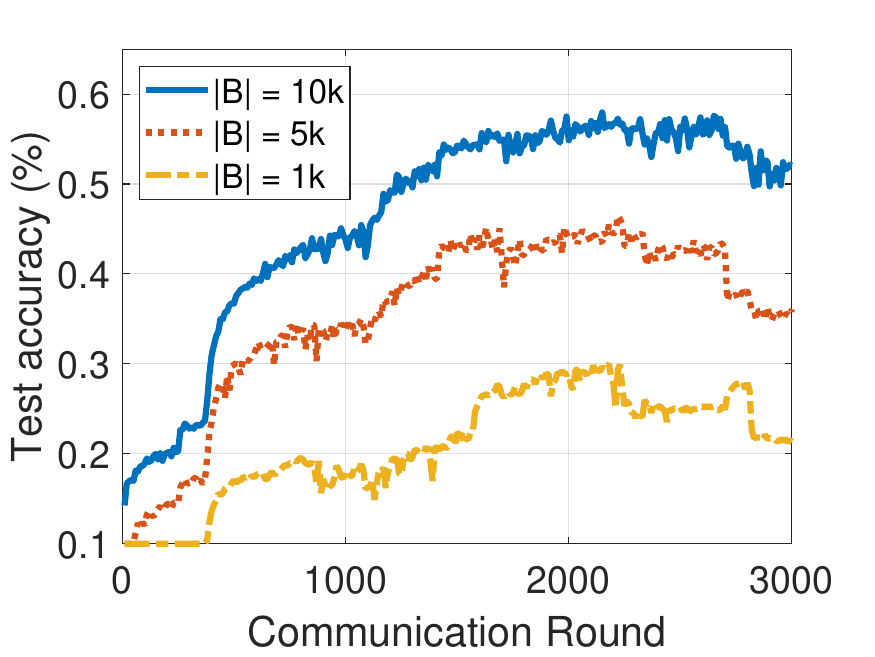}
    \caption{Random arrival - Random}
    
  \end{subfigure}
  
%---------------------------------------------------------
% FIFO Sampling
  \begin{subfigure}{0.32\textwidth}
    \centering
    \includegraphics[width=\linewidth]{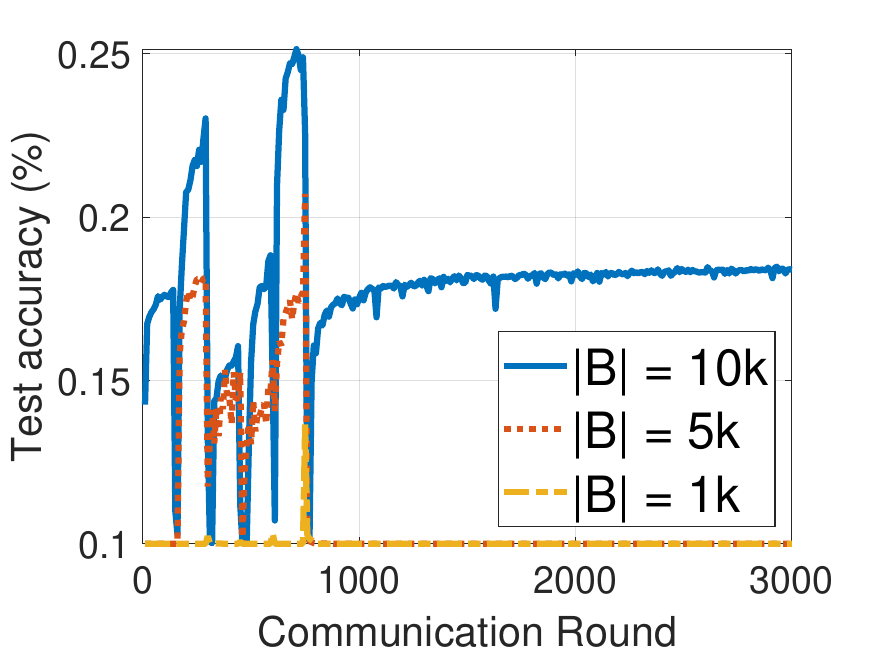}
    \caption{Smooth arrival - FIFO}
    
  \end{subfigure}
  \hfill
  \begin{subfigure}{0.32\textwidth}
    \centering
    \includegraphics[width=\linewidth]{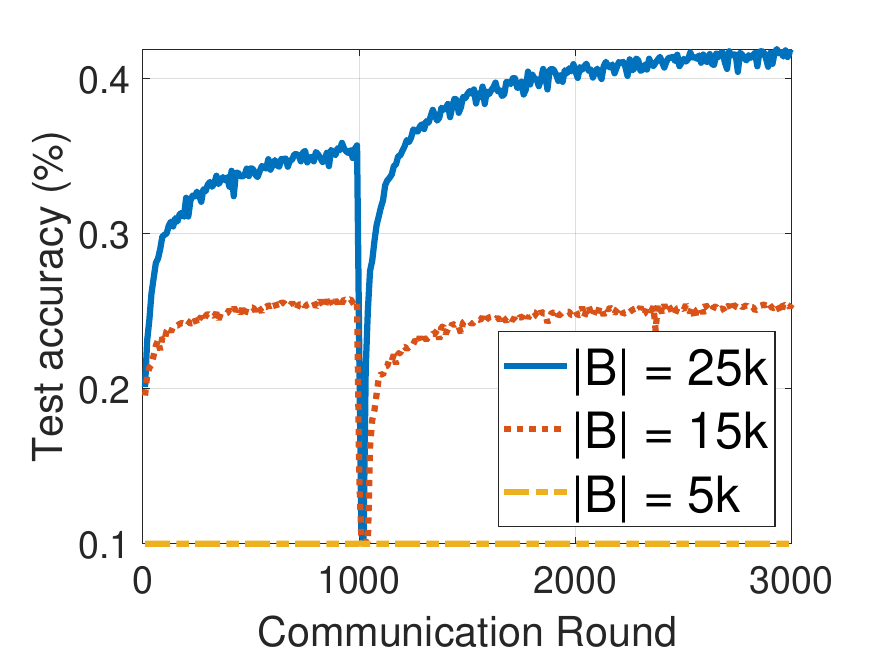}
    \caption{Burst arrival - FIFO}
    
  \end{subfigure}
  \hfill
  \begin{subfigure}{0.32\textwidth}
    \centering
    \includegraphics[width=\linewidth]{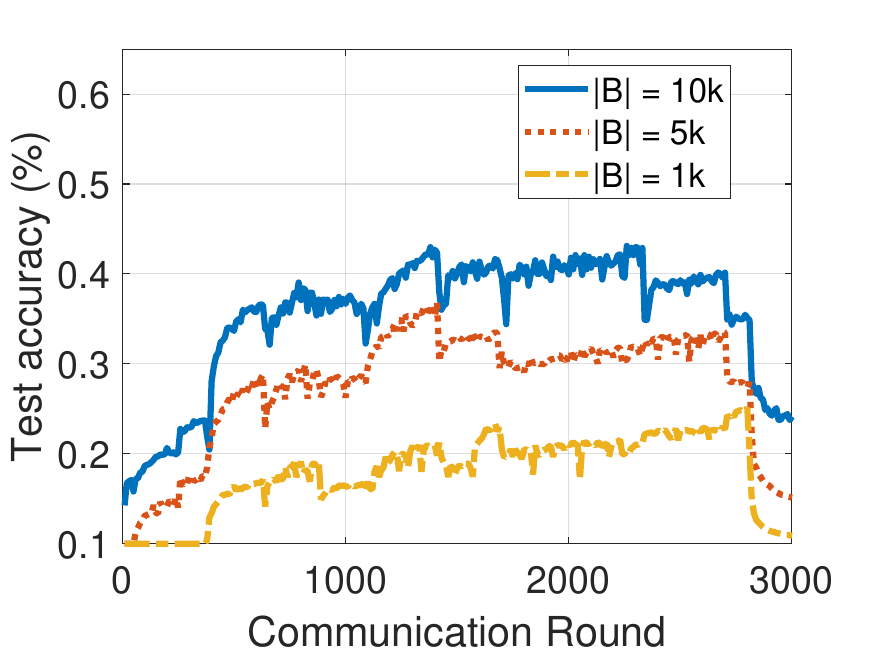}
    \caption{Random arrival - FIFO}
    
  \end{subfigure}
  
  \caption{The impact of buffer size $|B|$ on DYNAMITE under different sampling methods and arrival patterns}
  \label{fig:buffer_samp_arrive}
\end{figure*}

We first present the proof of Theorem \ref{thm:opt_uniform} with uniform batch-size. Given that the objective function is monotonically decreasing with the client batch size, we can obtain the optimal uniform batch size $s^*(\tau)$ by finding its maximum value under both completion time and training cost constraints. Then we can substitute $s^*(\tau)$ into the objective function, i.e., $f(\tau) = q^{K\tau}G(0)+\frac{1-q^K}{1-q}\left(\frac{\beta\eta^2(1-q^{\tau})}{2D^2(1-q)}\sum\limits_{i\in \mathcal{N}}\frac{M_iD_i^2}{s^*(\tau)}+\rho h(\tau)^2\right)$, which can be easily proved to be a convex function when $\tau<2/log(1/q)$. Since the value of $q = 1 - \eta c \mu$ is only slightly less than $1$ , this interval of $\tau$  can be large enough. Thus, we can solve the $\hat{\tau}$ to minimize the expected error bound by letting the derivative of $f(\tau)$ to be zero. However, $\hat{\tau}$ can be fractional, so we need to compare the values of $f(\lfloor\hat{\tau}\rfloor)$ and $f(\lceil\hat{\tau}\rceil)$ to find the optimal $\tau^*$. In Corollary \ref{cor:bs-gpu}, the optimal batch size $s_i^*$ for each client $i$ is determined by the Cauchy-Schwarz inequality and the total batch size $s_{tot}(\tau^*)$, where the optimal $\tau^*$ can be obtained following the same procedures as described in the proof of Theorem \ref{thm:opt_uniform}.

\section{Proof of Theorem \ref{thm:opt_alg_1} (Algorithm 1)}
\label{sec:proof_alg}

\begin{figure*}[t]
    \centering
	\begin{subfigure}[t]{0.32\linewidth}
		\includegraphics[width=\textwidth]{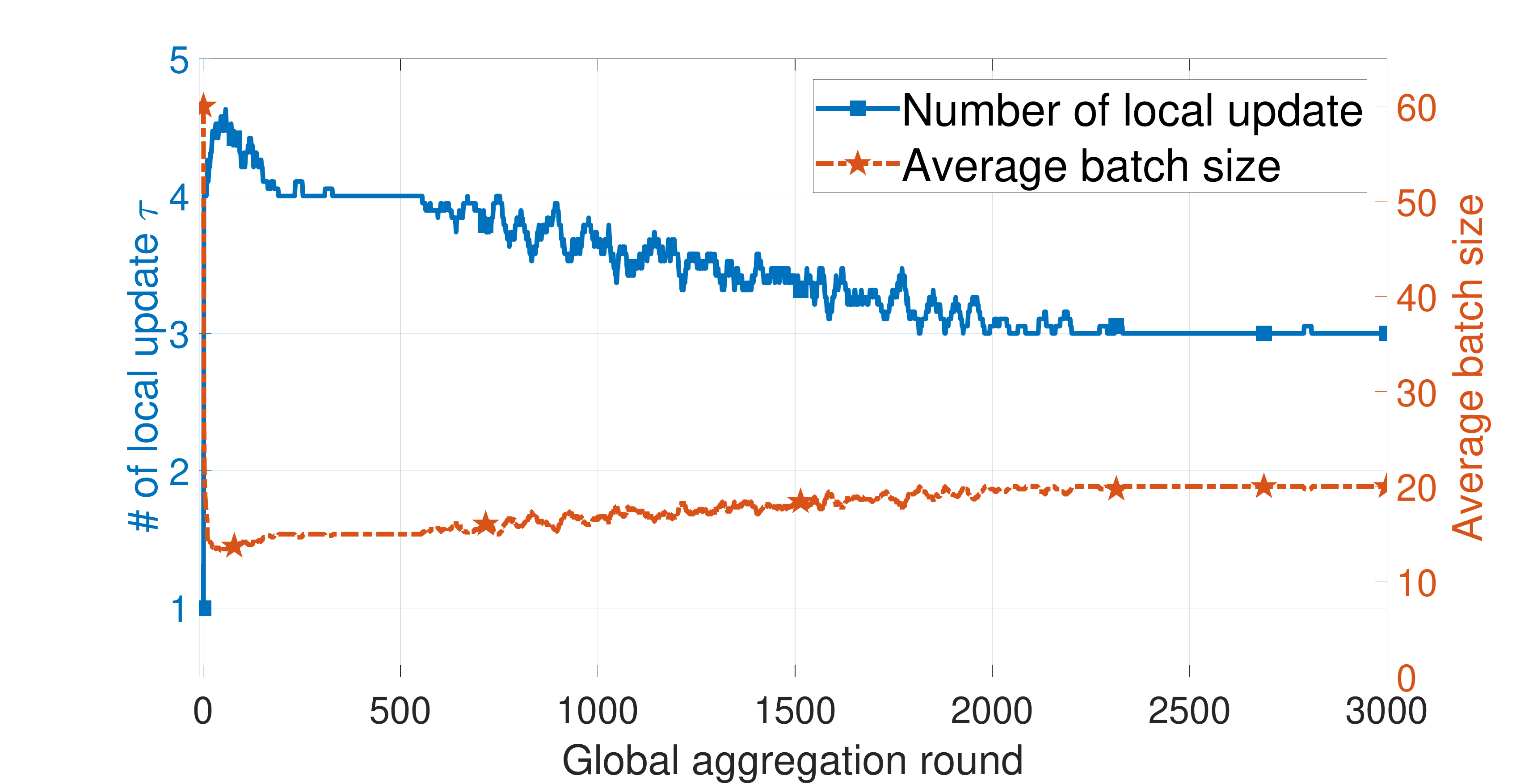}
		\caption{Smooth arrival}
		
	\end{subfigure}
	\begin{subfigure}[t]{0.32\linewidth}
		\includegraphics[width=\textwidth]{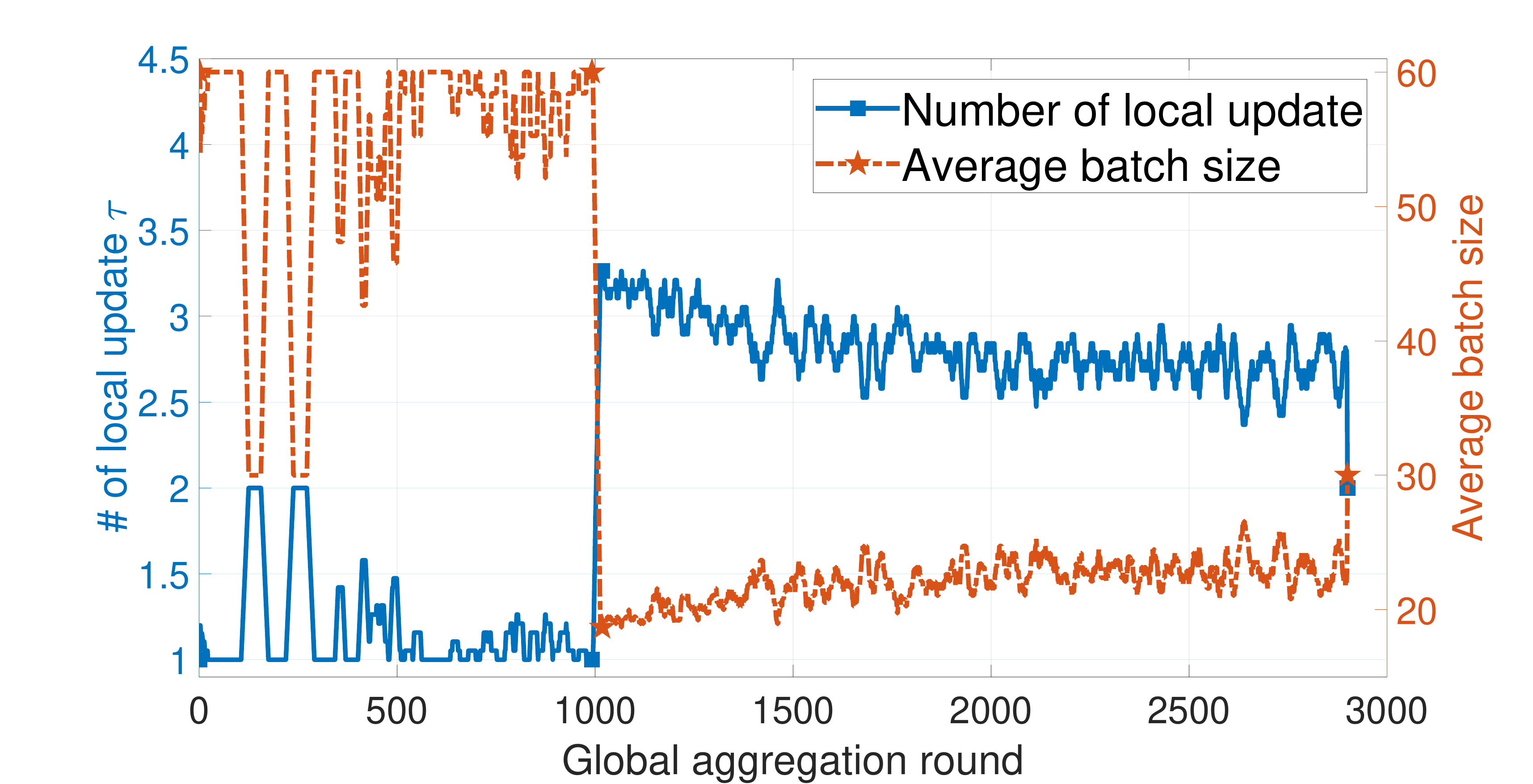}
		\caption{Burst arrival}
		
	\end{subfigure}
	\begin{subfigure}[t]{0.32\linewidth}
		\includegraphics[width=\textwidth]{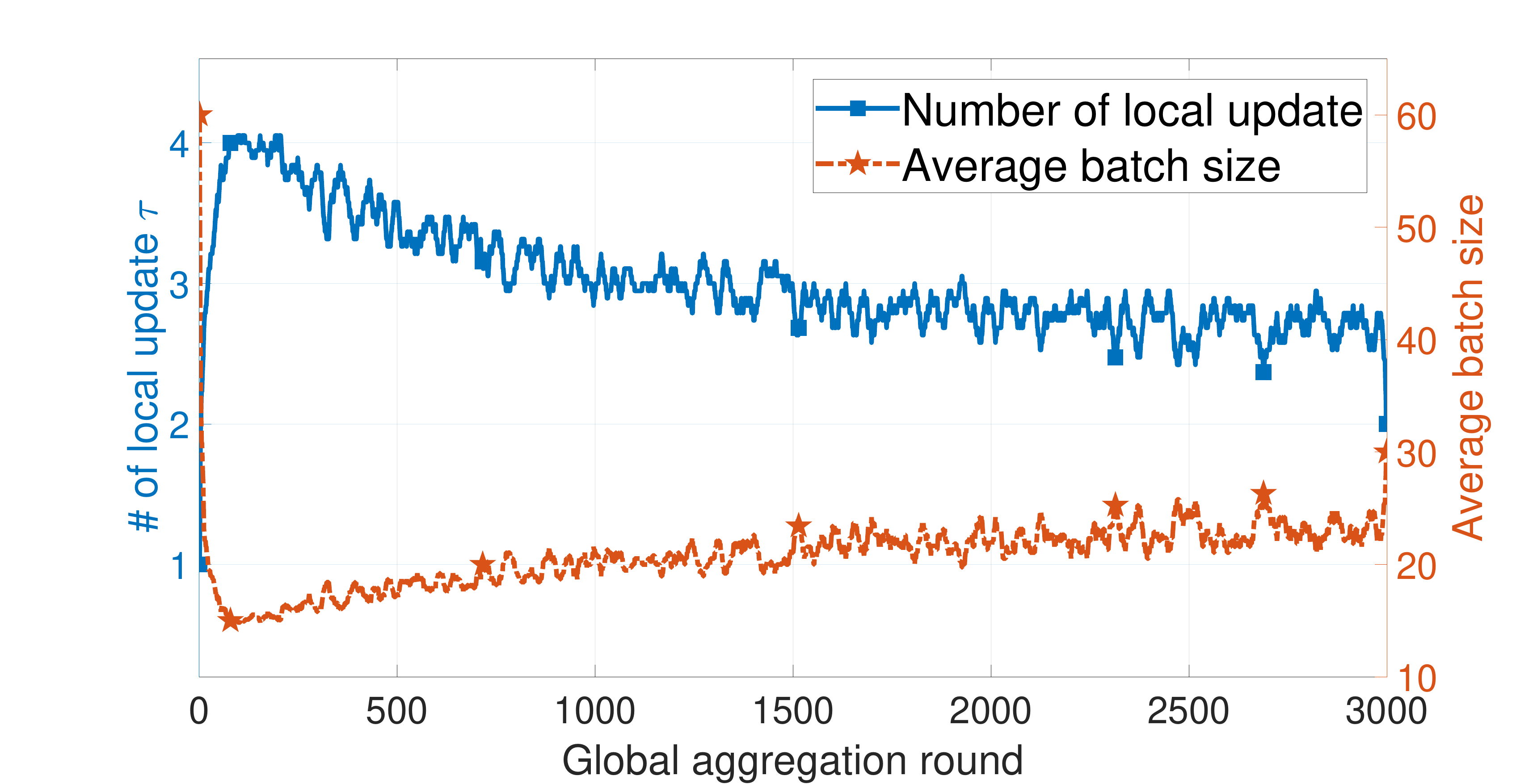}
		\caption{Random arrival}
		
	\end{subfigure}
	\caption{The change in the number of local updates and average batch size under streaming datasets CIFAR-10} 
	\label{fig:opttau_stream}
\end{figure*}
The optimality of the initial assignment based on Cauchy–Schwarz inequality has been proved in Section \ref{sec:case3}. Here we continue to prove that the batch size of the 'time-constrained' devices $(s_i\geq s_{i}(\theta))$ will always satisfy $s_i=s_{i}(\theta)$ in the optimal batch-size distribution.

\begin{lem}\label{lem:time-constrained}
	Compared to a normal device $j$, a time-constrained device $i$ satisfy:
	\begin{equation}\label{eq:md_compare1}
		\frac{\sqrt{M_i}D_i}{s_i(\theta)} > \frac{\sqrt{M_j}D_j}{s_j}
	\end{equation}
	\begin{equation}\label{eq:md_compare2}
		\frac{M_iD_i^2}{s_i(\theta)}+\frac{M_jD_j^2}{s_j} < \frac{M_iD_i^2}{s_i(\theta)-1}+\frac{M_jD_j^2}{s_j+1}
	\end{equation}
	
\end{lem}
Every device $i$ will have the same value of $\frac{\sqrt{M_i}D_i}{s_i}$ after the initial batch-size distribution based on Cauchy inequality. But the 'time-constrained' devices have to reduce their batch-size $s_i$ to $s_{i}(\theta)$ due to the limited time budget and yields Lemma \ref{lem:time-constrained}.
We can assume a batch-size distribution scheme $[s_1, ..., s_i(\theta), s_j,..., s_N]$. According to (\ref{eq:md_compare2}), we can never find a better batch-size distribution scheme $[s_1, .., s_i(\theta)-1, s_j+1,..., s_N]$ to obtain a smaller objective function, where device $j$ can be any other normal devices. Therefore, the batch size of the 'time-constrained' devices $(s_i\geq s_{i}(\theta))$ will always satisfy $s_i=s_{i}(\theta)$ in the optimal batch-size distribution, and we can exclude these constrained devices from the following batch-size computation. 

As for the final adjustment, the value of $\frac{M_iD_i^2}{s_i}-\frac{M_iD_i^2}{s_i+1}$ represents the reduction of the objective function if the batch-size of device $i$ plus one, so we can still have the optimal solution if we increase the batch-size of the devices with the largest value of $\frac{M_iD_i^2}{s_i(s_i+1)}$ one at a time.

\section{Error bound with client selection}
\label{sec:bound_client_select}

\begin{thm}[Error bound with heterogeneous batch sizes $s_i$, local update $\tau$ and client selection]\label{thm:bound-client-select}
	Suppose that the loss functions satisfy Assumptions 1-4 proposed in \cite{li2019}, and $L, \mu, \sigma_k, G$ defined therein. If $N$ clients are selected randomly with replacement according to the sampling probabilities $p_1,...,p_N$, given the initial global parameter $\mathbf{w}(0)$, the expected error after $K$ aggregation rounds with $\tau$ local updates per round is:
	\begin{equation*}
	\begin{split}	\mathbb{E}\left[F\left(\mathbf{w}(K\tau)\right)\right]-F^{*} &\leq \frac{\kappa}{\gamma+K\tau-1} \cdot\\ &\left(\frac{2 (B+C)}{\mu}+ \frac{\mu \gamma}{2} \mathbb{E}\left\|\mathbf{w}(0)-\mathbf{w}^{*}\right\|^{2} \right),
 \end{split}
	\end{equation*}where
		$\kappa = L/\mu, \gamma = \max\{8\kappa, \tau\}, \Gamma = F^{*} -\sum_{i=1}^{N}p_iF_i^{*}, B=\sum_{i=1}^{N} \frac{p_{i}^{2} \sigma_{i}^{2}}{s_i}+6 L \Gamma+8(\tau-1)^{2} G^{2}, C = \frac{4}{K}\tau^2G^2$.
\end{thm}

Following the main proof of \cite{li2019}, this theorem can be easily derived by analyzing different variance reductions brought by the heterogeneous batch size ($s_i$ in term $B$) across clients, using the same theoretical analysis as described in Lemma \ref{lem:var_s}. This implies that we can incorporate the device selection mechanism into the DYNAMITE algorithm based on this theorem by adopting a similar workflow as presented in our DYNAMITE algorithm. 
\section{The impact of buffer size on DYNAMITE under different arrival patterns and sampling methods}
\label{sec:buffer_impact}

Figure \ref{fig:buffer_samp_arrive} shows that, under different arrival patterns and sampling methods, DYNAMITE always has a better model performance and faster convergence rate with a larger buffer. These experimental results are consistent with our theoretical expectations and intuitive understanding. The insight is that clients can store more diverse training samples with a larger buffer, leading to a more stable and better FL training process. Additionally, compared to smooth arrival and random arrival, DYNAMITE is more sensitive to the buffer size under burst arrival scenarios, where most data samples are received in a short period of time, which poses greater challenges to clients with limited buffers. We also observe that DYNAMITE often requires a larger buffer to reach the same test accuracy in burst arrival settings (Burst arrival: 25k, Smooth and Random arrival: 10k). Further, the reservoir sampling method adopted by DYNAMITE algorithm consistently yields superior model performance across all buffer size settings. In contrast, the other sampling methods (Random sampling and FIFO sampling) have demonstrated varying degrees of catastrophic forgetting, resulting in model degradation issues due to their biased selection principles.

\section{The number of local updates and average batch sizes of DYNAMITE under streaming datasets}
\label{sec:Appendix_taubs_stream}

In Figure \ref{fig:opttau_stream}, we present the change in the number of local updates $\tau$ and average batch size $\sum_{i\in [N]} s_i/N$ during the FL training using \textbf{DYNAMITE} algorithm under streaming datasets. The new experiments are performed under three different data arrival patterns (explained in Section 7.1.3) to fully simulate the data dynamics in online training. In the case of smooth arrival and random arrival patterns of streaming datasets, the number of local updates demonstrates a general decrease while the average batch size displays a consistent increase during the training process. Compared to the smooth arrival pattern, both of these two variables exhibit more pronounced fluctuations under the random arrival pattern due to the higher degree of uncertainty in random arrival settings. On the other hand, we have observed that, under burst arrival settings, the number of local updates exhibits an abrupt increase in the 1000-th round, precisely aligning with the time when the participating clients receive an enormous amount of data and therefore have a significant increase in the model training loss. These results match our theoretical analysis in Theorem \ref{thm:bound-vanilla} and Remark \ref{rmk:opt_tau_1}, as well as the experimental results in static datasets shown in Figure \ref{fig:case1_optau}.

\end{appendices}

}

\end{document}